%% file: main_arxiv.tex
\documentclass[10pt,twocolumn,letterpaper]{article}

\usepackage{cvpr}              %
\usepackage{utfsym}
\usepackage[abs]{overpic}
\usepackage{multirow}
\usepackage{xcolor}
\usepackage[sectionbib]{chapterbib}
\usepackage{microtype}

\definecolor{over_pink}{RGB}{194,123,160}
\definecolor{over_blue}{RGB}{118,165,175}
\definecolor{over_green}{RGB}{147,196,125}
\definecolor{over_orange}{RGB}{221,126,107}
\input{definitions}

\input{preamble}
\definecolor{cvprblue}{rgb}{0.21,0.49,0.74}
\usepackage[pagebackref,breaklinks,colorlinks,citecolor=cvprblue]{hyperref}
\usepackage{diagbox}

\def\paperID{00268} %
\def\confName{3DV\xspace}
\def\confYear{2024\xspace}

\title{\close{}: A 3D Clothing Segmentation Dataset and Model}

\author{\begin{tabular}{ccccccccccccccc}\multicolumn{2}{c}{\hspace{-0.6cm}Dimitrije Antić \textsuperscript{1}} & \multicolumn{2}{c}{Garvita Tiwari \textsuperscript{2,3,4}} & \multicolumn{2}{c}{Batuhan Ozcomlekci \textsuperscript{2}} & \multicolumn{2}{c}{Riccardo Marin \textsuperscript{2,3}} &  \multicolumn{2}{c}{Gerard Pons-Moll \textsuperscript{2,3,4}} \end{tabular}\\
{\small\textsuperscript{1}University of Amsterdam, Netherlands \qquad \textsuperscript{2}University of Tübingen, Germany \qquad \textsuperscript{3}Tübingen AI Center, Germany}    \\
{\vspace{-0.2cm}\small\textsuperscript{4}Max Planck Institute for Informatics, Saarland Informatics Campus, Germany\vspace{0.2cm}
}\\
{\vspace{-0.1cm}\tt\scriptsize d.antic@uva.nl, gtiwari@mpi-inf.mpg.de, batuhan.oezcoemlekci@student.uni-tuebingen.de,} \\ {\vspace{-0.2cm} \tt\scriptsize \{riccardo.marin, gerard.pons-moll\}@uni-tuebingen.de} \vspace{-5mm}}

\begin{document}
\makeatletter
\let\@oldmaketitle\@maketitle%
\renewcommand{\@maketitle}{
	\@oldmaketitle%
		\begin{center}

	\begin{overpic}[width=\textwidth,unit=1mm]{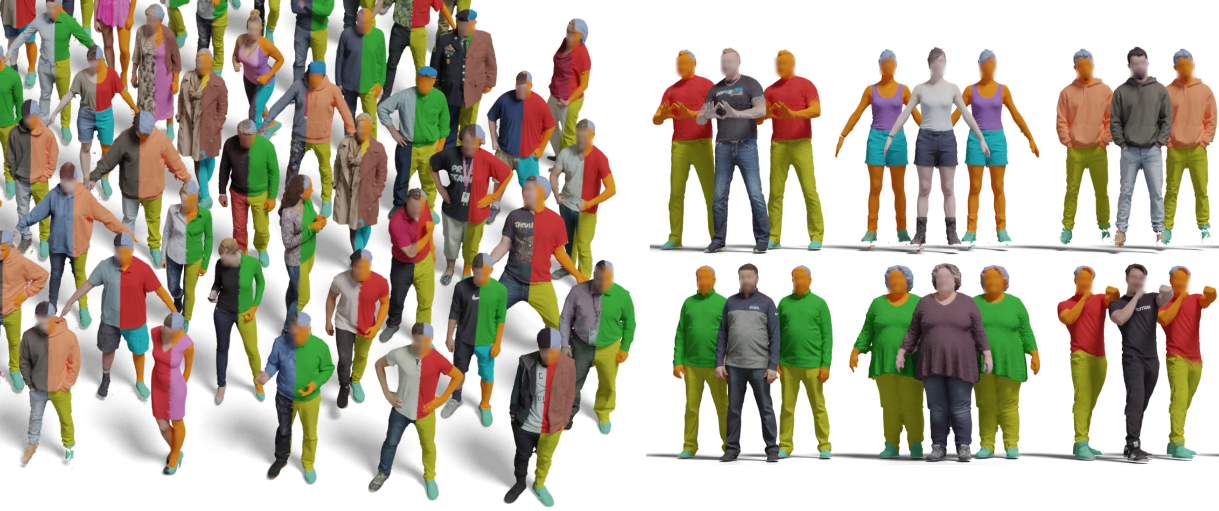}
 \scriptsize
                \put(124,67){GT}
                \put(130.5,67){Input}
   	        \put(137.5,67){\clothmodel{}}
\end{overpic}
    	\end{center}
     \vspace{-16pt}

\refstepcounter{figure}\normalfont Figure~\thefigure.  \emph{Left}: We present \textbf{\clothdata{}}, a large-scale dataset of people in clothing with fine-grained segmentation labels. We use this dataset to train our clothing segmentation model, \textbf{\clothmodel{}}, tailored to segment clothing from 3D scans.  \emph{Right}: We show results of \clothmodel{} on the diverse set of scans, where each instance represents GT, Input, and Prediction.
	\label{fig:teaser}
	\newline
 
}
\makeatother
\maketitle

\input{sec/00_Abstract}    
\input{sec/01_Introduction_Ric}

\input{sec/02_RelatedWork}

\input{sec/03_Dataset}

\input{sec/04_Method}

\input{sec/05_Results}

\input{sec/06_Conclusions}

{\small
\noindent{\bf Acknowledgments:} Thanks to RVH, CVLab team, and reviewers for valuable feedback. The project was made possible by funding from the Carl Zeiss Foundation. This work is supported by the Deutsche Forschungsgemeinschaft (DFG, German Research Foundation) - 409792180 (Emmy Noether Programme, project:  Real Virtual Humans), German Federal Ministry of Education and Research (BMBF): Tübingen AI Center, FKZ: 01IS18039A. Gerard Pons-Moll is a member of the Machine Learning Cluster of Excellence, EXC number 2064/1 - Project number 390727645. Riccardo Marin has been supported by the European Union’s Horizon 2020 research and innovation program under the Marie Skłodowska-Curie grant agreement No 101109330. 
}

{
    \small
    \bibliographystyle{ieeenat_fullname}
    \bibliography{main}
}

\clearpage
\newpage

\input{supmat_arxiv}

\end{document}

%% file: definitions.tex
\usepackage{amsmath}

\newcommand{\mat}[1]{\mathbf{#1}}

\newcommand{\vect}[1]{\mathbf{#1}}

\newcommand{\template}[0]{\mat{T}}

\newcommand{\pose}[0]{\boldsymbol{\theta}}
\newcommand{\shape}[0]{\boldsymbol{\beta}}

\newcommand{\smpl}[0]{M}

\usepackage{amsthm}

\newcommand{\clothmodel}{CloSe-Net}
\newcommand{\clothdata}{CloSe-D}
\newcommand{\clothtool}{CloSe-T}
\newcommand{\close}{CloSe}
\newcommand{\sota}{SotA}

\newcommand{\clothclass}{{\mathrm{g}}}
\newcommand{\pointenc}{f_{\mathrm{point}}}

\newcommand{\segdecoder}{f_{\mathrm{dec}}}

\newcommand{\pointfeat}{F^{\mathrm{p}}}
\newcommand{\bodyfeat}{F^{\mathrm{b}}}
\newcommand{\clothfeat}{F^{\mathrm{c}}}
\newcommand{\allfeat}{F^{\mathrm{all}}}

\newcommand{\bodyencoder}{Body Encoder}
\newcommand{\garencoder}{Clothing Encoder}
\newcommand{\pcencoder}{Point Encoder}

\newcommand{\figref}[1]{Fig.~\ref{#1}}
\newcommand{\secref}[1]{Sec.~\ref{#1}}
\newcommand{\eqnref}[1]{Eq.~\eqref{#1}}
\newcommand{\tabref}[1]{Table~\ref{#1}}

\newcommand{\intclothdata}{CloSe-Di}
\newcommand{\comclothdata}{CloSe-Dc}

\newcommand{\nclasses}{18}

\newcommand{\totalscans}{3167}

\newcommand{\mgn}{MGN-Seg~\cite{bhatnagar2019mgn}}

\newcommand{\nR}{\mathbb{R}}

\newcommand{\cC}{\mathcal{C}}

\newcommand{\cE}{\mathcal{E}}
\newcommand{\cF}{\mathcal{F}}
\newcommand{\cG}{\mathcal{G}}

\newcommand{\cL}{\mathcal{L}}

\newcommand{\cV}{\mathcal{V}}

\DeclareOldFontCommand{\bf}{\normalfont\bfseries}{\mathbf}

%% file: sec/00_Abstract.tex
\urlstyle{same}
\begin{abstract}
3D Clothing modeling and datasets play crucial role in the entertainment, animation, and digital fashion industries. Existing work often lacks detailed semantic understanding or uses synthetic datasets, lacking realism and personalization. To address this, we first introduce \textbf{\clothdata{}}: a 
novel large-scale dataset containing 3D clothing segmentation of $\totalscans$ scans, covering a range of $\nclasses$ distinct clothing classes. Additionally, we propose \textbf{\clothmodel{}}, the first learning-based 3D clothing segmentation model for fine-grained segmentation from colored point clouds. \clothmodel{} uses local point features, body-clothing correlation, and a garment-class and point features-based attention module, improving performance over baselines and prior work. The proposed attention module enables our model to learn appearance and geometry-dependent clothing prior from data. We further validate the efficacy of our approach by successfully segmenting publicly available datasets of people in clothing. We also introduce \textbf{\clothtool{}}, a 3D interactive tool for refining segmentation labels. Combining the tool with \clothmodel{} in a continual learning setup demonstrates improved generalization on real-world data. Dataset, model, and tool can be found at \url{https://virtualhumans.mpi-inf.mpg.de/close3dv24/}.
\end{abstract}

%% file: sec/01_Introduction_Ric.tex
\section{Introduction}
\label{sec:intro}
Clothing plays a vital role in shaping our identity. Our dressing choices contribute to our representation, conveying cultural traits, religious beliefs, geographical origin, or mood. 
In light of the recent attention on the expressiveness of avatarization processes, Computer Vision scholars dedicate tremendous efforts in acquiring, modeling, and comprehending digital clothing, enabling uncountable applications: from digital fashion to AR/VR; from home entertainment to industrial-scale content creation. 

Semantic understanding of humans in clothing from 2D images has seen significant progress~\cite{pgn,u2net,iMaterialist,yang2014cvpr}, but the lack of geometrical information is an obstacle for AR/VR applications, where actions occur in a 3D world. Despite the development of 3D data capture techniques~\cite{3dmd,lumaai,treedys} and the consequent abundant geometrical data of 3D people in clothing~\cite{cai2022humman,thuman2,cape,Jinka2022}, garment analysis and its semantic understanding remains an open problem.

The primary challenge lies in representing 3D digital garments and their semantics. Certain existing approaches rely on synthetic clothing meticulously designed by experts~\cite{cloth3d,patel2020} or involve expensive acquisitions~\cite{deepfashion3d,deepwrinkles}, where accurate information comes at the expense of scalability. Some recent methods address human and clothing models within a unified representation~\cite{tiwari21neuralgif,scanimate,chen2022gdna} but lack a semantic understanding of the distinct parts. We posit that the absence of large-scale and high-quality segmented datasets serves as a fundamental barrier.

Consequently, the development of robust methods for understanding the 3D cloth semantics has been hampered. Prior work like~\mgn{} requires expensive pipelines involving rendering, SMPL+D registration~\cite{lazova2019360}, 2D segmentation~\cite{pgn}, and hand-crafted clothing priors, making it time-consuming (15-20 minutes per scan) and prohibitive for complex clothing items. In contrast, GIM3D~\cite{gim3d} uses a \sota{} part segmentation method trained on synthetic data, accounting only for geometric information, namely location and normals. Both are limited by 3-class prediction: upper and lower garments and the human body, which trivialize the problem. Considering the clothes variations in style and textures, getting semantic information from 3D scans requires fast, accurate, generalizable, and scalable method. 

This work aims to fill this gap with a three-fold contribution. First, we introduce \textit{\clothdata{}}, a large-scale dataset of people in clothing with fine-grained segmentation labels, for a total of $\totalscans$
scans and \nclasses{} garments categories. To our knowledge, this is the first real-world dataset with such fine-grained segmentation labels. Secondly, we use \clothdata{} to train \textit{\clothmodel{}}, a 3D clothing segmentation model to predict \nclasses{} distinct types of clothing. Our model is based on two key intuitions: we correlate body parts with clothing classes, leveraging the SMPL~\cite{SMPL:2015} body model(\secref{sec:canon}); and we address the relationship between local geometric-appearance cues and clothing class in segmentation using an attention module (\secref{sec:attention}), learning associations between point features and clothing classes. 
We demonstrate a significant improvement over baselines and prior works (\secref{sec:results} and \secref{sec:prior}).
Finally, to achieve the most comprehensive generalization possible, we develop \textit{\clothtool{}}, an interactive 3D tool to provide quick human feedback/annotation. This tool allows users to refine segmentation predictions and rectify segmentation labels. Our tool can also be integrated with \clothmodel{}, where the feedback is backpropagated, and the network is fine-tuned in a continual learning setup(~\secref{sec:realworld}). We leverage this tool to prepare high-quality segmentation training data \clothdata{}. We also use the tool in conjunction with \clothmodel{} to prepare high-quality 
segmentation labels on public datasets, which we release as \clothdata++.

In summary, our contributions are:
\begin{itemize}
    \item \textbf{\clothdata{}}: A high-quality fine-grained clothing segmentation dataset containing segmentation labels for $\totalscans$ scans, covering $\nclasses$ clothing classes. %
    \item \textbf{\clothmodel{}}: A human prior and clothing classes attention-based 3D clothing segmentation method that outperforms baselines and prior work.
    \item \textbf{\clothtool{}}:  A 3D interactive tool to refine the model in a continual learning framework, improving generalization to new datasets.
    \item \textbf{\clothdata++}: Fine-grained semantic segmentation for a subset of publicly available real-world datasets.
\end{itemize}

We release our data, model, and tool for further research.

%% file: sec/02_RelatedWork.tex
\section{Related Work}
\label{sec:relatedwork}

Our work includes a new 3D clothing dataset, a 3D segmentation model, and an interactive refinement method, thus the related work covers these three areas.

\subsection{3D Clothing Datasets}
\label{rw:dataset}
The rise of learning-based digital fashion and virtual try-on led to the creation of image-based clothing datasets with semantic labels~\cite{iMaterialist,liu2016deepfashion}. However, these datasets lack pose variation, are 2D and mostly frontal, and are unsuitable for learning overall human/clothing shape, deformations, and 3D/4D models. 

3D/4D Clothing datasets can be grouped into two categories: Synthetic and Captured ones. Synthetic datasets~\cite{patel2020,garmentdesign_Wang_SA18,cloth3d,resynth} are obtained using physics-based simulation software~\cite{md}, their generation requires expert intervention and mostly contain geometric information without texture. Moreover, this may not scale for complex clothing and multiple layers, often resulting in non-realistic deformations. 

On the other hand, the accessibility of capturing datasets has increased recently, courtesy of advancements in 3D/4D capture systems~\cite{3dmd, treedys}. THuman1-4~\cite{thuman2,thuman3,thuman4,thuman5} propose medium to high-quality static scans of individuals in limited clothing styles/variations with limited pose variations. BUFF/CAPE~\cite{buff,cape,clothcap}, and HuMMan~\cite{cai2022humman} provide dynamic scans of subjects in different clothing items. These datasets contain point clouds, occasionally texture, and SMPL parameters, but lack clothing semantic segmentation. Prior work such as MGN~\cite{bhatnagar2019mgn}, SIZER~\cite{tiwari20sizer}, GIM3D~\cite{gim3d} provide coarse 3D clothing segmentation labels, containing only three categories, namely \emph{upper garment, lower garment, and body}. Deepfashion3D~\cite{deepfashion3d} is a dataset of high-quality and diverse 3D Clothing items, scanned on mannequins consisting of 10 clothing classes with keyline annotations and SMPL pose parameters. However, the clothing items are scanned separately and cannot be used to create a fully clothed person. In contrast, our dataset is more realistic, featuring a broader range of clothing classes.

None of the existing real-world clothing datasets contains fine-grained clothing segmentation labels of clothed humans. In \clothdata{}, we provide clean and fine-grained clothing segmentation labels of scans along with colored point clouds, and SMPL~\cite{SMPL:2015} parameters. We compare \clothdata{} with existing real-world datasets in~\tabref{tab:dataset}.

\begin{table}[t]
\centering
\resizebox{0.48\textwidth}{!}{
\fontsize{40pt}{40pt}\selectfont
\begin{tabular}{lcccccc}
\toprule
\textbf{Dataset} & \textbf{$\#$ scans }& \textbf{Segmentation} &  \textbf{Garment Class} & \textbf{Texture} & \\ 
\\

\midrule

MGN~\cite{bhatnagar2019mgn}                         & $\sim$300        & \usym{1F5F8} &   3    & \usym{1F5F8}\\
SIZER~\cite{tiwari20sizer}                          & $\sim$2000       & \usym{1F5F8} &   3    
 &   \usym{1F5F8}\\
DeepFashion3D~\cite{deepfashion3d}                    & 2078(563) & \usym{1F5F8} &   10    & \usym{1F5F8}\\
THuman~\cite{thuman2,thuman3,multihuman}                    & $\sim$1000 & \usym{2718} &   -   & \usym{1F5F8}  \\
\midrule

\clothdata                                          & $\sim$3000         & \usym{1F5F8} &   18  &  \usym{1F5F8}   \\
\clothdata ++                                          &    $\sim$( +1000 )   & \usym{1F5F8} &   18   &  \usym{1F5F8}  \\

\bottomrule
\end{tabular}

}
\caption{Compared to current 3D clothing(real and static) datasets, \clothdata{} is the first extensive dataset featuring fine-grained clothing segmentation labels and a variety of clothing items.}
\label{tab:dataset}
\end{table}

\subsection{3D Clothing Segmentation}
\label{rw:partsegmentation}

 2D clothing segmentation and human parsing have been extensively studied. Several prior works~\cite{yamaguchi2013paperdoll,crowdedscenes,mhparser,pgn,parsingclothing} propose learning-based human parsing in images, trained using 2D clothing datasets such as~\cite{lip,iMaterialist}. These methods exploit human body parts and pose information to improve accuracy and generalization. However, they cannot be used for 3D segmentation, as they are not trained to produce multi-view consistent results, and lifting labels from 2D to 3D requires a slow optimization process. 
 
\mgn{} is an optimization-based 3D clothing segmentation method that involves registering scans to SMPL+D~\cite{bhatnagar2019mgn,bhatnagar2020ipnet}, applying PGN~\cite{pgn} for 2D segmentation of multi-view renderings of a mesh. These 2D segmentation images are then lifted back to 3D by solving GrabCut~\cite{rother2004grabcut} in SMPL-UV space with a handcrafted clothing class based prior. This takes roughly $20$ minutes per scan, is limited to 3 clothing classes, and requires expensive SMPL+D registration and hand-crafted clothing prior. On the other hand, \clothmodel{} only needs colored point cloud, SMPL($\pose, \shape$) parameters, and clothing classes present in the scan and learns clothing prior from data. 

3D clothing segmentation from a point cloud resembles a part segmentation setup. There are various existing works in learning-based 3D part segmentation, such as the seminal PointNets~\cite{8099499_pointnet,pointnet2} and self-attention-based PointTransformer~\cite{pointtransformer}. More recent methods like KPConv~\cite{thomas2019KPConv} use kernel points defined in Euclidean space to apply convolution. SGPN~\cite{wang2018sgpn} uses a single network to predict point grouping proposals and a corresponding semantic class for each proposal. Recent \sota{} methods~\cite{3d_segmentation_survey} like DGCNN~\cite{dgcnn} introduces EdgeConv, a differentiable layer representing data on graphs dynamically computed in each network layer. DeltaConv~\cite{Wiersma2022DeltaConv} uses anisotropic convolution layers and introduces a convolution layer that combines geometric operators from vector calculus to construct anisotropic filters on point clouds. Existing works lack evaluations on clothing segmentation and do not integrate human-clothing-specific knowledge. In contrast, \clothmodel{} employs a DGCNN-based point-feature module, leveraging human body and clothing priors for improved performance in clothing segmentation. We also show in our experiments that \sota{} DGCNN and DeltaConv, have limitations when applied to this task. 

Most relevant to our work is GIM3D~\cite{gim3d}, which uses \sota{} part segmentation methods to segment directly from a point cloud, but similar to~\cite{bhatnagar2019mgn,tiwari20sizer} it is limited to three classes. To our knowledge, no existing method directly operates on a colored 3D point cloud/mesh to generate fine-grained 3D clothing segmentation. GIM3D+~\cite{gim3dplus} extends GIM3D, by including more diverse fabrics, sizes, and poses in the dataset, but is still limited to three classes and doesn't consider texture information.

\subsection{3D Interactive Segmentation and Refinement}
\label{rw:interactive}

Tools for interactive segmentation refinement are crucial for developing large-scale datasets and incorporating human feedback for improving segmentation. In 2D, classical methods like GrabCut~\cite{rother2004grabcut} and a more recent interactive segmentation refinement~\cite{petrov20fBRS} use human feedback input to improve segmentation. In 3D, 3D-GrabCut~\cite{Meyer20153DGI} and mesh cutting tools~\cite{easymeshcutting} are used for foreground/background segmentation. However, these methods do not leverage learned neural features and initial predictions. Recent works like~\cite{iseg3d,interacitve_siyu} are related to 3D interactive part segmentation annotation tools. iSeg3D~\cite{iseg3d} utilizes primitive-aware embedding and doesn't consider color information of data. InterObject3D~\cite{interacitve_siyu} proposes a generalizable pipeline for interactive 3D segmentation, where the network is refined based on user clicks for a given target domain.  Yet, none of these methods is designed for clothing segmentation and doesn't consider catastrophic forgetting. We introduce \clothtool{}, a novel, fast, and easy-to-use interactive tool, tailored for the clothing domain. We propose the network refinement in a continual learning setup~\cite{continualsurvery}, such that the network not only performs well on the target domain but also learns from it, improving generalization. Prior works like~\cite{8107520,Kirkpatrick_2017} introduce weighted loss term and EWC(elastic weight consolidation) to avoid catastrophic forgetting. We use~\cite{8107520} based weighted loss in our setup.

%% file: sec/03_Dataset.tex
\section{\clothdata{}: 3D \underline{Clo}thing \underline{Se}gmentation \underline{D}ataset }
\label{sec:closed}

We introduce \clothdata{}, a large-scale 3D clothing segmentation dataset that contains labels for $\totalscans$ scans comprising of $\nclasses$ garment classes. \clothdata{} consists of scans from two kinds of sources: 1) \intclothdata{}: Scans captured using Treddy~\cite{treedys} scanner, and 2) \comclothdata{}: Scans from the commercial datasets, such as twindom, renderpeople~\cite{renderpeople,twindom,treedys,axyz}. For \intclothdata{} we will release scans, SMPL parameters, and segmentation labels, while for \comclothdata{} we will provide SMPL parameters and segmentation labels only, due to license concerns. We show examples from our dataset in~\figref{fig:teaser}(left), and the details in~\tabref{tab:dataset_stats}. 

\noindent{\bf Ground Truth Segmentation Labels.} To obtain the ground truth segmentation labels, we adopt the pipeline utilized in~\cite{bhatnagar2019mgn,tiwari20sizer}, as mentioned in~\secref{rw:partsegmentation}. However, unlike \mgn{}, our pipeline does not require SMPL+D registration, as we directly apply all the steps on scan and use~\cite{metashape} for lifting labels to 3D. Due to inconsistent predictions across views and limited generalization of 2D segmentation methods, the 3D segmentation labels may contain noise. To address this, we manually refine the segmentation using \clothtool{}, which is explained in \secref{sec:tool}.

\begin{table}[t]
\centering

\resizebox{0.48\textwidth}{!}{
\fontsize{40pt}{40pt}\selectfont

\begin{tabular}{lcccccccccc}
\toprule
\toprule
{\diagbox{Data}{Class}}  & \textbf{T-shirt} & \textbf{Shirt}  &  \textbf{Vest} & \textbf{Coat} & \textbf{Jacket} & \textbf{Hoodies} & \textbf{Short-Pants}  & \textbf{Pants} & \textbf{Skirts} \\

\midrule
\intclothdata & 415 & 556 & 85 & 191 & - & 209 & 500 & 897 & 59 \\
\comclothdata &  775 & 739 & 107 & 306 & 42 & 42 & 252 & 1404 & 34\\

\midrule
\midrule
{\diagbox{Data}{Class}}   &   \textbf{Dress} & \textbf{JumpS.} & \textbf{SwimS.} & \textbf{UnderG}. & \textbf{Scarf} & \textbf{Hat} & \textbf{Shoes} & \textbf{Body} & \textbf{Hair}  \\

\midrule
\intclothdata & -  & -  & - & - & - & 114 & 1437 & 1455 & 1382   \\
\comclothdata & 50 & 6 & 23 & 10 & 25 & 64 & 1686 & 1732 & 1717 \\

\bottomrule
\bottomrule
\end{tabular}
}
    \caption{Number of scans per clothing class in CloSe-D (\ie, the union of CloSe-Di and CloSe-Dc).}
    
\label{tab:dataset_stats}
\vspace{-0.3cm}
\end{table}

%% file: sec/04_Method.tex
\section{Method}
\label{sec:method}

\begin{figure*}[t]
	\centering

    \begin{overpic}[width=0.98\textwidth,unit=1bp,tics=20]{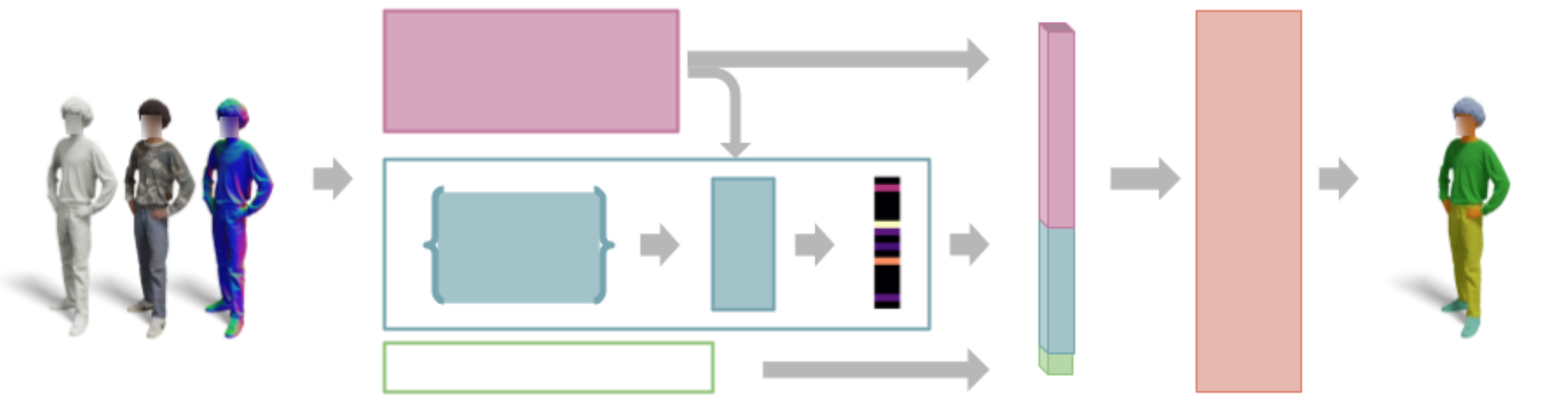}

        \put (5,10){%
        \pgfsetfillopacity{0.0}%
        \colorbox{white}{%
            \parbox{0.15\textwidth}{\pgfsetfillopacity{1} \color{black}\hspace{1pt} $\vect{p}_i = \{ \vect{x}_i | \vect{c}_i | \vect{n}_i\}$} }}

        \put (5,0){%
        \pgfsetfillopacity{0.0}%
        \colorbox{white}{%
            \parbox{0.15\textwidth}{\pgfsetfillopacity{1} \color{black}\hspace{1pt} $\mathrm{SMPL: (\pose, \shape)}, \& \;\mathrm{\clothclass}$} }}
            
    \put (127,102){%
        \pgfsetfillopacity{0.0}%
        \colorbox{white}{%
            \parbox{0.15\textwidth}{\pgfsetfillopacity{1} \color{black}\hspace{1pt} $\mathrm{\pcencoder{}}$} }}

    \put (283,116){%
        \pgfsetfillopacity{0.0}%
        \colorbox{white}{%
            \parbox{0.15\textwidth}{\pgfsetfillopacity{1} \color{black}\hspace{1pt} $\mathrm{\pointfeat_i}$} }}
  \put (225,90){%
        \pgfsetfillopacity{0.0}%
        \colorbox{white}{%
            \parbox{0.15\textwidth}{\pgfsetfillopacity{1} \color{black}\hspace{1pt} $\mathrm{\vect{p'}^2_i} \in \nR^{l}$} }}
   \put(230,50.5){\makebox(0,0){\rotatebox{90}{Attention}}}

                \put (132,50){%
        \pgfsetfillopacity{0.0}%
        \colorbox{white}{%
            \parbox{0.15\textwidth}{\pgfsetfillopacity{1} \color{black}\hspace{1pt} $\mat{G} \in \nR^{ \nclasses \times l}$} }}
        \put (283,60){%
        \pgfsetfillopacity{0.0}%
        \colorbox{white}{%
            \parbox{0.15\textwidth}{\pgfsetfillopacity{1} \color{black}\hspace{1pt} $\mathrm{\clothfeat_i}$} }}

                \put (283,20){%
        \pgfsetfillopacity{0.0}%
        \colorbox{white}{%
            \parbox{0.15\textwidth}{\pgfsetfillopacity{1} \color{black}\hspace{1pt} $\mathrm{\bodyfeat_i}$} }}
            
                \put (134,10){%
        \pgfsetfillopacity{0.0}%
        \colorbox{white}{%
            \parbox{0.25\textwidth}{\pgfsetfillopacity{1} \color{black}\hspace{1pt} $\mathrm{\bodyencoder}$} }}

               \put(389,60){\makebox(0,0){\rotatebox{90}{Segmentation Decoder}}}
               \put(455,10){\makebox(0,0){$y_i$}}
            
    \end{overpic}

	\caption{\textbf{\clothmodel{}}: Given a colored point cloud $\mat{P} = \{ \vect{p}_i \hdots \vect{p}_n \}$ with SMPL parameters ($\pose, \shape$), and clothing classes ($\clothclass$) detected in the scan, where $\vect{p}_i = \{ \vect{x}_i | \vect{c}_i | \vect{n}_i\}$ represent point location, color and normal of a point, \clothmodel{} predicts fine-grained per-point segmentation labels. (a) \textbf{\textcolor{over_pink}{\pcencoder}}(\secref{sec:point}) takes $\mat{P}$, as input and predicts per-point features $\pointfeat$. (b) \textbf{\textcolor{over_blue}{\garencoder}}(\secref{sec:attention}) consists of a learnable codebook $G$ and an attention module, which predicts $\clothfeat$, based on per-point feature $\mathrm{\vect{p'}^2_i}$ and $G$. This $\mathrm{\vect{p'}^2_i}$ is intermediate feature of \pcencoder{}. (c) \textbf{\textcolor{over_green}{\bodyencoder}}(\secref{sec:canon}), finds per-point canonical vertex in SMPL template, given SMPL $\pose$, $\shape$ parameters. (d) Finally, the \textbf{\textcolor{over_orange}{Segmentation Decoder}}(\secref{sec:decoder}) takes $\pointfeat, \clothfeat, \bodyfeat$ and predicts segmentation labels, $y_i$ for $i^\mathrm{{th}}$ point. Solid boxes in model are learnable, while others are fixed. }
	\label{fig:overview}
 \vspace{-0.5cm}
\end{figure*}

In this section, we introduce \clothmodel{} (\secref{sec:close}), a 3D clothing segmentation model that predicts fine-grained clothing labels from a colored point cloud. Additionally, we present \clothtool{} (\secref{sec:tool}), a 3D interactive tool used for creating high-quality segmentation labels of \clothdata{}. We demonstrate the utility of \clothtool{} in enhancing the generalization of our model on real-world datasets.

\subsection{\hspace{-0.13cm} \clothmodel: 3D \underline{Clo}thing \underline{Se}gmentation \underline{Net}work}
\label{sec:close}

\noindent{\bf Overview.} \clothmodel{} predicts fine-grained clothing segmentation labels directly from colored point clouds, given SMPL parameters and clothing classes as input. As shown in~\figref{fig:overview}, \clothmodel{} consists of four modules: \textit{\pcencoder{}}, \textit{\bodyencoder{}}, \textit{\garencoder{}},
and \textit{Segmentation Decoder}. Previous methods (e.g., \mgn{}) manually define clothing priors, leading to poor generalization across garment styles. In contrast, our model learns clothing priors by establishing the correlation between body parts and local clothing, utilizing the \textit{\bodyencoder{}}, and understanding the connection between point features and clothing class through the \textit{\garencoder{}}. As a result, our approach learns a prior that incorporates the garment's style, body information, and the combined local geometric and appearance features.

\noindent{\bf Input/Output.} \clothmodel{} takes a point cloud $\mat{P} \in \nR^{n \times 9}$ as input, consisting of $n$ points, denoted as $\vect{p}_i = \{ \vect{x}_i | \vect{c}_i | \vect{n}_i\}$, where $\vect{x}_i \in \nR^3 $ represent Euclidean coordinates, $\vect{c}_i \in \nR^3 $ represent per-point colors and $\vect{n}_i \in \nR^3 $ represent normals. 
\clothmodel{} predicts per-point segmentation labels, $y_i \in \{1 \hdots K\}, \;$ where $K$ is the number of classes.

As \clothmodel{} incorporates human body prior and clothing class-based attention module, our method also needs SMPL ($\pose, \shape$) parameters and clothing classes ($\clothclass$) of the scan. We use SMPL registration library~\cite{bhatnagar2020ipnet,lazova2019360} to obtain SMPL parameters. For clothing class labels, we render a single viewpoint of the scan and infer the clothing classes using a \sota{} human parsing network~\cite{u2net}.

\subsubsection{\pcencoder{}}
\label{sec:point}

Semantic/part segmentation of a point cloud needs meaningful local and global geometric features~\cite{3d_segmentation_survey}. Following this, we implement our Point Encoder, $\pointenc$ using \sota{} \emph{EdgeConv} based architecture, called DGCNN~\cite{dgcnn}. DGCNN operates on a point cloud by constructing a directed graph $\cG = (\cV, \cE)$, where $\cV \in \{ 1,\hdots n \}$ and $\cE \subseteq \cV \times \cV$. The edges $\cE$ are obtained using $k$-nearest neighbors of $\vect{p}_i \in \mathbb{R}^{F}$, where $F=9$ for first layer, and $64$ for subsequent layers. We then calculate per-point features using Edge Convolution given in~\eqnref{eq:dgcnn}, where $h_{\theta}$ is the learnable edge feature layer.

\vspace{-10pt}

\begin{equation}
\label{eq:dgcnn}
\vect{p'}_i = \max_{j: (i, j) \in \cE}  \; h_{\theta} (\vect{p}_i, \vect{p}_j) \; 
\end{equation}

Similar to~\cite{dgcnn}, we use 3 EdgeConv layers followed by an MLP ($ f_{\mathrm{MLP}}$) to learn a global encoding. This results in multi-scale per-point features given by $\pointfeat_i = \{ \vect{p'}^s_i | \vect{p'}_{\mathrm{g}} \}$, where  $s =\{0,1,2\}$, $\vect{p'}^s_i \in \nR^{l}$ is per-point feature learned by $s^{\mathrm{th}}$ EdgeConv layer and $ \vect{p'}_{\mathrm{g}} \in \nR^{1024}$ is a global encoding of the point cloud. $\vect{p'}_{\mathrm{g}}$ is obtained using an MLP, $ \vect{p'}_{\mathrm{g}} = f_{\mathrm{MLP}}(\mat{P'})$, where 
$\mat{P'} = (\vect{p'}_0, \dots, \vect{p'}_n) \in \mathbb{R}^{n \times (l+l+l)}, \vect{p'}_i = \{\vect{p'}^0_i | \vect{p'}^1_i | \vect{p'}^2_i\}$ is the concatenation of intermediate per-point features.

\subsubsection{\bodyencoder}
\label{sec:canon}

 We incorporate the correlation between body parts and clothing class using SMPL mesh, given by $\smpl(\shape,\pose)$ and SMPL template mesh $\template$. In particular, for every point $\vect{x}_i$ in the input point cloud, we find the index ($j$) of the nearest vertex on SMPL mesh, given by $\smpl_j(\shape,\pose)$. We then find the corresponding vertex location in SMPL template $\template$. In this way, we associate each point in the point cloud with fine-grained semantic information about the human body. This module is not learnable and only requires a nearest-neighbor search in $\nR^{n \times 3}$. We represent the encoded body feature as $ \bodyfeat_i$ = $\template_j$ and call this a Canonical Body Encoder.

\subsubsection{Clothing Encoding and Class-based Attention}
\label{sec:attention}

Clothing classification is ambiguous due to its subjective nature (\eg,  think about the difference between jackets and coats). To address this, we use a learnable codebook $\mat{G} \in \nR^{ \nclasses \times l}$. Our model learns distinct latent vectors~\cite{vqvae} for each clothing class in an auto-decoding manner~\cite{park2019deepsdf}. This enables the model to acquire per-clothing attributes and relevant characteristics for segmentation. The learned codebook is fixed during inference. Compared to a non-learnable binary/one-hot encoding-based representation, the learnable codebook yields better performance in challenging regions, like clothing boundaries and uncommon clothing items like jumpsuits(see~\secref{sec:ablation}).
The key idea of our model is learning an explicit association between per-point features and clothing class. We implement this using an attention module, where we compute how much each point feature attends to a clothing feature. The attention between a point feature and a specific clothing latent code increases when that particular clothing is present at the query point. \\
We define query vector using per-point EdgeConv features $\vect{p'}^s_i,$ $s$ = $2$, and key-value pair using the learnable codebook $\mat{G}$ and define the attention mechanism in \eqnref{eq:attention}, where $\circ$ is masking operator. Simply using  $\vect{p'}^s_i \times \mat{G} $, would also yield features for clothing items not present in the scan. This might result in learning spurious correlations between features and labels. To avoid this, we mask the key matrix $\mat{G}$, using a binary encoding $\vect{g}$ of length $K=\nclasses$, where $g_{j} = 1$, if $j^{\mathrm{th}}$ class is present in the scan, and $g_{j} = 0$ otherwise. Our model is steered towards learning the clothing-specific prior, leveraging fine-grained local point features and clothing latent codes.

\begin{equation}
\label{eq:attention}
\clothfeat_i = \mathrm{softmax} (\vect{p'}^s_i \times (\vect{g} \circ \mat{G} )^{T}) \mat{G}.
\end{equation}

\subsubsection{Segmentation Decoder}
\label{sec:decoder}

Finally, we concatenate all the features $\allfeat_i = \{ \pointfeat_i | \bodyfeat_i | \clothfeat_i \} $, and pass it through a segmentation decoder $\segdecoder$, which predicts per-point segmentation labels. The network $\segdecoder$  is implemented as an MLP.

\noindent{\bf Loss.} We train \clothmodel{} with the cross-entropy loss in \eqnref{eq:loss}, where $\hat{y}^k_i$ is the true label, $y^k_i$ is the predicted probability of the $k^{\mathrm{th}}$ class, and $K$ is the number of classes.

\begin{equation}
\label{eq:loss}
\cL_{\mathrm{CE}} = -\sum_{k=1}^{K} \hat{y}^k_i \log({y}_i^k)
\end{equation}

\subsection{\clothtool}
\label{sec:tool}

In the previous section, we presented our method which learns clothing prior from data for improved generalization over garment styles, appearance and categories. However, we foresee that a variety of clothes often shows unique characteristics that are difficult to catch statistically, especially with the datasets available at the present date. For this reason, we introduce \clothtool{}, a fast-interactive tool to streamline the label refinement process. It provides a graphical interface built explicitly for clothing segmentation and relies on Open3D library~\cite{open3d}, offering a broad set of functionalities (\eg, points selection, labels updating, segmentation prediction, model refinement). We use \clothtool{} to refine the training dataset, and to improve our network generalization on publicly available dataset by backpropagating user feedback in a continual learning setup~\cite{continualsurvery}.

\noindent{\bf Continual Learning \clothmodel{} Refinement.} Let $y^k_i$ be the predicted probability of the $k^{\mathrm{th}}$ class label of $i^{\mathrm{th}}$ point in given pointcloud, and  $\hat{y}^k_i$ be the correct segmentation label provided by user using \clothtool{}. We define the loss as:
\begin{equation}
\label{eq:refine}
\cL_{\mathrm{refine}} = \lambda_{\mathrm{c}} \underset{i \in \cC}{\cL_{\mathrm{CE}}} (y_i^k,\hat{y}_i^k) + \lambda_{\mathrm{f}}   \underset{{i \in \cF}}{\cL_{\mathrm{CE}}} (y_i^k,\hat{y}_i^k) + \lambda_{\mathrm{w}}  {\cL_{\mathrm{W}}} (\theta, \theta'),
\end{equation}
where $\cC$ is the set of indices of point cloud corrected by the user and  $\cF$ is the set of remaining points. $\cL_{\mathrm{CE}}$ is cross-entropy loss defined in~\eqnref{eq:loss}, $\cL_{\mathrm{W}}$ is weight regularization, penalizing weights difference between refined model ($\theta'$) and original pre-trained model ($\theta$). $\lambda_{\mathrm{c}}, \lambda_{\mathrm{f}}, \lambda_{\mathrm{w}}$ are the weights associated with each loss term. We only fine-tune the last layer of the segmentation decoder and MLP of the \pcencoder{}. Following~\cite{8107520}, we use $\lambda_{\mathrm{c}} \ll \lambda_{\mathrm{f}}$, in order to avoid catastrophic forgetting.

%% file: sec/05_Results.tex
\section{Experiments and Results}
\label{sec:exp}

We describe the experiment setup in \secref{sec:setup}, evaluate our proposed \clothmodel{} and compare it with \sota{} part segmentation in~\secref{sec:results}, and with prior methods in~\secref{sec:prior}. We analyze each module of \clothmodel{} in~\secref{sec:ablation} and discuss the attention-based clothing prior in~\secref{sec:garmentprior}. Additionally, we present results on publicly available clothing datasets in~\secref{sec:realworld} and showcase improvements in generalization through \clothtool{}-based refinement.

\subsection{Experimental Setup}
\label{sec:setup}
\noindent{\bf Implementation Details:}
We use the official DGCNN implementation~\cite{dgcnn} with 3 EdgeConv layers (feature-length $l=64$ and $|\vect{p'}_\mathrm{global}| =1024$). For the clothing codebook($\mat{G}$), we use $l=64$. The clothing-class-based attention module is based on multi-head attention. The train-val-test splits of \clothdata{} are 2652/265/270.

\noindent{\bf Error Metric.} Intersection over Union (IoU) is a popular metric for segmentation, quantifying the overlap between predicted and ground truth labels. We consider both per-class IoU and mean IoU ($\mathrm{IoU_{mean}}$) over all classes.

\subsection{3D Clothing Segmentation}
\label{sec:results}

\begin{figure}[t]
	\centering
    \begin{overpic}[width=0.47\textwidth,unit=1bp,,tics=10]{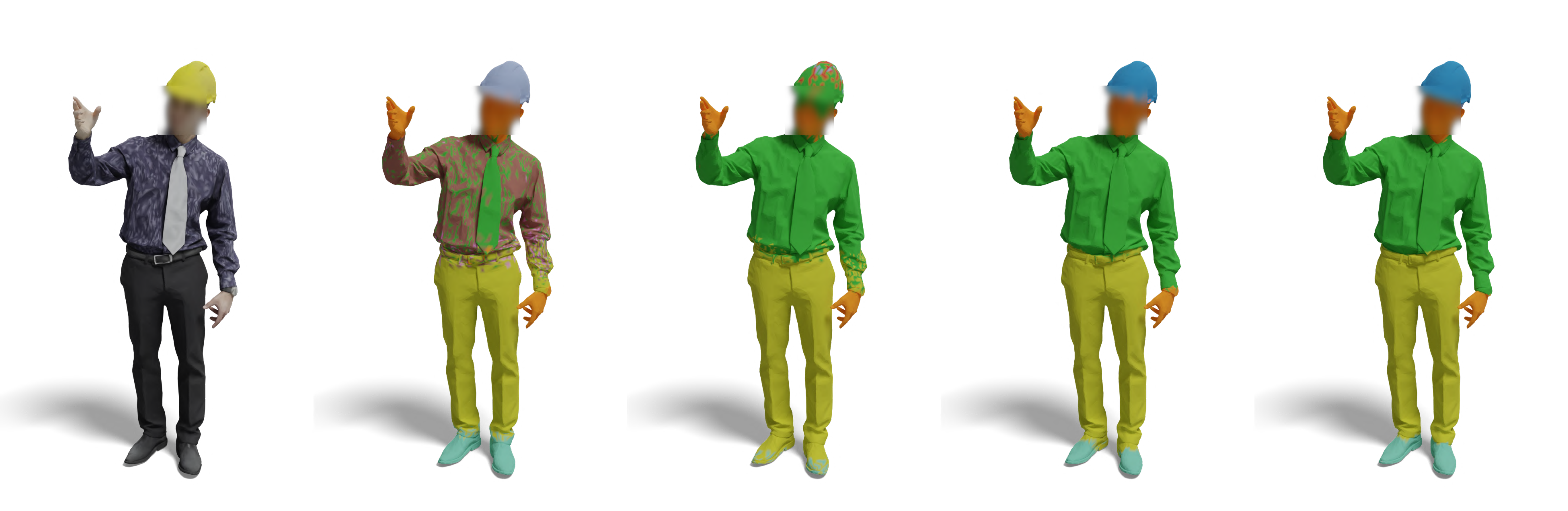}
     		\put(15,75){\colorbox{white}{\parbox{0.15\textwidth}{%
     \scriptsize{Input}}}}
       	    \put(50,75){\colorbox{white}{\parbox{0.15\textwidth}{%
     \scriptsize{DGCNN~\cite{dgcnn}}}}}
          		\put(100,75){\colorbox{white}{\parbox{0.15\textwidth}{%
     \scriptsize{DeltaConv~\cite{Wiersma2022DeltaConv}}}}}
       	    \put(155,75){\colorbox{white}{\parbox{0.15\textwidth}{%
     \scriptsize{\textbf{Ours}}}}}
       	    \put(205,75){\colorbox{white}{\parbox{0.15\textwidth}{%
     \scriptsize{GT}}}}
    \end{overpic}
\leavevmode\newline
	\includegraphics[width=0.47\textwidth]{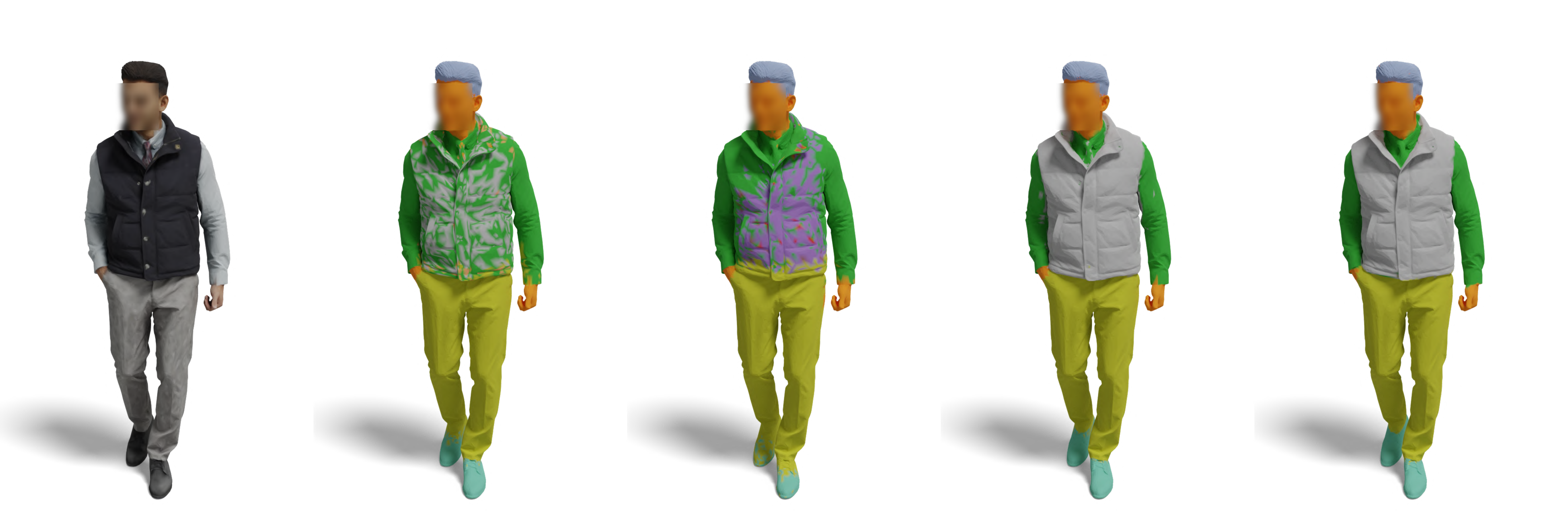}
\leavevmode\newline
	\includegraphics[width=0.47\textwidth]{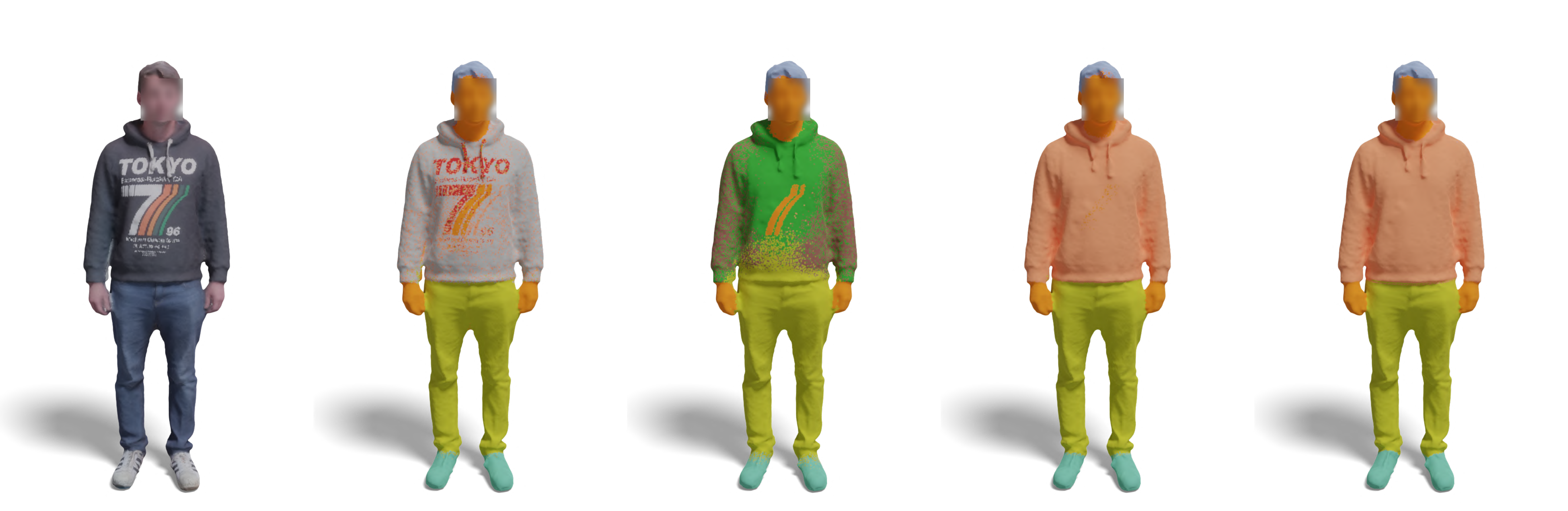}
   \caption{Comparison with \sota{} part segmentaiton models:\textbf{ DGCNN~\cite{dgcnn}} and \textbf{DeltaConv~\cite{Wiersma2022DeltaConv}}. Our model  predicts accurate clothing classes and finer boundaries in complex scans. This can be attributed to our model's utilization of local point features, body priors, and clothing class-based attention features.}
	\label{fig:baseline}
 \vspace{-16pt}
\end{figure}

We evaluate \clothmodel{} on test split of \clothdata{}, both qualitatively (\figref{fig:baseline}) and quantitatively (\tabref{tab:baseline}). Given the similarity between clothing segmentation and part segmentation tasks~\cite{gim3d}, we compare our method with \sota{} 3D part segmentation models trained on \clothdata{}. For this, we employ methods~\cite{3d_segmentation_survey} like DGCNN~\cite{dgcnn} and DeltaConv~\cite{Wiersma2022DeltaConv}. As observed in \figref{fig:baseline}-middle, DGCNN and DeltaConv struggle with multi-layer clothing. This arises from the absence of clothing-related information in the model and its inclination to overfit to single-layer scans, which constitute the majority of the dataset. On the other hand, our model incorporates clothing information via the learned distinct codebook and hence predicts correct labels with crisp boundaries. Moreover, as seen in \figref{fig:baseline}-top both baseline methods inadequately segment less common classes, such as hats.  We also observe texture bias in both baseline methods (\figref{fig:baseline}-bottom); both networks fail to predict the same class for nearby points with different texture colors. We attribute this to the codebook and masked attention module in our model, which enables the network to learn distinct features for each clothing and avoid learning spurious correlations between point features and absent classes.

\begin{table*}[t]
\centering

\resizebox{\textwidth}{!}{
\begin{tabular}{lccccccccccccccccccc}
\toprule
Method & Mean  & T-shirt & Shirt  &  Vest & Coat & Jacket & Hoodies & Short-Pants & Pants  & Skirts & Dress &  JumpS. & SwimS. & UnderG. & Scarf & Hat & Shoes & Body & Hair  \\

\midrule
DGCNN~\cite{dgcnn}  & 87.11 &87.88 & 81.02 & 90.58 & 81.60 & 96.16 & 97.46 & 94.60 & 82.50 & 96.44 & 73.61 & 76.67 & 98.89 & 99.26 & 73.33 & 95.19 & 79.47 & 80.45 & 82.83 \\ 
DeltaConv~\cite{Wiersma2022DeltaConv} & 84.78 & 87.22 & 73.56 & 84.68 & 80.06 & 98.52 & 96.99 & 89.58 & 78.37 & 94.11 & 67.08 & 77.04 & 99.26 & 99.26 & 73.33 & 95.19 & 72.98 & 76.69 & 82.13 \\
Ours & \textbf{91.23} & \textbf{95.47} &  \textbf{92.94} &  \textbf{98.86} &  \textbf{90.12} &  \textbf{99.23} &  \textbf{99.43} &  \textbf{98.32} &  \textbf{85.96} &  \textbf{98.12} &  \textbf{79.11} &  \textbf{77.73} &  \textbf{99.78} &  \textbf{99.96} &  \textbf{73.23} &  \textbf{97.72} &  \textbf{82.96} &  \textbf{85.76} &  \textbf{87.49} \\

\bottomrule
\end{tabular}

}
\vspace{-0.2cm}

\caption{We quantitatively compare the results of our method \sota{} part segmentation methods, DGCNN~\cite{dgcnn} and DeltaConv~\cite{Wiersma2022DeltaConv}. We report IoU for every class and mean over all the classes($\mathrm{IoU_{mean}}$).}
\label{tab:baseline}
\end{table*}

\subsubsection{Comparison with Prior Work}
\label{sec:prior}

We compare \clothmodel{} with prior 3D clothing segmentation methods like \mgn{} and GIM3D~\cite{gim3d}. Since both GIM3D and \mgn{} predict 3 classes, we merge the predictions of our model into the respective classes. For GIM3D, we only compare with PointNet++~\cite{pointnet2}, as it is the only one authors provided to us. We observe from~\tabref{tab:prior} that \clothmodel{} largely improves over prior work. \\
Moreover, \clothmodel{} takes approximately $5-6 \; \mathrm{seconds}$ to infer the segmentation labels for the whole scan($270k$ vertices) on 12 GiB GPU (3080Ti). Whereas for~\mgn{} it takes roughly $\sim20$ minutes to get the segmentation labels for SMPL+D mesh ($27k$ vertices). 

\begin{figure}[t]
\vspace{-0.1cm}
    \centering
        \begin{overpic}[width=0.47\textwidth,unit=1bp,tics=20]{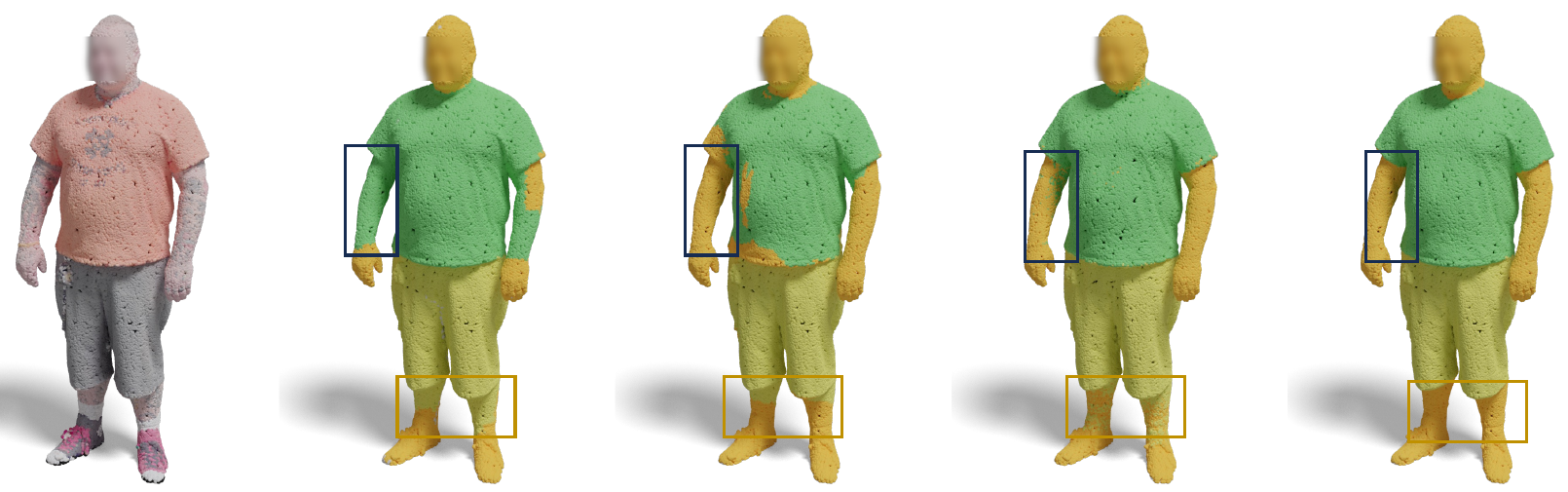}
     		\put(10,75){\colorbox{white}{\parbox{0.15\textwidth}{%
     \scriptsize{Input}}}}
       	    \put(50,75){\colorbox{white}{\parbox{0.15\textwidth}{%
     \scriptsize{\mgn}}}}
          		\put(100,75){\colorbox{white}{\parbox{0.15\textwidth}{%
     \scriptsize{GIM3D~\cite{gim3d}}}}}
       	    \put(155,75){\colorbox{white}{\parbox{0.15\textwidth}{%
     \scriptsize{\textbf{Ours}}}}}
       	    \put(210,75){\colorbox{white}{\parbox{0.15\textwidth}{%
     \scriptsize{GT}}}}
    \end{overpic}

	\caption{Comparison with  \textbf{\mgn{}} and \textbf{GIM3D~\cite{gim3d}}. } 
	\label{fig:results_prior}
\end{figure}

We further provide qualitative results in \figref{fig:results_prior}. We observe that \mgn{} is not able to generate precise labels at boundaries, \eg at the leg. This is because the prior designed for lower garments comes from a fixed template of ``long pants", whereas the pant in this scan is smaller than the pre-defined template. We also observe that sometimes the handcrafted features are not able to correct segmentation errors due to texture bias and inconsistent multiview prediction of 2D human parsing~\cite{pgn,u2net}. This is visible in the right hand of the scan. We notice that GIM3D fails near boundaries. This stems from the fact that normal information is not sufficient to distinguish between different clothing, especially in the case of real-world scans, where normals can be noisy. On the other hand, our model results in accurate boundaries as it takes texture information into account as well and doesn't rely on any pre-defined prior.

\begin{table}[t]
\centering

\resizebox{0.4\textwidth}{!}{
\begin{tabular}{lcccccc}
\toprule
Dataset &  MGN~\cite{bhatnagar2019mgn} &GIM3D~\cite{gim3d}  & Ours \\

\midrule
\clothdata{}-Test  &   88.88 & 72.04 &  \textbf{92.47}  \\

\bottomrule
\end{tabular}

}

\caption{\label{tab:prior} Comparison with MGN~\cite{bhatnagar2019mgn} and GIM3D~\cite{gim3d} on three class(upper, lower and body) segmentation.}

\vspace{-0.5cm}
\end{table}

\subsubsection{Learned Clothing Prior}
\label{sec:garmentprior}

\begin{figure}[t]
\vspace{-0.5cm}
\centering
    \begin{overpic}[width=0.4\textwidth,unit=1bp,tics=10]{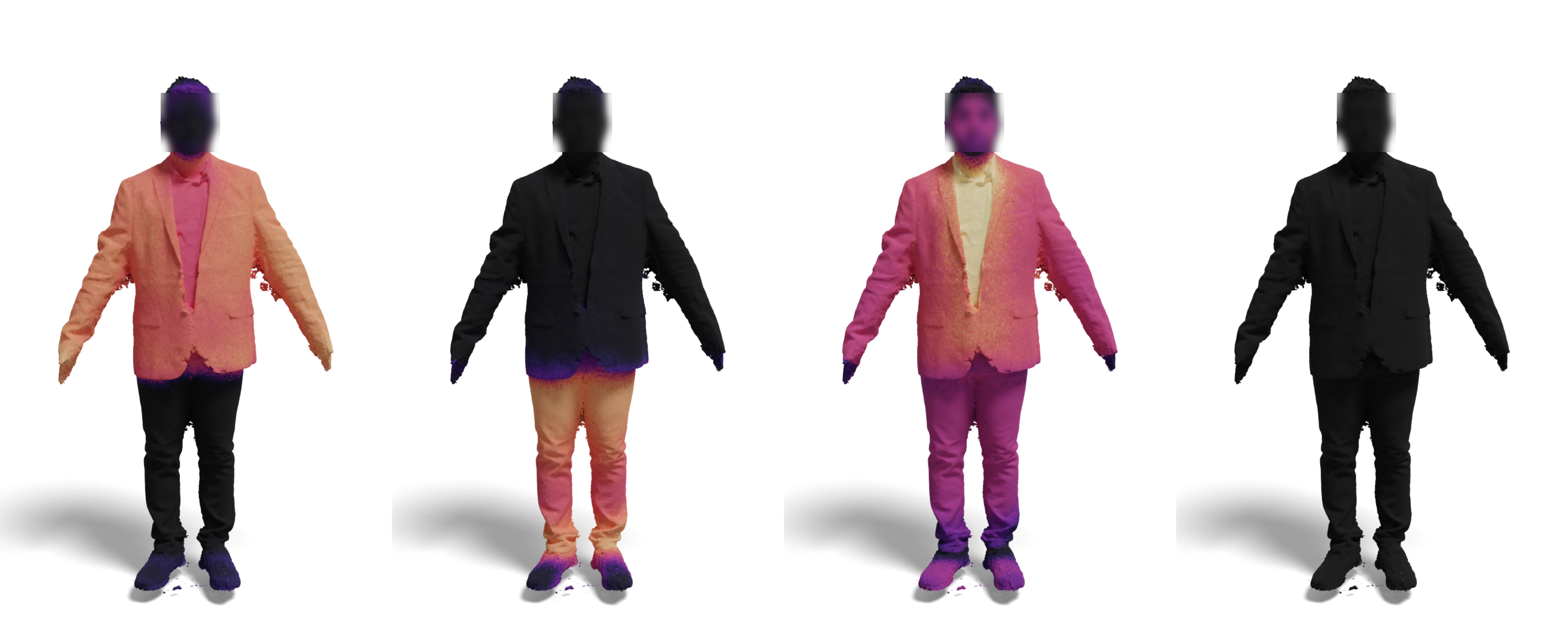}
     		\put(13,75){\colorbox{white}{\parbox{0.15\textwidth}{%
     \scriptsize{Coat}}}}
       	    \put(63,75){\colorbox{white}{\parbox{0.15\textwidth}{%
     \scriptsize{Pants}}}}
          		\put(110,75){\colorbox{white}{\parbox{0.15\textwidth}{%
     \scriptsize{Shirt}}}}

       	    \put(160,75){\colorbox{white}{\parbox{0.15\textwidth}{%
     \scriptsize{Hoodies}}}}
    \end{overpic}
\leavevmode\newline

\begin{overpic}
[width=0.4\textwidth,unit=1bp,tics=10]{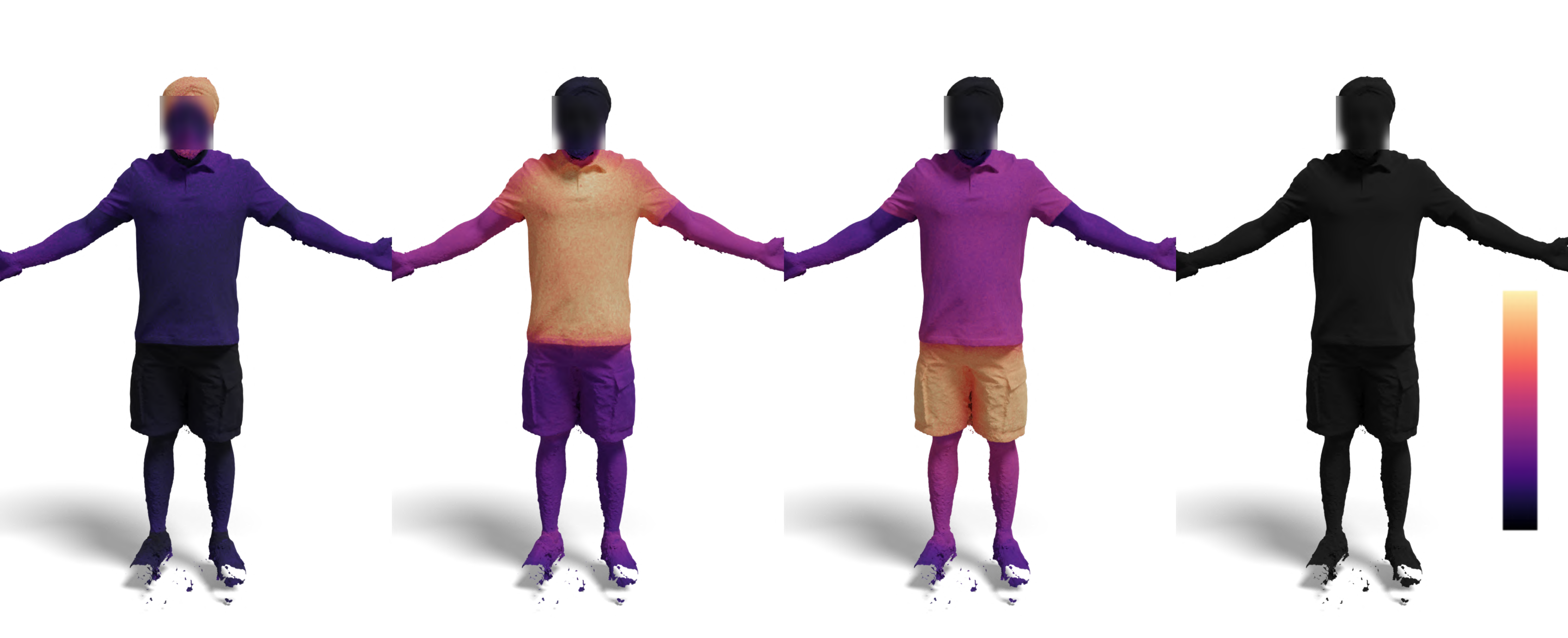}
\scriptsize
     		\put(15,76){\colorbox{white}{\parbox{0.15\textwidth}{%
     \scriptsize{Hat}}}}
       	    \put(60,76){\colorbox{white}{\parbox{0.15\textwidth}{%
     \scriptsize{TShirt}}}}
          		\put(109,76){\colorbox{white}{\parbox{0.15\textwidth}{%
     \scriptsize{Short Pants}}}}

       	    \put(157,76){\colorbox{white}{\parbox{0.15\textwidth}{%
     \scriptsize{Hoodies}}}}

     \put(195,12){0}
     \put(195,40){1}
    \end{overpic}
	\caption{\textbf{Clothing prior learned using attention module}: Attention module in \garencoder{}, learns a robust clothing prior based on point features. Here we visualise the attention of point feature on different clothing class.}
	\label{fig:garment_prior}

 \vspace{-0.5cm}
\end{figure}

A good clothing prior is crucial for segmentation. In \mgn{}, a geodesic distance-based prior was manually crafted. This approach lacks scalability for new clothing and struggles with varying shapes within the same class (e.g., different jacket lengths or shirt sleeve styles \etc). Interestingly, our model's attention-based clothing encoder learns clothing prior from point features ($\pointfeat$) and a garment codebook ($\clothfeat$). This is visualized in \figref{fig:garment_prior}, where attention for the $j^{\mathrm{th}}$ garment class at point $i$ is calculated as $\vect{p'}^k_i \times \mat{G}_j$. As compared to \mgn{} prior, this is not manually defined and also incorporates the local properties of the clothing.

\vspace{-0.3cm}
\subsubsection{Ablation}
\label{sec:ablation}

\begin{figure}[t]
        \centering
\begin{overpic}[width=0.47\textwidth,unit=1bp,tics=10]{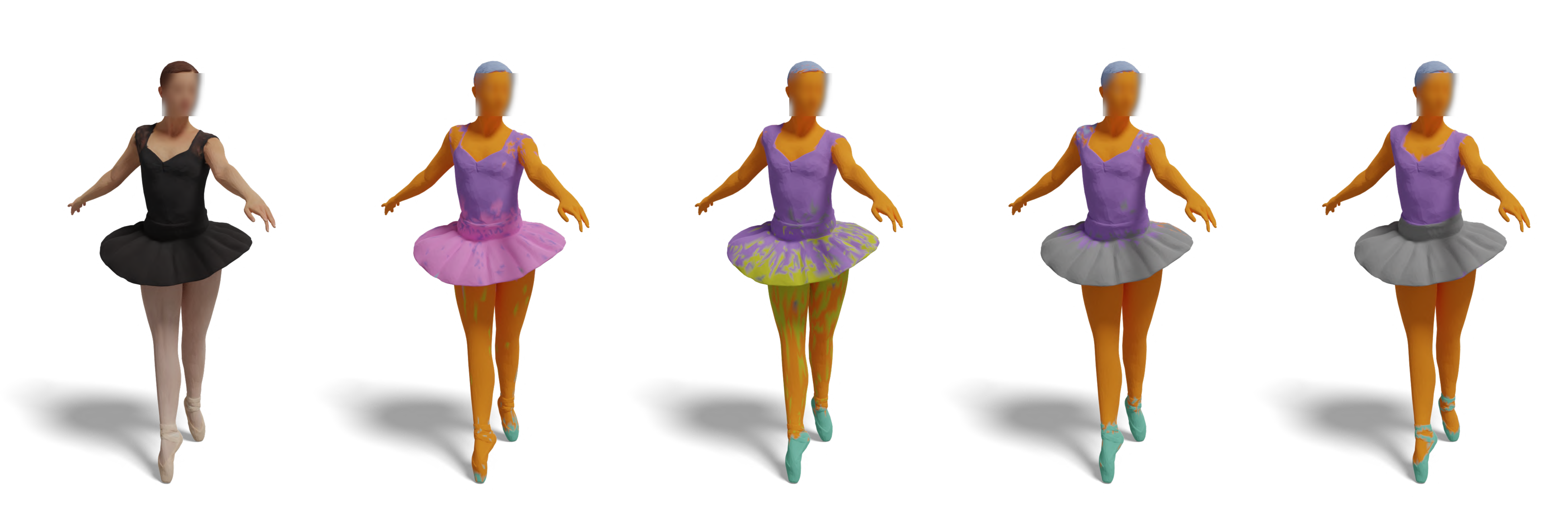}
     		\put(15,75){\colorbox{white}{\parbox{0.15\textwidth}{%
     \scriptsize{Input}}}}
       	    \put(65,75){\colorbox{white}{\parbox{0.15\textwidth}{%
     \scriptsize{$\times$}}}}
          		\put(100,75){\colorbox{white}{\parbox{0.15\textwidth}{%
     \scriptsize{ + Hybrid}}}}
          		\put(150,75){\colorbox{white}{\parbox{0.15\textwidth}{%
     \scriptsize{ + Canonical}}}}
       	    \put(205,75){\colorbox{white}{\parbox{0.15\textwidth}{%
     \scriptsize{GT}}}}
    \end{overpic}
\leavevmode\newline
    \begin{overpic}[width=0.47\textwidth,unit=1bp,tics=20]{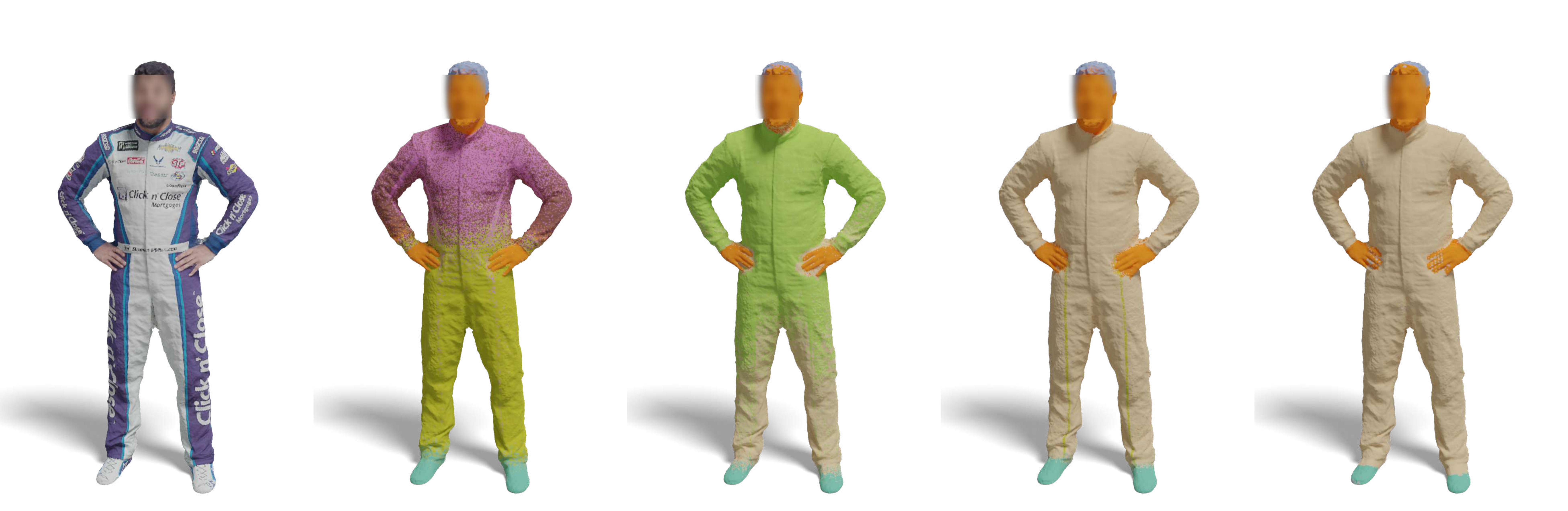}
     		\put(15,75){\colorbox{white}{\parbox{0.15\textwidth}{%
     \scriptsize{Input}}}}
       	    \put(65,75){\colorbox{white}{\parbox{0.15\textwidth}{%
     \scriptsize{$\times$}}}}
          		\put(100,75){\colorbox{white}{\parbox{0.15\textwidth}{%
     \scriptsize{ + Binary}}}}
          		\put(150,75){\colorbox{white}{\parbox{0.15\textwidth}{%
     \scriptsize{ + Attention}}}}
       	    \put(205,75){\colorbox{white}{\parbox{0.15\textwidth}{%
     \scriptsize{GT}}}}

     \vspace{-1cm}
    \end{overpic}

	\caption{\textbf{\bodyencoder}(Top): As opposed to others, the proposed \bodyencoder (Canonical) is simple, generalizes to difficult poses, and produces fine boundaries. \textbf{\garencoder}(Bottom): Attention-based encoder and codebook learn distinct garment features and prior, achieving accurate segmentation prediction. }
	\label{fig:results_ablation}

\end{figure}

We experimentally validate and discuss the design choices of each module of \clothmodel{} in \tabref{tab:ablation} and \figref{fig:results_ablation}.

\noindent{\bf \pcencoder.} We experiment with two part segmentation models: DGCNN~\cite{dgcnn} and DeltaConv~\cite{Wiersma2022DeltaConv} as \pcencoder. ~\tabref{tab:ablation} shows the DGCNN-based model significantly outperforms the DeltaConv-based one. This behavior is similar to the standalone DGCNN and DeltaConv, as shown in \tabref{tab:baseline}. DGCNN surpasses DeltaConv by learning feature spaces where semantically similar features are closely clustered. Hence, we use DGCNN as \pcencoder{} in our model.

\noindent{\bf \bodyencoder.} We explore different body encodings: \emph{Canonical Body Encoder}, $\bodyfeat$ (\secref{sec:canon}), and a fusion of $\bodyfeat$ with a coarse feature encoder based on COAP~\cite{coap}, resulting in the \emph{Hybrid \bodyencoder{}}. The lack of \bodyencoder{} in the model leads to mislabeled regions in improbable locations, such as a t-shirt label appearing in the skirt region. These mislabels are evident as patches in~\figref{fig:results_ablation}-top. The \emph{Hybrid \bodyencoder{}} learns vertex and body part associations with the clothing, but results in smudged boundaries(see skirt and legs in \figref{fig:results_ablation}-top ). This is because the hybrid model contains body-part features, so it tends to associate the same labels to all the points in a body part if there is no significant difference in geometry or appearance. Our proposed \bodyencoder{} uses fine-grained correspondence and establishes accurate correlations between clothing labels and body locations.

\noindent{\bf \garencoder.} We compare our attention-based clothing encoder, with a binary encoding-based one. In binary encoding, we use $\vect{g}$, instead of $\clothfeat$. The attention-based model boosts performance and also learns garments prior from data. Binary encoding is not consistently effective in predicting the correct garment, particularly when confronted with uncommon clothing styles, as exemplified by the jumpsuit case in \figref{fig:results_ablation}-bottom, similar to models without \garencoder{}. Moreover, without any \garencoder{}, models exhibit texture bias. Leveraging the learned clothing prior(\secref{sec:garmentprior}) significantly improves performance, and alleviates the mentioned problems.

  \begin{table}[t]
  \vspace{-0.5cm}
    \setlength{\tabcolsep}{4pt}
    \centering
    \footnotesize
    \begin{tabular}{cccc}
    \toprule
    \textbf{\multirow{ 2}{*}{\shortstack{Points \\ Encoder}}} & \textbf{\multirow{ 2}{*}{\shortstack{\bodyencoder{}}}} & \textbf{\multirow{ 2}{*}{\shortstack{Clothing \\ Encoder}}} &{\textbf{$\mathrm{IoU_{mean}}$ $\uparrow$}} \\ 
\\
    \midrule
    \multirow{7}{*}{\textbf{DGCNN}} & \textbf{Canonical} & \textbf{Attention} & \textbf{91.23}  \\
    & Canonical & Binary & 90.41 \\
    & Hybrid & Attention & 89.70 \\ %
    & Hybrid & Binary & 89.68 \\
    & $\times$ & Attention & 89.90  \\
    & Canonical & $\times$ & 87.18 \\
    & $\times$ & $\times$  &  87.10  \\

    \midrule
    \multirow{1}{*}{DeltaConv} &Canonical & Attention & 86.84 \\

    \bottomrule
    \end{tabular}
    \caption{Quantitative evaluation of the ablation study on different modules of the proposed \clothmodel{} model. The table shows the performance of the model with different combinations of the Point Encoder, Body Encoder, and Clothing Encoder.}
    \label{tab:ablation}
\end{table}

\subsection{\clothdata{}++}
\label{sec:realworld}

\begin{figure}[t]

    \begin{overpic}[width=0.238\textwidth,unit=1bp,tics=20]{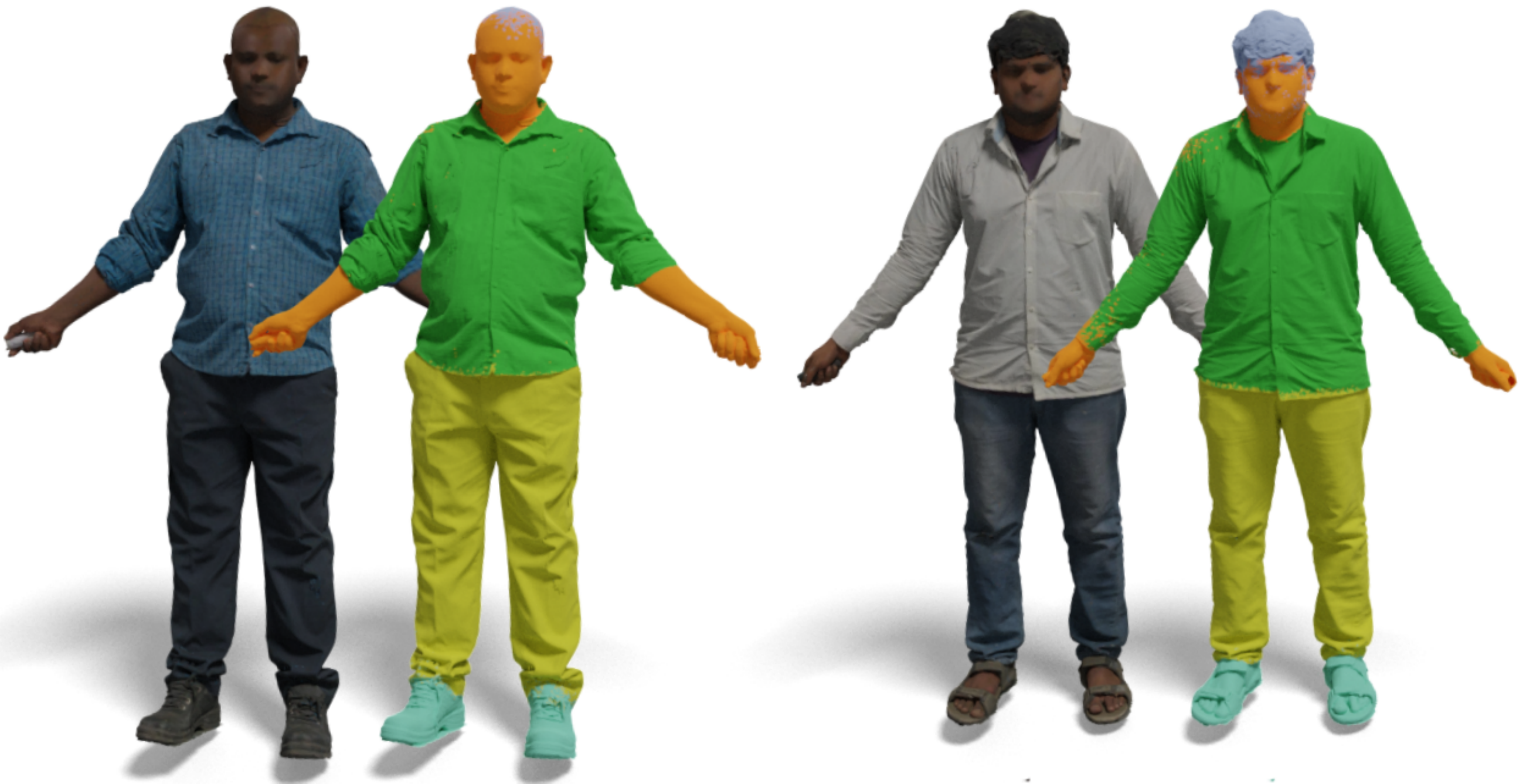}
     		\put(30,65){\colorbox{white}{\parbox{0.15\textwidth}{%
     \scriptsize{3DHumans-IIITH~\cite{Jinka2022}}}}}

    \end{overpic}
    \begin{overpic}[width=0.238\textwidth,unit=1bp,tics=20]{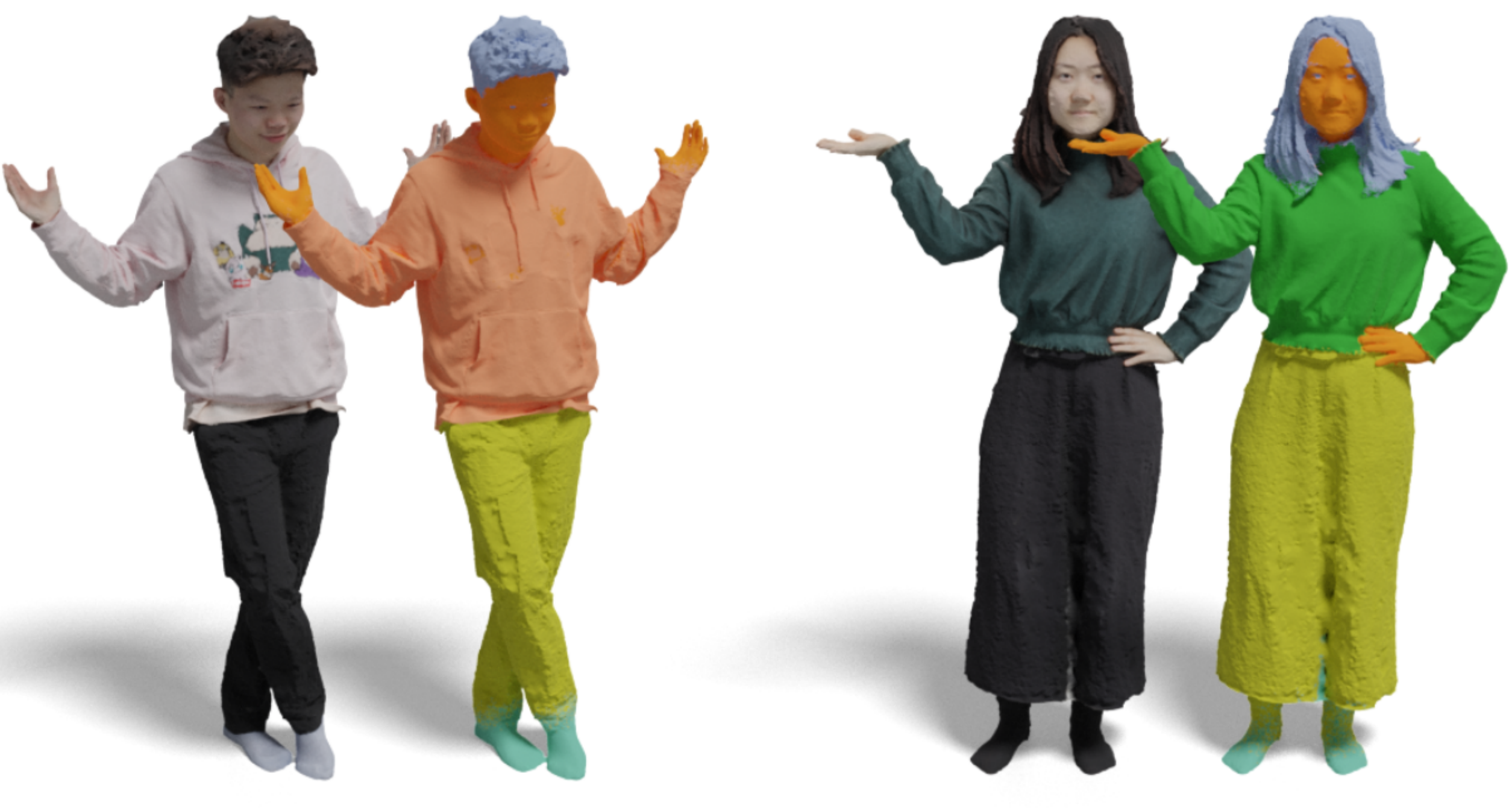}
     		\put(30,65){\colorbox{white}{\parbox{0.15\textwidth}{%
     \scriptsize{THuman2.0~\cite{thuman2}}}}}
    \end{overpic}

	\caption{Results of \clothmodel{} on publicly available datasets~\cite{Jinka2022,thuman2}, showing  generalization capability of the model.}
	\label{fig:results_public}
 \vspace{-0.6cm}
\end{figure}

\begin{figure}[t]
\vspace{-0.5cm}
    \begin{overpic}[width=0.49\textwidth,unit=1bp,tics=20]{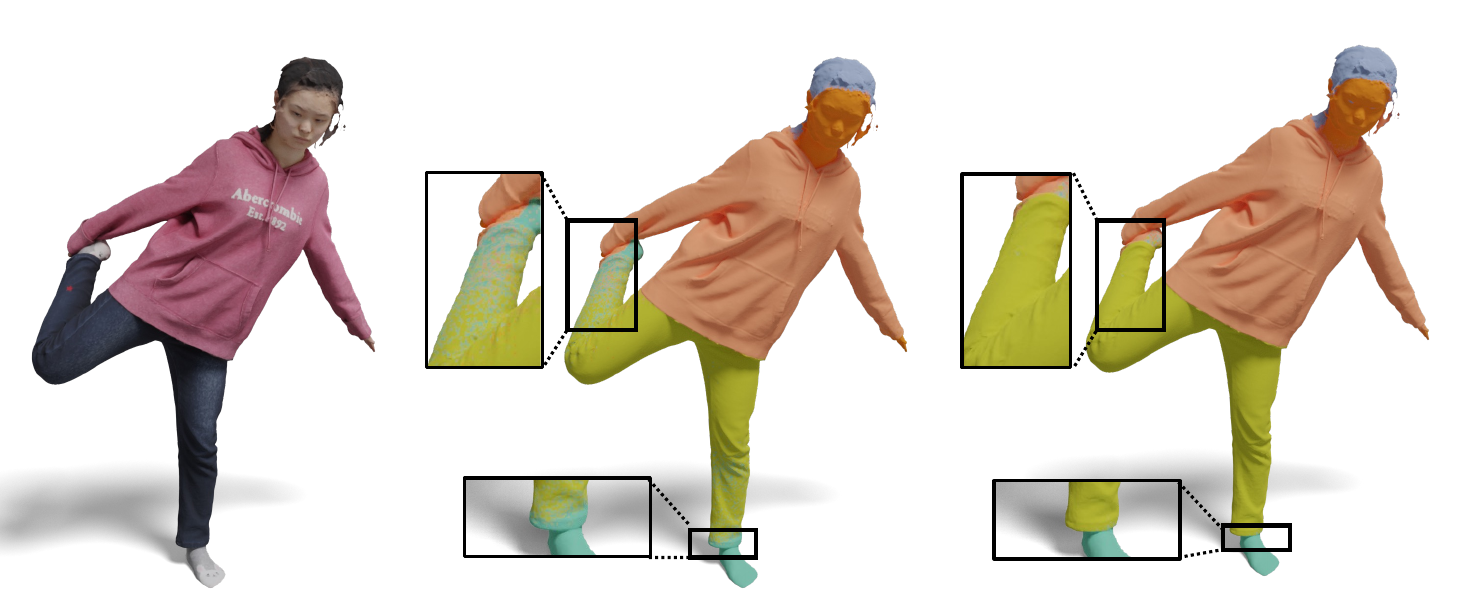}
     		\put(30,95){\colorbox{white}{\parbox{0.15\textwidth}{%
     \scriptsize{Input}}}}

          		\put(80,95){\colorbox{white}{\parbox{0.15\textwidth}{%
     \scriptsize{ Original Network}}}}

       	    \put(160,95){\colorbox{white}{\parbox{0.15\textwidth}{%
     \scriptsize{Refined Network}}}}
    \end{overpic}

	\caption{Improving \clothmodel{} performance on THuman2.0~\cite{thuman2} by fine-tuning the model on few samples from THuman2.0 using \clothtool{}. }
	\label{fig:results_tool}
 \vspace{-0.5cm}
\end{figure}

Our model generalizes well on real-world public datasets, showing good results in~\figref{fig:results_public}. However, it exhibits blurry boundaries and texture bias for some scans, as seen in~\figref{fig:results_tool} (middle). Generalizing to out-of-distribution real-world datasets is challenging due to the vast variability of clothing styles and few differences between classes. 

To address this, we use the proposed \clothtool{} in a continual learning approach(\secref{sec:tool}). The goal is to improve performance over new datasets, without catastrophic forgetting. We show the results of the original model and fine-tuned model on a scan from THuman2~\cite{thuman2} in~\figref{fig:results_tool}. After fine-tuning, results on \clothdata{}-test is a mean IoU of $90.35$, which is a small decrease from the original model($91.23$) and still better than prior work and baselines. We will provide the labels of publicly available datasets such as THuman2,3~\cite{thuman2,thuman3}, CAPE~\cite{cape}(textured-scans), 3DHumans-IIITH~\cite{Jinka2022}, a subset of HuMMan~\cite{cai2022humman} and we call this dataset \clothdata{}++.

%% file: sec/06_Conclusions.tex
\section{Conclusions}
\label{sec:conclusion}
We present a novel fine-grained 3D clothing segmentation model that works directly on point clouds, and to train it we introduce a large-scale dataset of people in diverse clothing items, poses, and with high-quality segmentation labels. We incorporate human body information to improve the pose-generalization of our model and introduce a novel garment class attention module, which learns clothing prior from data, as opposed to hand-crafted priors~\cite{bhatnagar2019mgn}. Our model outperforms prior work and baselines and generalizes to public out-of-distribution datasets. We further introduce a continual learning-based refinement strategy to improve the generalization of the model, without catastrophic forgetting. 

\noindent{\bf Limitations and Future Work}
To the best of our knowledge, this is the first dataset and model for 3D clothing segmentation from colored point clouds, which contains diverse and fine-grained segmentation labels. Future work may broaden our work by adding more clothing items through \clothtool{}, \eg, to include a variety of cultural styles. Our approach necessitates the garment class worn by the subject as network input, requiring a preprocessing step. Potentially, this can be obviated by incorporating clothing prediction directly within the network. Lastly, to enhance network generalization, we have integrated the continual learning~\cite{8107520} framework, paving the way for future exploration of recent strategies such as EWC~\cite{Kirkpatrick_2017}.

%% file: supmat_arxiv.tex
\twocolumn[{%
\renewcommand\twocolumn[1][]{#1}%
\newpage
\null
\vskip .375in
\begin{center}
  {\Large \bf SUPPLEMENTARY MATERIALS \\ \close{}: A 3D Clothing Segmentation Dataset and Model \par}
  \vspace*{24pt}
\end{center}
}]

\input{sec/suppl/01_data}
\input{sec/suppl/02_method}

\input{sec/suppl/03_tool}
\input{sec/suppl/04_results}

%% file: sec/suppl/01_data.tex
\section{Dataset}
\label{supp:dataset}

Our dataset \clothdata{} comes from two sources, 1) \clothdata{}i, which is dataset captured in our lab and 2) \clothdata{}c, dataset from commercial data sources. We explain the dataset capturing details in the following section, followed by the process for obtaining segmentation labels.

\paragraph{\clothdata{}c Data.} We collect scans from different commercial dataset such as AXYZ~\cite{axyz}, Twindom~\cite{twindom}, Treedy~\cite{treedys}, Renderpeople~\cite{renderpeople}. Due to licensing issues, we will not provide the scans from these datasets, but we will release the segmentation labels and detailed instructions to purchase these datasets from respective sources. 

\paragraph{\clothdata{}i Data Capture.} Following data capture setup in~\cite{tiwari20sizer,yenamandra2020i3dmm}, we create a dataset of approximately 100 subjects in 7 diverse poses, wearing 12 garment classes. We use Treedy's scanner~\cite{treedys}, which consists of $\sim 130$ high-resolution camera at a fixed position. We use Metashape~\cite{metashape} for 3D reconstruction, which is photogrammetry-based reconstruction. Reconstructed scans are highly detailed and have high-resolution texture maps associated with them. We also register SMPL~\cite{SMPL:2015} to each scan, with the registration method used in~\cite{bhatnagar2019mgn,tiwari20sizer,lazova2019360}.

\begin{figure}[t]
	\includegraphics[width=0.45\textwidth]{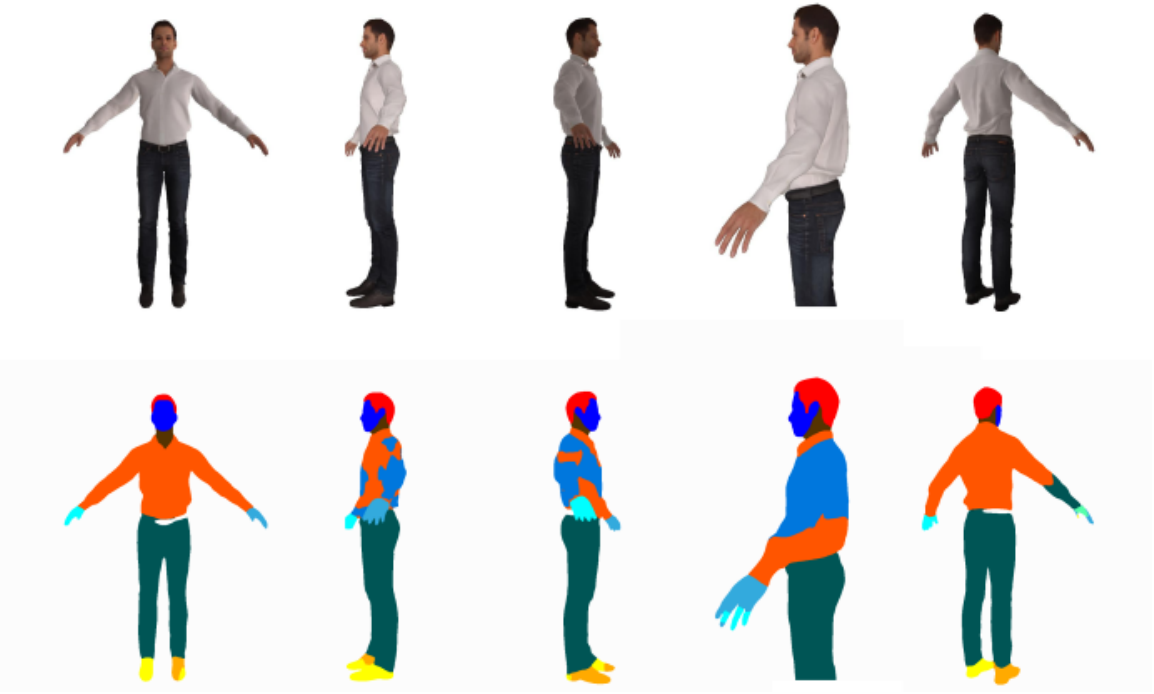}

	\caption{\emph{Top}: Multiview rendered images of a scan.\emph{Bottom}: Clothing segmentation obtained using 2D Parsing method~\cite{pgn}. 2D Parsing method generates inconsistent labels across views. Consequently, when these labels are elevated from 2D to 3D using the 2D-to-3D lifting technique, the resulting segmentation is noisy. }
	\label{fig:pgn_fail}
\end{figure}

\begin{figure}[t]

\centering
    \begin{overpic}[width=0.45\textwidth,unit=1bp,tics=10]{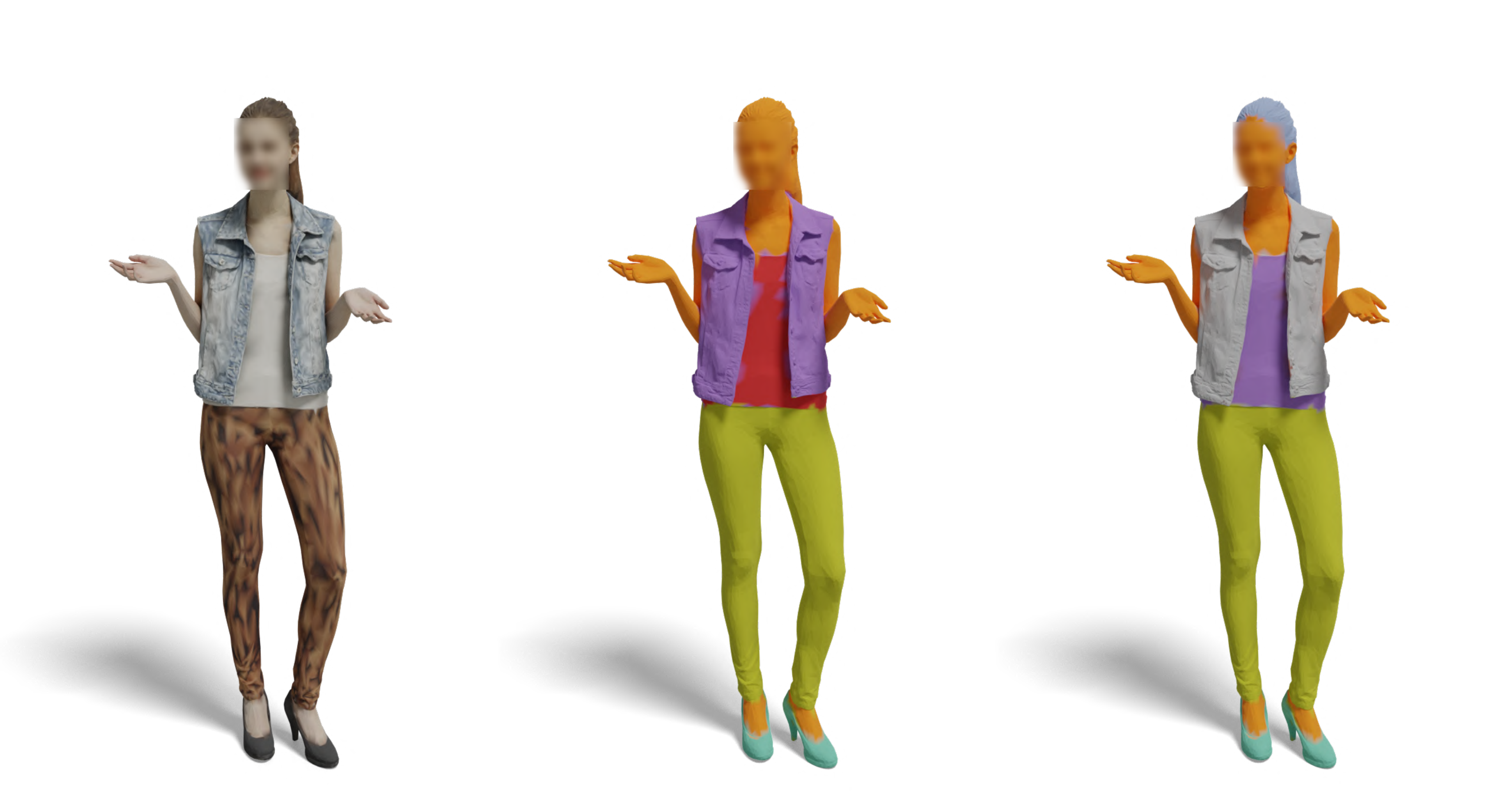}

  \put(40,125){\makebox(0,0){\scriptsize{Textured Scan}}}
  \put(110,125){\makebox(0,0){\scriptsize{Initial Labels}}}
  \put(180,125){\makebox(0,0){\scriptsize{Corrected Labels}}}
  \put(180,118){\makebox(0,0){\scriptsize{(using \clothtool{})}}}
  
    \end{overpic}

 	\includegraphics[width=0.47\textwidth]{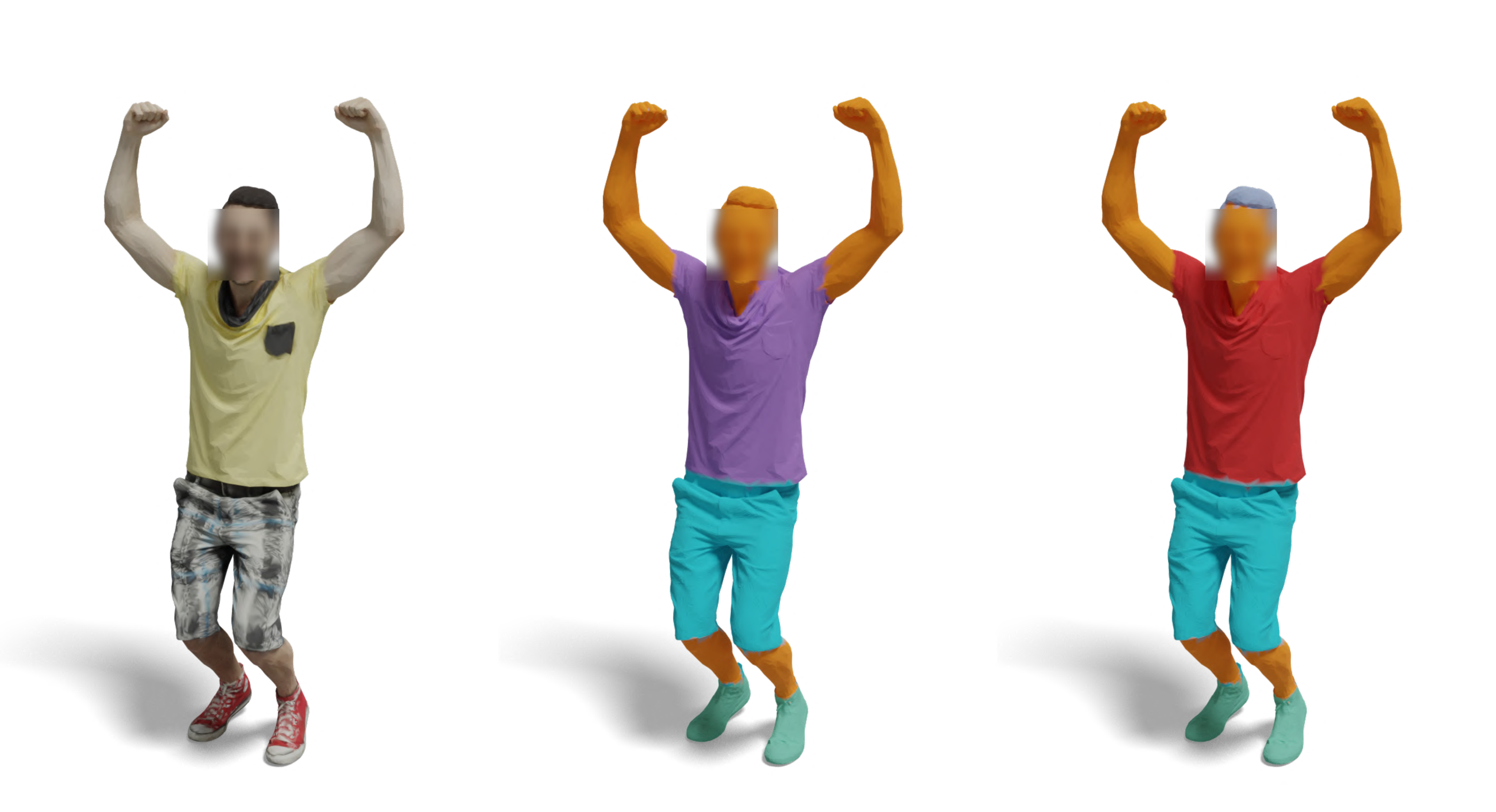}

	\caption{Due inherent uncertainty in clothing classification, the segmentation labels acquired through~\cite{pgn} might be incoherent. However, such labeling discrepancies can be easily corrected using \clothtool{}.}
	\label{fig:dataset2}
\end{figure}

\paragraph{Ground Truth Segmentation Labels of \clothdata{}c Scans.} We follow the pipeline similar to the one in \mgn{}. We first register the scans to SMPL and SMPL+D~\cite{bhatnagar2019mgn}. We then render the registered meshes from 72 different views and apply \sota{} 2D Human Parsing method, PGN~\cite{pgn}. One of the major limitations of such a pipeline is inconsistent multiview prediction of the 2D Human Parsing method, as shown in~\figref{fig:pgn_fail}. This is expected behavior from such methods as 1) they are not trained with any explicit loss to produce multi-view consistent results, and 2) they are not trained on multi-view images of the same scene. As a result, we observe many patches of undesired clothing classes in the 2D segmentation and hence in the lifted 3D segmentation as well. \mgn{} tried to solve this problem by using a pre-defined prior, but these priors are limited to 3 classes. We propose to clean such inconsistency using our hand-crafted heuristics and \clothtool{}(see ~\figref{fig:dataset_tool}(left)). Moreover, PGN labels are inconsistent with our \clothmodel{} labels, so we apply some merging and splitting in labels. We first explain heuristics for merging and segregation of labels in the following points:

\begin{itemize}
    \item \textit{Merging body parts:} In PGN there are separate labels for left-leg, right-leg, left-arm and right-arm. We instead use a single label for all these parts, so we merge them into a single category.
    \item \textit{Separate labels for Upper and Lower Garments :} PGN generates only two kinds of upper garment labels, namely `Shirt' and `Coat'. On the other hand, our model uses more fine-grained labels, \eg `Shirt' is further divided into `TShirt', `Vest', `Hoodies' \etc. We use the \emph{change all} option provided in \clothtool{} to correct such labels, as shown in~\figref{fig:dataset2}. Similarly, there is only one label for lower garments: `Pants', which we split into `Pants' and `Short-Pants'.
\end{itemize}

\begin{figure}[t]

\centering
    \begin{overpic}[width=0.45\textwidth,unit=1bp,tics=10]{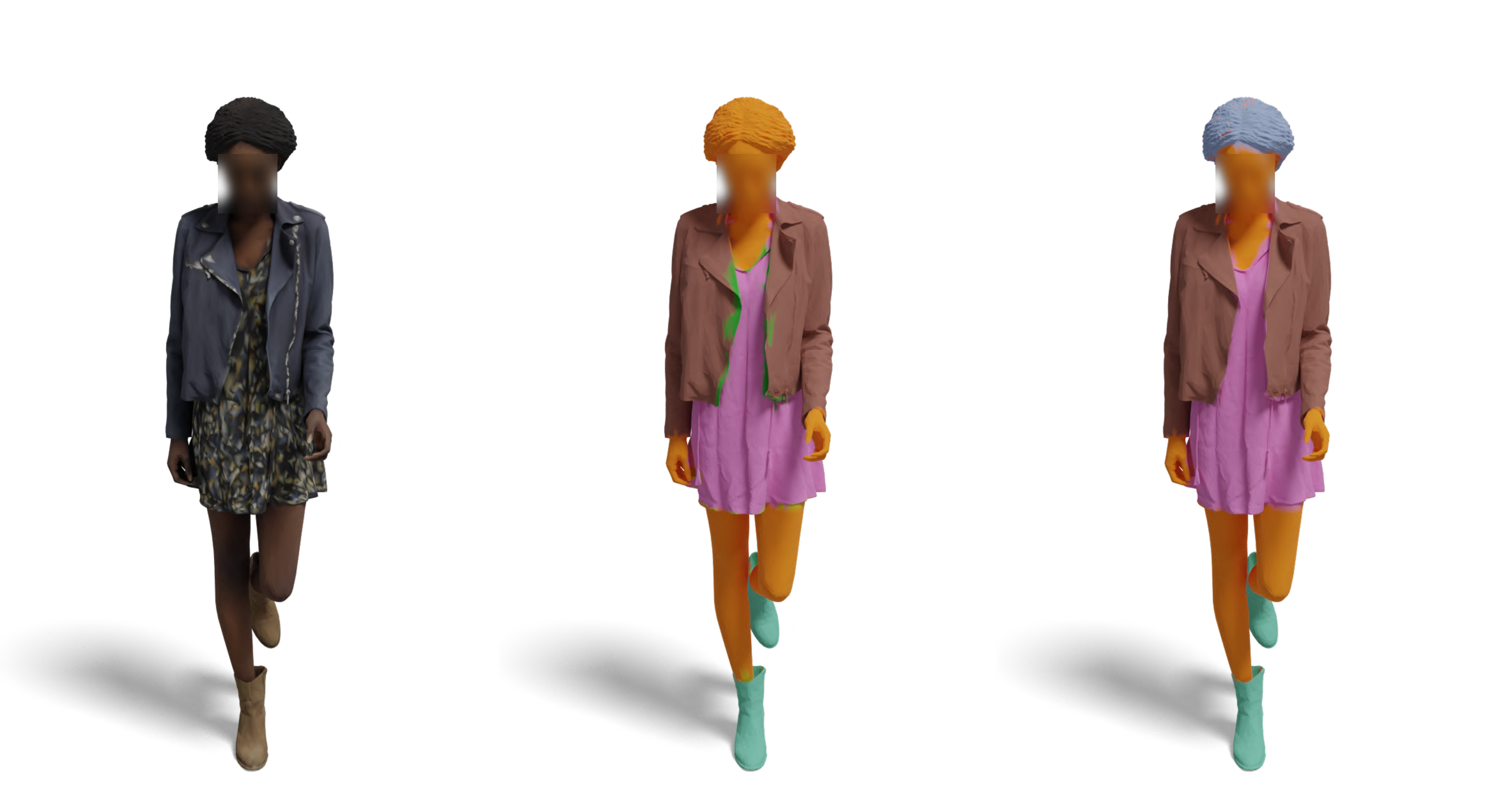}

  \put(35,120){\makebox(0,0){\scriptsize{Textured Scan}}}
  \put(110,120){\makebox(0,0){\scriptsize{Initial Segmentation}}}
  \put(190,120){\makebox(0,0){\scriptsize{Clean Segmentation}}}
  \put(190,112){\makebox(0,0){\scriptsize{(using \clothtool{})}}}
  
    \end{overpic}

 	\includegraphics[width=0.47\textwidth]{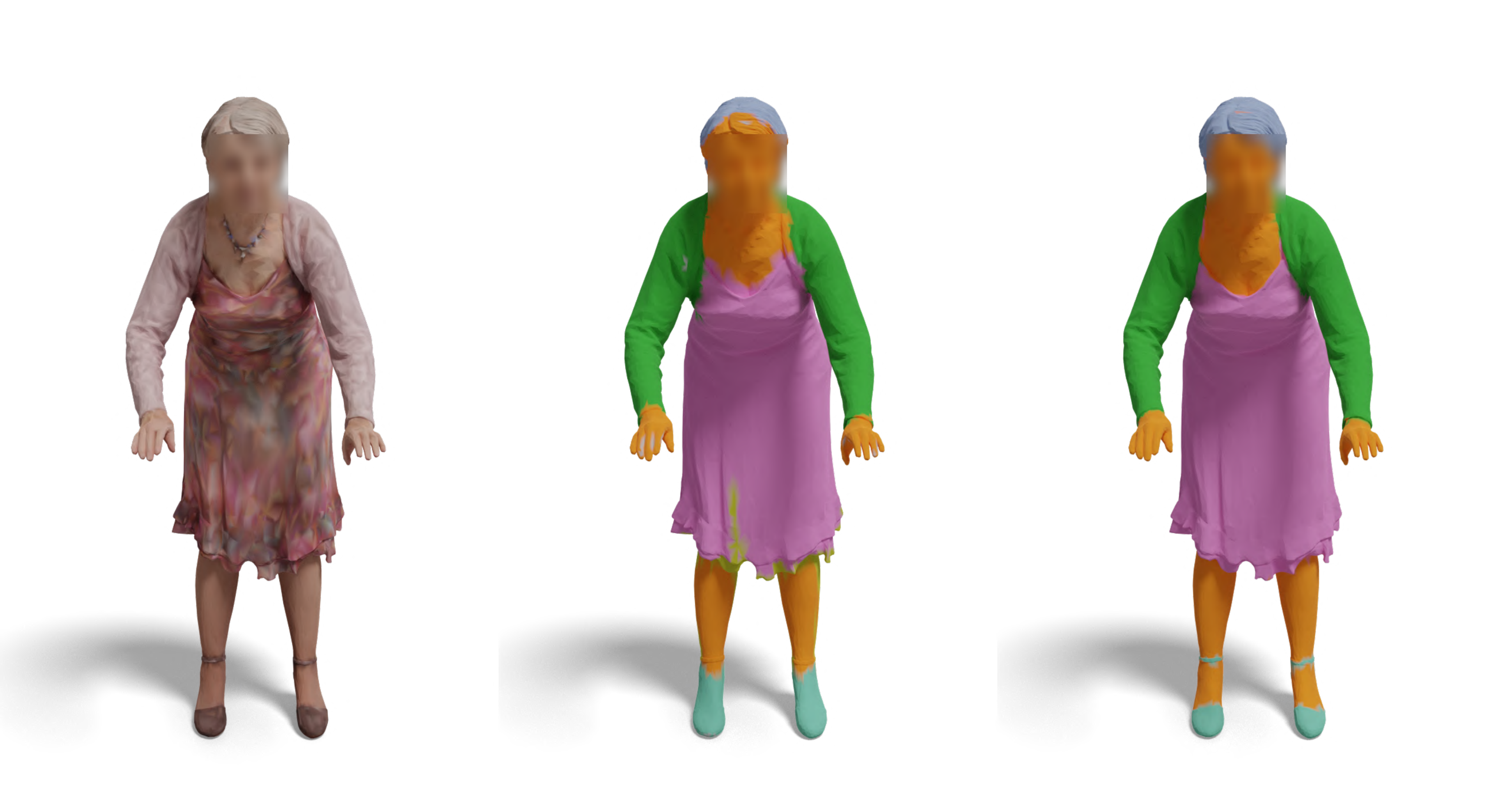}
  \includegraphics[width=0.47\textwidth]{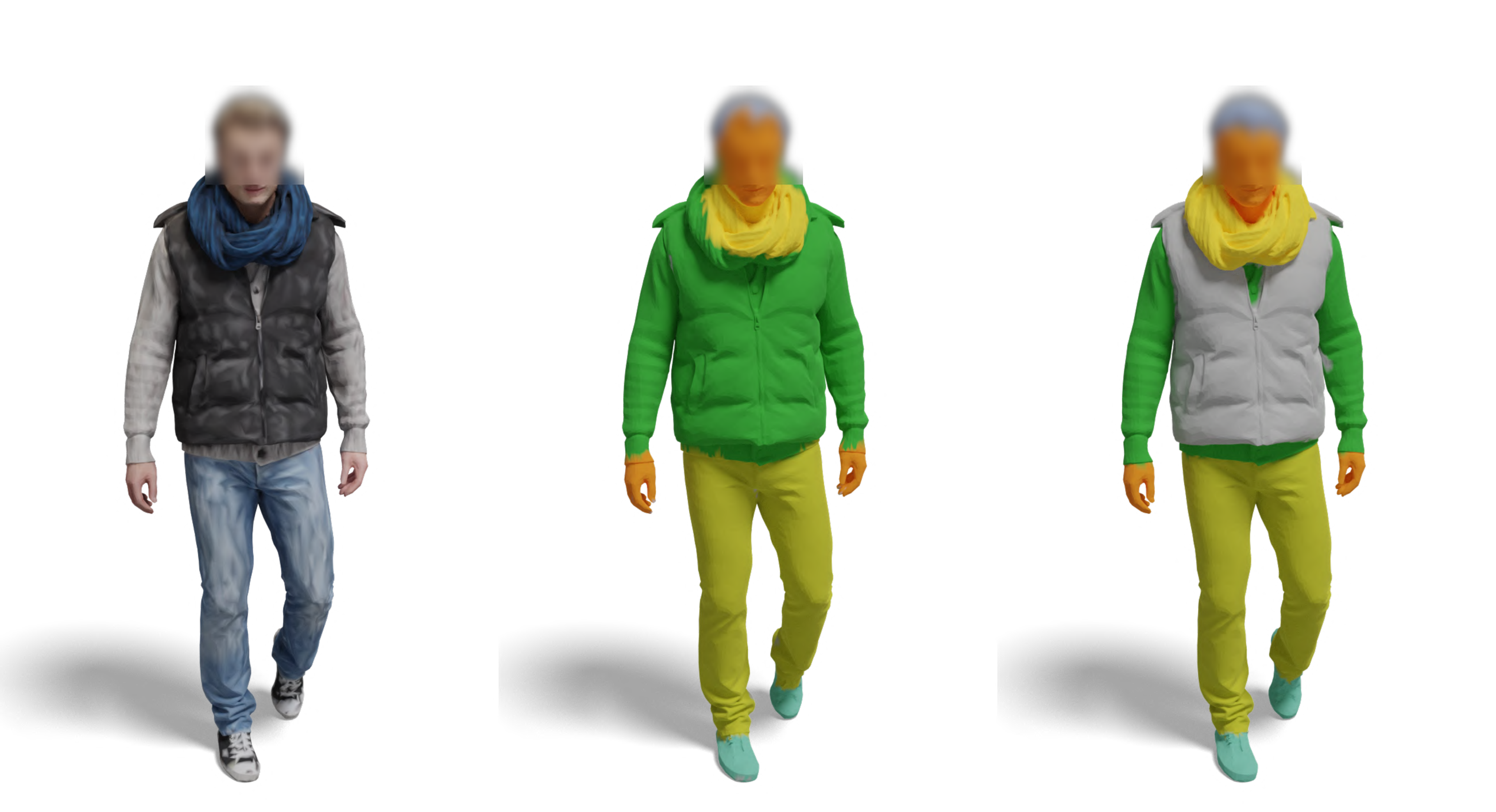}
  \includegraphics[width=0.47\textwidth]{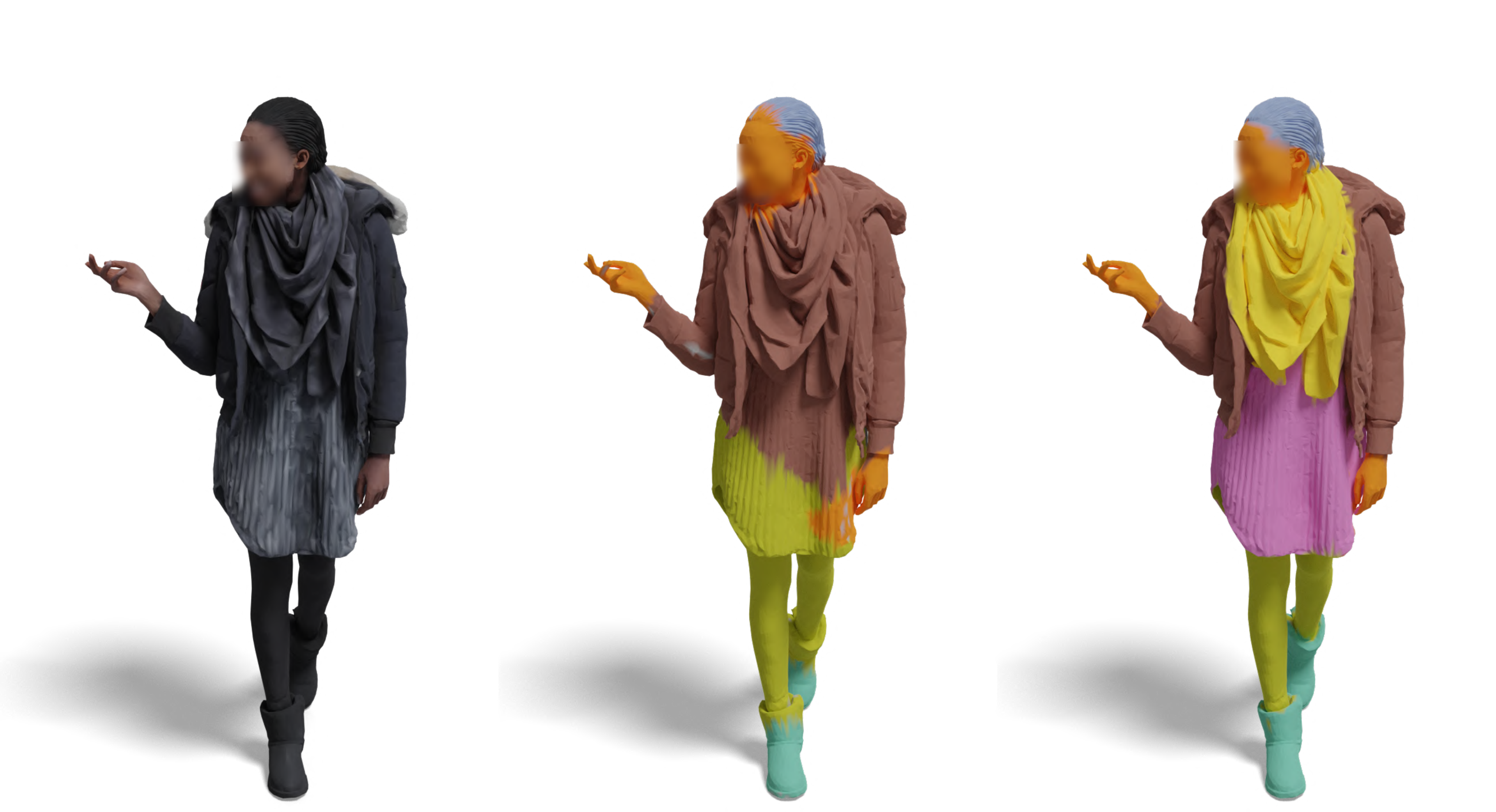}

	\caption{Segmentation labels obtained using our heuristics might result in unclear boundaries(top, middle) and undesired noisy patches (bottom, middle). We clean such noise using \clothtool{} and obtain high-quality labels, as shown on the right. }
	\label{fig:dataset}
\end{figure}

We show some examples of heuristics-based segmentation and manually refined segmentation in ~\figref{fig:dataset} and ~\figref{fig:dataset2}. We explain more details about our interactive tool in ~\secref{supp:tool}.

\paragraph{Ground Truth Segmentation Labels of \clothdata{}i Scans.} 
For  \clothdata{}i, we follow a similar idea, but instead of using SMPL+D registration and SMPL UV space, we use Metashape~\cite{metashape} to perform 2D-to-3D lifting of segmentation labels. The recovered 3D segmentation might be inaccurate because of 1) inaccurate 2D segmentation prediction, and 2) inconsistent 2D segmentation labels across different views. Similar to our processing of \clothdata{}c, we clean noise using heuristics. We define heuristics-based priors on SMPL mesh and clean the labels in for scan points directly. This alleviates the problem of obtaining SMPL+D~\cite{bhatnagar2019mgn} registrations. We deployed two different classes of heuristics:
\begin{itemize}
    \item \textit{Body Parts Heuristics:} We rely on the prior knowledge that some garments should not belong to unusual body parts (e.g., t-shirts on feet, trousers on arms \etc).
    \item \textit{Garments Class Heuristics:} In some cases, we observed artifacts related to specific combinations of garments. In these cases, we deploy an additional set of rules to address these issues specifically.
\end{itemize}

%% file: sec/suppl/02_method.tex
\section{Method}
\label{supp:method}
We explain the details of our model \clothmodel{} in this section. 

\paragraph{Point Encoder.} We use the official implementation of DGCNN~\cite{dgcnn} and use 3 layers of EdgeConvoluation operation, followed by a single-layer MLP.  

\paragraph{Clothing Encoder.} We use a multi-head attention module in the encoder, where $\mathrm{n_{head}} = 4$ in our case. We also apply positional encoding to the query vector($\mathrm{\vect{p'}^2_i}$), before calculating the attention score.

\paragraph{Body Encoder.} $\bodyfeat$ requires the computation of nearest neighbors for each point within the batch, potentially leading to computational overhead during the training process. To mitigate this, we opt to precompute $\bodyfeat$. This is done by finding the nearest point for each scan point from the posed SMPL mesh ($\smpl(\shape, \shape)$). Subsequently, during inference, a preprocessing step is employed to calculate $\bodyfeat$ beforehand, which is then used during inference.

%% file: sec/suppl/03_tool.tex
\section{Interactive Tool}
\label{supp:tool}
In this section, we explain common functionalities provided by our tool and its usage in data annotation and network refinement.

\paragraph{Interactive Tool Interface.} We implement \clothtool{} using Open3D~\cite{open3d} in C++ and introduce an easy-to-use, light-weight interactive 3D tool, which provides following functionalities:

\begin{itemize}
    \item \textbf{I/O operations}: Loading/Saving meshes and labels, Loading/evaluating pre-trained model, Saving/Evaluating refined network.
    \item \textbf{Scene}: Move in the scene with mouse control, change lights, background, \etc
    \item \textbf{User selection}: Easy polygon-based region selection by selecting the polygon edges by clicking.
   \item \textbf{Labeling}: Relabel region based on user selection/majority vote.

\end{itemize}

\begin{figure}[t]
	\includegraphics[width=0.45\textwidth]{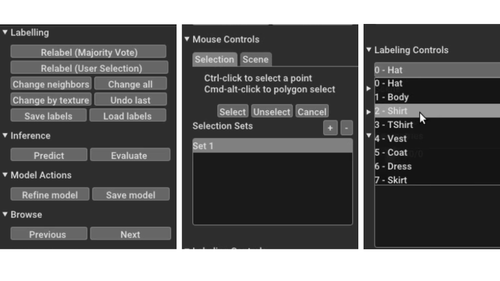}

	\caption{Functionalities provided in \clothtool{} interface includes based I/O operations, mouse-controlled camera movement in the scene, region selection, relabelling, evaluation, and fine-tuning \clothmodel{}.}
	\label{fig:tool}
\end{figure}

\paragraph{Label Correction.} There are multiple options to label the selected regions
\begin{itemize}
        \item  \textit{User selected class }: Manually set the class assigned to the selected areas. The predefined list of classes is shown in the dropdown menu; see \figref{fig:tool}(right).
        \item  \textit{Majority Vote}: If a partial/inaccurate initial segmentation of the scan already exists, the selected region can be labeled more efficiently using the "majority vote" procedure. More precisely, for example, if there is a patch of mislabeled points, the user can select the wider region around it, and label the whole region by the class that is the most commonly present in the region. This makes the labeling procedure much faster.
    \end{itemize}

\paragraph{\clothtool{} for Data Annotation}.

\begin{figure*}[t]

\centering
    \begin{overpic}[width=0.45\textwidth,unit=1bp,tics=20]{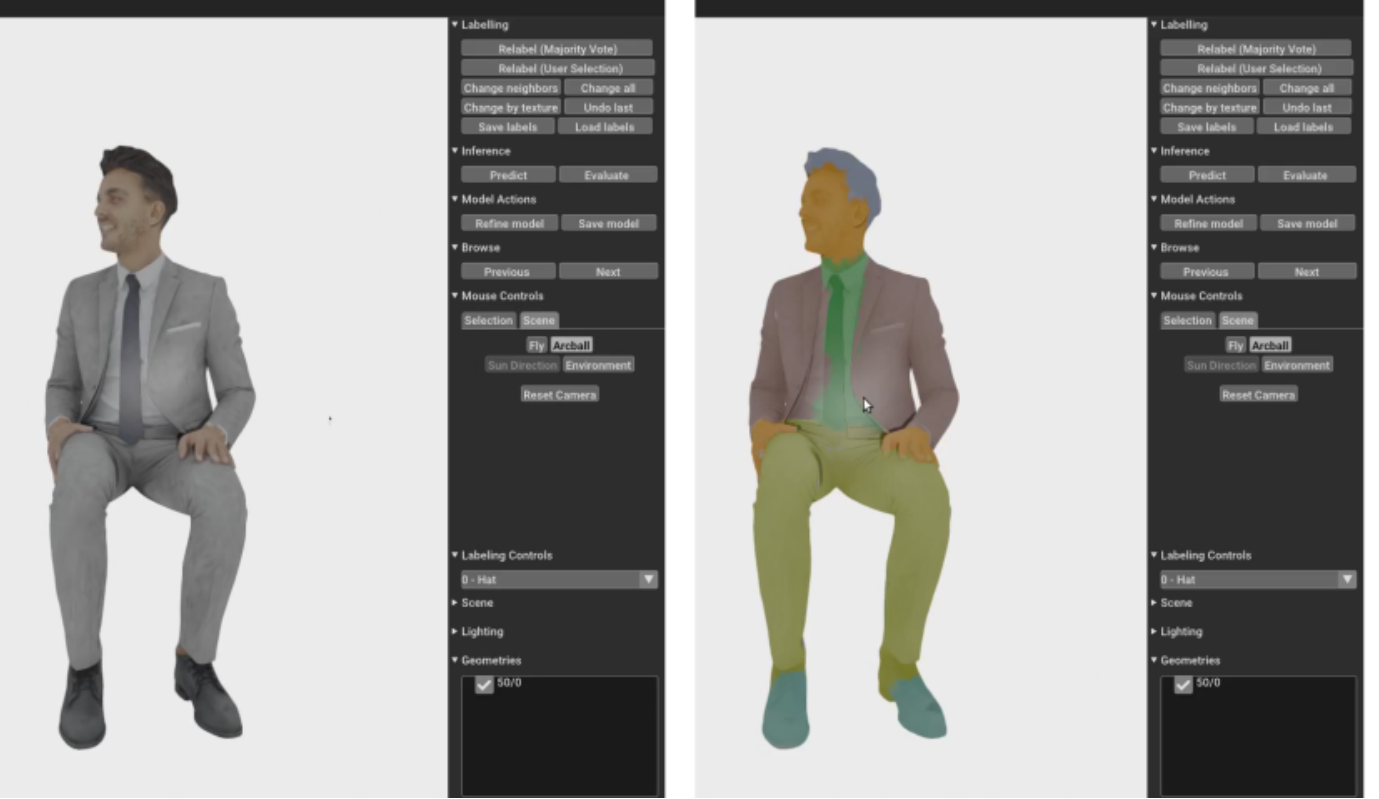}

  \put(40,140){\makebox(0,0){\scriptsize{Load scan with texture}}}
  \put(160,140){\makebox(0,0){\scriptsize{Visualize current segmentation}}}

    \end{overpic}
        \begin{overpic}[width=0.45\textwidth,unit=1bp,tics=20]{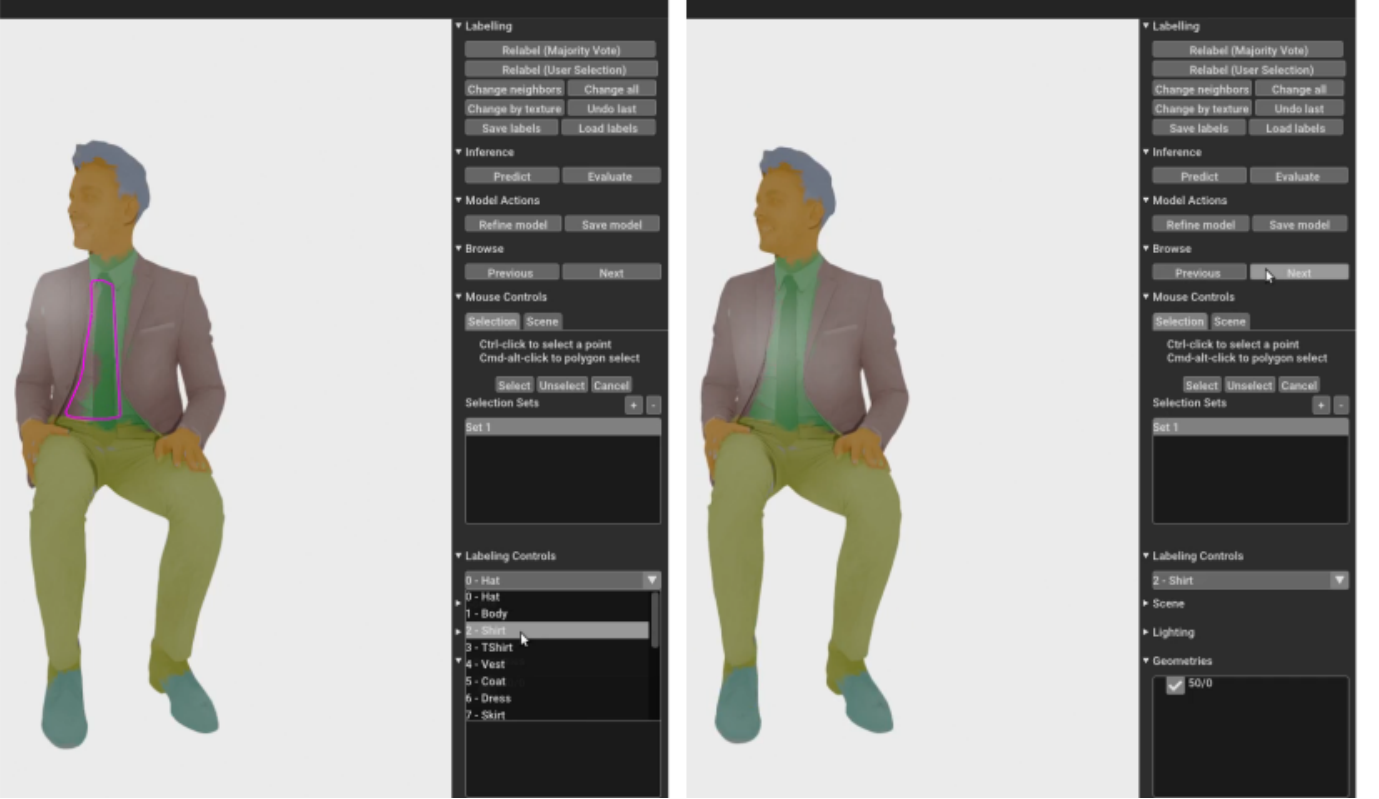}

  \put(40,140){\makebox(0,0){\scriptsize{Select region and relabel }}}
  \put(160,140){\makebox(0,0){\scriptsize{Visualize new segmentation}}}

    \end{overpic}

	\caption{\textbf{Annotation using \clothtool{}.}Using \clothtool{}, we first \emph{load the scan with texture} to understand the scan. We then \emph{visualize current segmentation} as an overlay on the textured scan. After inspection, we identify and \emph{select mislabeled regions} and assign them the correct label from a predefined set. Finally, we \emph{visualize the new segmentation} and inspect by moving the camera around the scene.}
	\label{fig:dataset_tool}
\end{figure*}

We use \clothtool{} to manually clean segmentation and generate high quality segmentation data, as shown in~\figref{fig:dataset} and ~\figref{fig:dataset2}. We provide a demo of labeling process in the supplementary video and visualize key-stage of pipeline in ~\figref{fig:dataset_tool}.

Due to the inherently error-prone nature of the segmentation label generation pipeline, numerous scans displayed noisy boundaries and improperly labeled clothing classes. To address this issue, approximately 1000 scans were annotated using \clothtool{} within the \clothdata{} dataset, while the remaining were carefully verified. Consequently, \clothdata{} comprises a curated segmentation label dataset that has been meticulously verified.

\paragraph{\clothtool{} for Network Refinement.}
We also use \clothtool{} to improve the generalization of our model for real-world datasets. We first predict the segmentation label for a given scan using the pre-trained \clothmodel{}. Since the given scan is out-of-distribution, network results might be incorrect and noisy. We then refine the network prediction for the given scan using the steps mentioned in \emph{data annotation}. We explain training details and experiments in~\secref{supp:results}. The new network is used to infer the given scan again and also evaluated on the test-set of \clothdata{}. All these functions are implemented as a simple button click in the tool, see~\figref{fig:tool}. The newly trained model can be saved and used of this new out-of-distribution dataset for better generalization.

%% file: sec/suppl/04_results.tex
\section{Results}
\label{supp:results}
In this section, we provide more results of our model. In~\secref{supp:baseline}, we provide more comparison with baseline methods, followed by comparison on BUFF~\cite{buff} dataset in ~\secref{supp:prior}. Finally, we provide ablation studies for continual learning setup of our model and show more results on real-world datasets in ~\secref{supp:realworld}.

\subsection{Comparison with baseline}
\label{supp:baseline}
In this section, we analyze more comparisons with part segmentation methods to understand the cause of superior performance of \clothmodel{}. We broadly classify them into 5 factors, as discussed below. These factors act mutually in many cases, widening the disparity between the performance levels of baseline techniques and our proposed approach. In the table \tabref{tab:closedi_test} we provide a quantitative comparison on the test split of \clothdata{}i.

\begin{table*}[t]
\centering

\resizebox{\textwidth}{!}{
\begin{tabular}{lccccccccccccccccccc}
\toprule
Method & Mean  & T-shirt & Shirt  &  Vest & Coat & Hoodies & Short-Pants & Pants  & Skirts & Hat & Shoes & Body & Hair  \\

\midrule
DGCNN~\cite{dgcnn}  & 92.65 &97.50 & 93.23 & 95.78 & 86.89 & 99.54 & 96.89 & 87.27 & 98.90 & 97.26 & 86.17 & 84.19 & 88.14 \\
DeltaConv~\cite{Wiersma2022DeltaConv} & 91.30 &97.19 & 88.12 & 96.57 & 86.98 & 98.55 & 94.39 & 86.87 & 98.69 & 97.26 & 83.42 & 80.33 & 87.29 \\
Ours & \textbf{95.19} & \textbf{99.12} & \textbf{96.18} & \textbf{99.48} & \textbf{87.93} & \textbf{99.69} & \textbf{97.98} & \textbf{89.39} & \textbf{99.05} & \textbf{99.06} & \textbf{89.97} & \textbf{89.78} & \textbf{94.66} \\

\bottomrule
\end{tabular}

}
\vspace{-0.2cm}

\caption{We quantitatively compare the results of our method \sota{} part-segmentation methods, DGCNN~\cite{dgcnn} and DeltaConv~\cite{Wiersma2022DeltaConv}. We report IoU for every class and mean over all the classes($\mathrm{IoU_{mean}}$).}
\label{tab:closedi_test}
\end{table*}

\begin{figure*}[t]
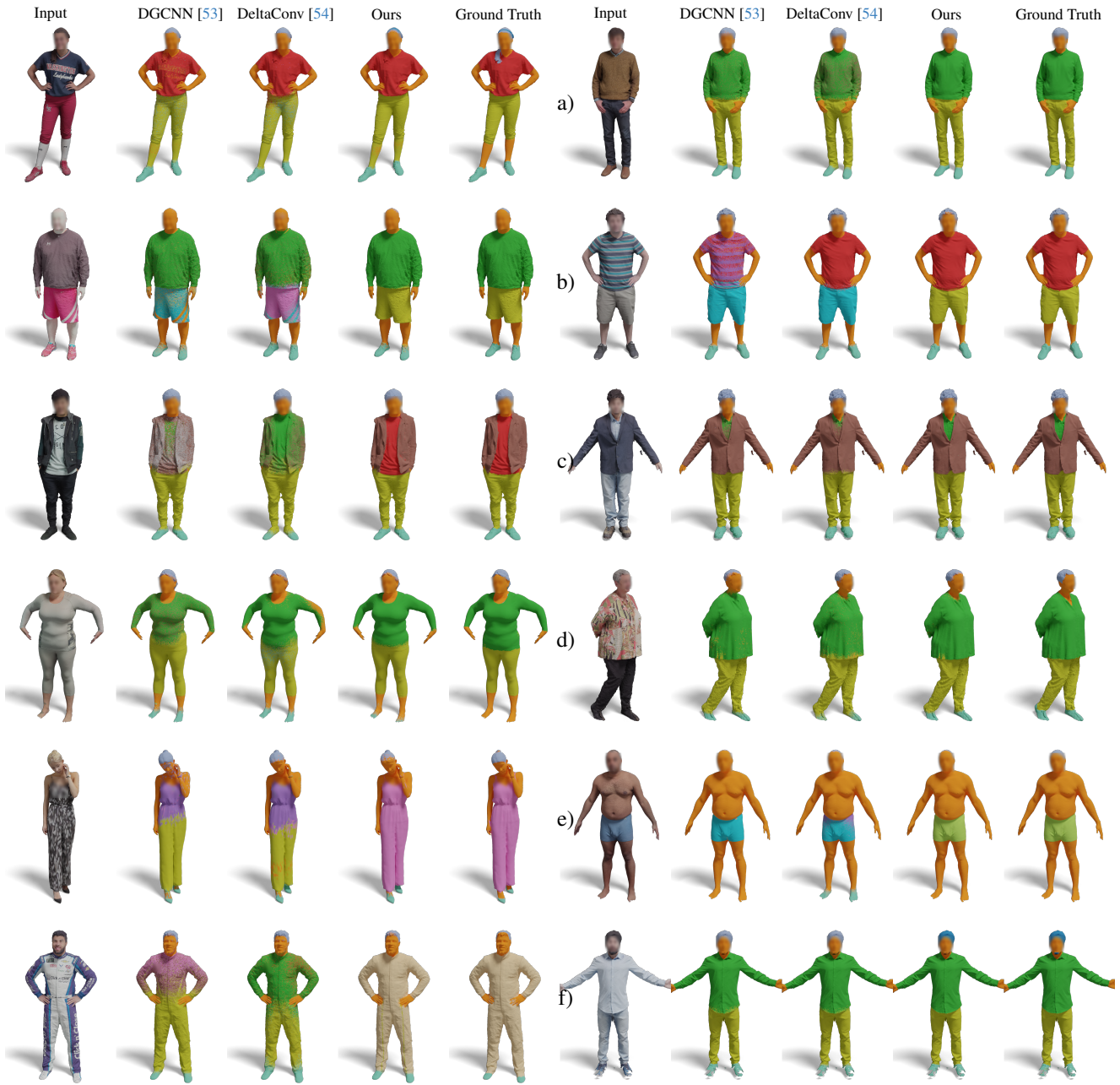


\centering
    \begin{overpic}[width=\textwidth,unit=1bp,tics=20]{images/suppl_im/full_fig_6_supmat.pdf}

  \put(20,480){\makebox(0,0){\scriptsize{Input}}}
  \put(78,480){\makebox(0,0){\scriptsize{DGCNN~\cite{dgcnn}}}}
  \put(125,480){\makebox(0,0){\scriptsize{DeltaConv~\cite{Wiersma2022DeltaConv}}}}
  \put(170,480){\makebox(0,0){\scriptsize{Ours}}}
  \put(220,480){\makebox(0,0){\scriptsize{Ground Truth}}}

  \put(270,480){\makebox(0,0){\scriptsize{Input}}}
  \put(320,480){\makebox(0,0){\scriptsize{DGCNN~\cite{dgcnn}}}}
  \put(370,480){\makebox(0,0){\scriptsize{DeltaConv~\cite{Wiersma2022DeltaConv}}}}
  \put(420,480){\makebox(0,0){\scriptsize{Ours}}}
  \put(470,480){\makebox(0,0){\scriptsize{Ground Truth}}}

  \put(250,440){\makebox(0,0){a)}}
  \put(250,360){\makebox(0,0){b)}}
  \put(250,280){\makebox(0,0){c)}}
  \put(250,200){\makebox(0,0){d)}}
  \put(250,120){\makebox(0,0){e)}}
  \put(250,40){\makebox(0,0){f)}}
  
    \end{overpic}

	\caption{Baseline method like DGCNN~\cite{dgcnn} and DeltaConv~\cite{Wiersma2022DeltaConv} have \textbf{Texture bias} (a, b), are unable to distinguish between \textbf{multi-layer clothing} (c), produces incorrect labels if \textbf{geometry deviates significantly from average body and clothing shapes} (d)  and underperform for \textbf{unbalanced classes} such as dress and hats(e, f).}
	\label{fig:baseline}
\end{figure*}

\paragraph{Clothing Information.} Baseline methods DGCNN~\cite{dgcnn} and DeltaConv~\cite{Wiersma2022DeltaConv} have no prior about clothing present in the scan. As a result, these methods rely on local/global geometric and appearance features. Given the diversity and complexity of clothing items, it is challenging to learn about robust semantics from limited information. As a result, baseline methods seem to generate multiple clothing classes in a vicinity, mislabel clothing classes, and are not able to learn the shape/structure of clothing items. This is evident from all the examples shown in~\figref{fig:baseline}. \clothmodel{} not only takes advantage of clothing information but also learns a more distinctive feature for each clothing class and consequently learns clothing prior based on local features and these clothing features(via attention module). 

\paragraph{Texture Bias.} As observed in ~\figref{fig:baseline}(first and second row), baseline methods are highly sensitive to changes in texture. As a result any steep change in texture results in a new clothing class. However \clothmodel{} produces accurate results. For baseline methods color, normal and location are the only guiding signal without any prior. Given limited training data, they tend to overfit to textures scene during training.  

\paragraph{Multi-layer Clothing.}
We also observe that baseline methods are not able to recover multi-layer clothing labels see~\figref{fig:baseline}(third row). As there is no prior knowledge about clothing present in the scan, baselines rely on texture and geometry information. In such cases, baselines seem to predict the most commonly seen example with texture during training such as hoodies or shirts. On the other hand, the clothing information used in \clothmodel{} helps with better comprehension, even if local features are very similar.

\paragraph{Shape/geometry Bias.}
Similar to texture bias, the baseline method also has geometry bias to some extent. As shown in ~\figref{fig:baseline}(fourth row) loose upper clothing with larger shapes are classified as hoodies, although the labels are not noisy.

\paragraph{Sparse Clothing Classes.}
We also observe that \clothmodel{} performs well for rare clothing classes such as dresses, hats, etc. On the other hand baseline methods fail to generate consistent labels.

\subsection{Comparison with Prior Work}
\label{supp:prior}

We compare or model with prior work GIM3D~\cite{gim3d} on Buff dataset~\cite{buff}. We use 15 scans from BUFF, as in~\cite{gim3d} for evaluation on the 3-class segmentation problem. We use PointNet++~\cite{pointnet2} based model from GIM3D and report the number in~\tabref{tab:prior2}. We observe that for both \clothdata{}-test and BUFF dataset, our model significantly outperforms GIM3D~\cite{gim3d}.

\begin{table}[t]
\centering

\resizebox{0.4\textwidth}{!}{
\begin{tabular}{lcccccc}
\toprule
Dataset &  MGN~\cite{bhatnagar2019mgn} &GIM3D~\cite{gim3d}  & Ours \\

\midrule
\clothdata{}-Test  &   88.88 & 72.04 &  \textbf{92.47}  \\
Buff~\cite{buff}  &  - & 75.41 & \textbf{90.13}  \\

\bottomrule
\end{tabular}

}

\caption{\label{tab:prior2} Comparison with MGN~\cite{bhatnagar2019mgn} and GIM3D~\cite{gim3d} on \clothdata{} and BUFF dataset.}
\end{table}

\subsection{\clothmodel{} on Real-world Datasets}
\label{supp:realworld}
\begin{figure}[t]
	\includegraphics[width=0.45\textwidth]{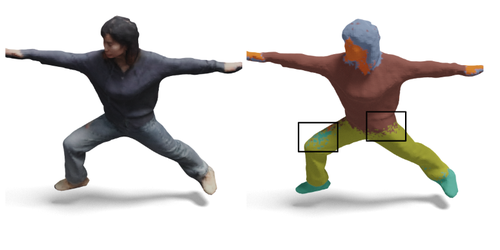}

	\caption{\clothmodel{} predicts blurry boundaries for out-of-distributions scans.}
	\label{fig:public_fail}
\end{figure}

\begin{figure*}[t]

        \begin{overpic}[width=0.24\textwidth,unit=1bp,tics=20]{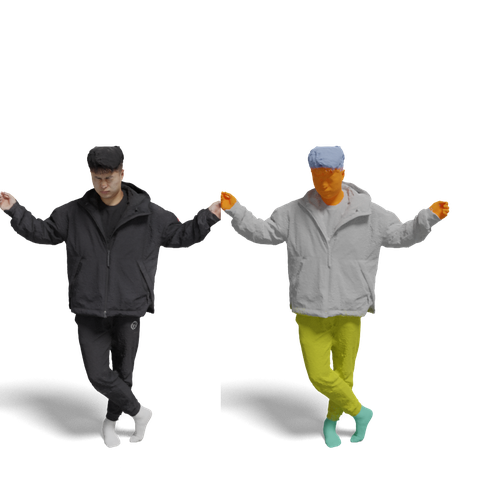}
  \put(25,100){\makebox(0,0){\scriptsize{Input}}}
  \put(80,100){\makebox(0,0){\scriptsize{Prediction}}}
    \end{overpic}
        \begin{overpic}[width=0.24\textwidth,unit=1bp,tics=20]{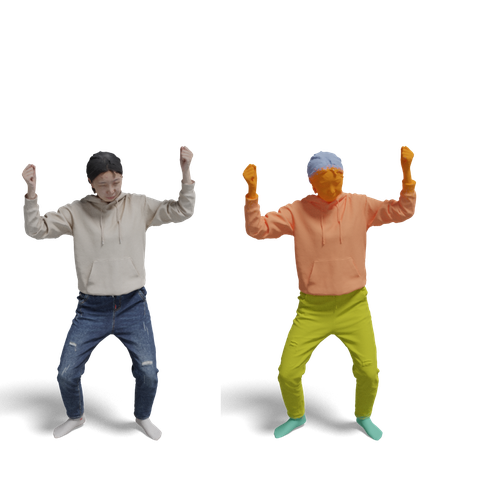}
  \put(25,100){\makebox(0,0){\scriptsize{Input}}}
  \put(80,100){\makebox(0,0){\scriptsize{Prediction}}}
  \put(130,0){\makebox(0,0){\scriptsize{THuman3.0 Scans~\cite{thuman3}}}}
    \end{overpic}
            \begin{overpic}[width=0.24\textwidth,unit=1bp,tics=20]{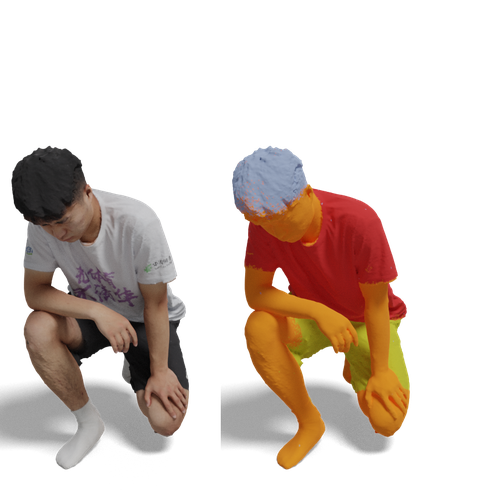}
  \put(25,100){\makebox(0,0){\scriptsize{Input}}}
  \put(80,100){\makebox(0,0){\scriptsize{Prediction}}}
    \end{overpic}
            \begin{overpic}[width=0.24\textwidth,unit=1bp,tics=20]{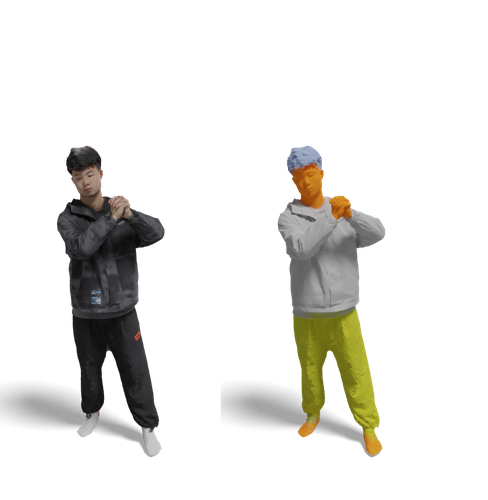}
  \put(25,100){\makebox(0,0){\scriptsize{Input}}}
  \put(80,100){\makebox(0,0){\scriptsize{Prediction}}}
    \end{overpic}

	\includegraphics[width=0.24\textwidth]{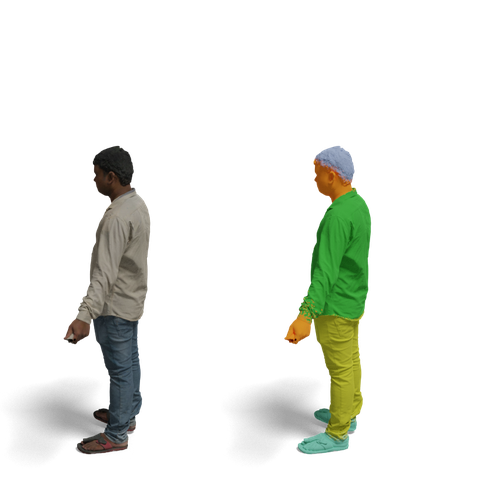}
         \begin{overpic}[width=0.24\textwidth,unit=1bp,tics=20]{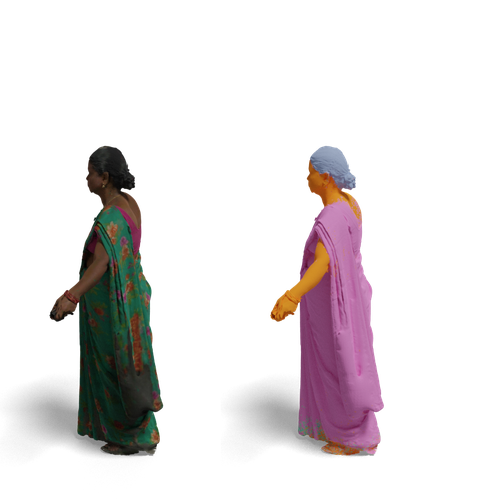}

  \put(130,0){\makebox(0,0){\scriptsize{3DHumans Scans~\cite{Jinka2022}}}}
    \end{overpic}
\includegraphics[width=0.24\textwidth]{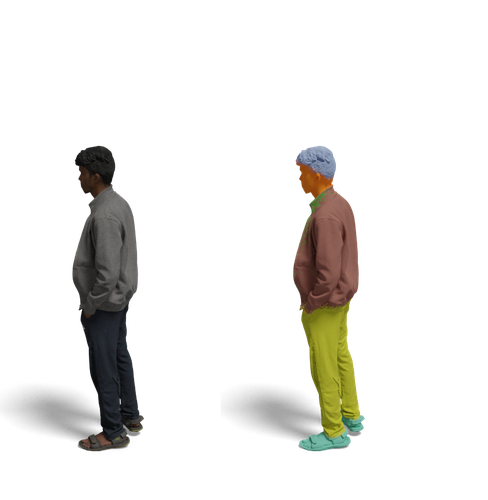}
\includegraphics[width=0.24\textwidth]{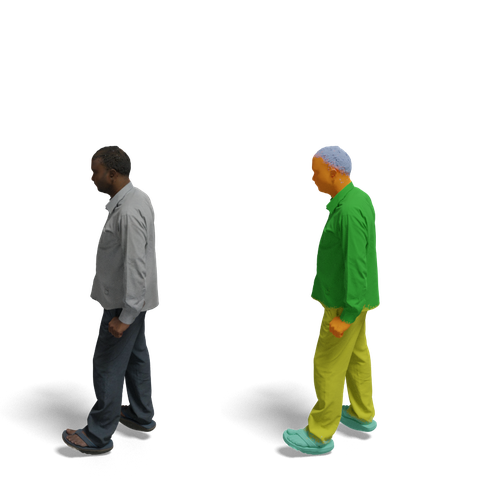}

	\includegraphics[width=0.24\textwidth]{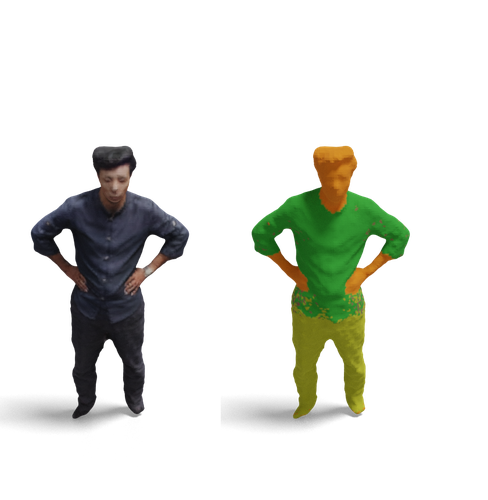}
          \begin{overpic}[width=0.24\textwidth,unit=1bp,tics=20]{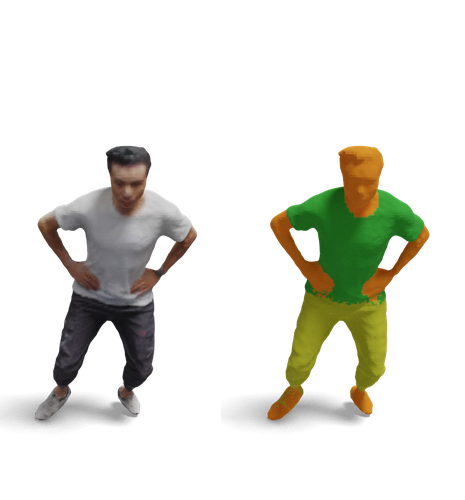}
  \put(130,0){\makebox(0,0){\scriptsize{HuMMan Scans~\cite{cai2022humman}}}}
    \end{overpic}
	\includegraphics[width=0.24\textwidth]{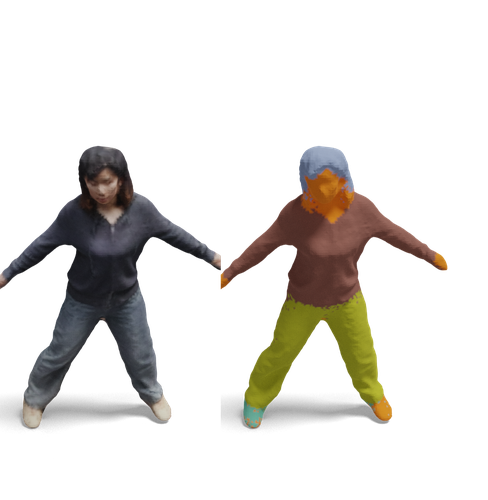}
	\includegraphics[width=0.24\textwidth]{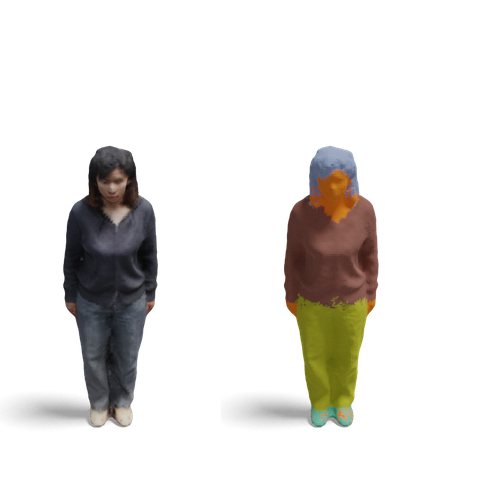}

	\includegraphics[width=0.25\textwidth]{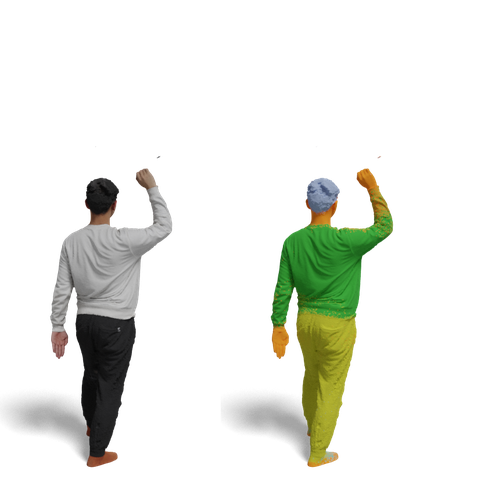}
          \begin{overpic}[width=0.24\textwidth,unit=1bp,tics=20]{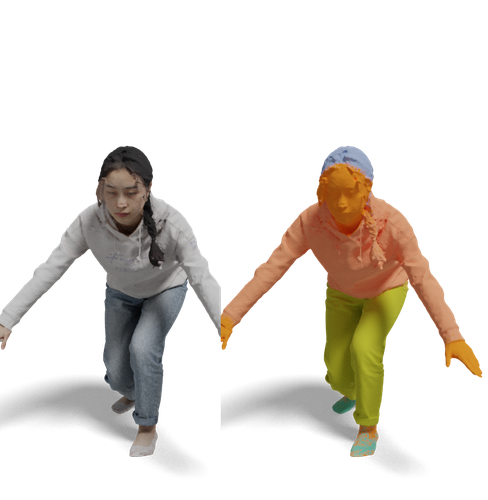}
  \put(130,0){\makebox(0,0){\scriptsize{THuman2.0 Scans~\cite{thuman2}}}}
    \end{overpic}
	\includegraphics[width=0.24\textwidth]{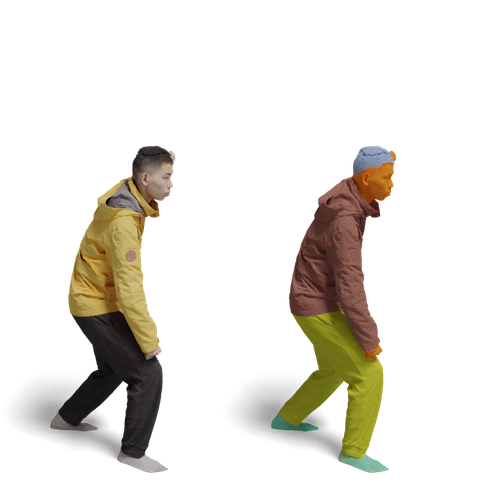}
	\includegraphics[width=0.24\textwidth]{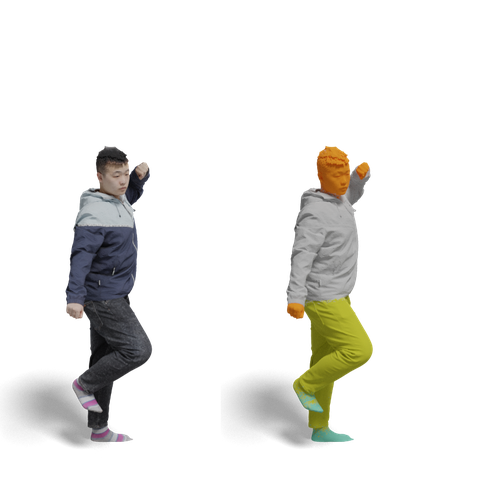}

	\caption{\clothmodel{} results on real-world public datasets~\cite{thuman2,thuman3,Jinka2022,cai2022humman}.}
	\label{fig:result_public}
\end{figure*}

We qualitatively evaluate \clothmodel{} on publicly avaialble real-world datasets such as THuman2.0~\cite{thuman2}, THuman3.0~\cite{thuman3}, HuMMan~\cite{cai2022humman}, 3DHumans~\cite{Jinka2022}. We have added more results in~\figref{fig:result_public}. We observe that for all datasets, \clothmodel{} generates good results and generalizes well. However, in some cases, it results in blurry boundaries and noisy patches of labels, as shown in~\figref{fig:public_fail}.

We propose to improve the performance of our model for such out-of-distribution scans, by fine-tuning the model in a continual learning framework. We follow~\cite{continualsurvery} and experiment with various loss combinations and training configurations to find an optimal setup, such that network performance improves on new out-of-distribution scans without catastrophic forgetting. We show  the ablation in~\tabref{tab:closet_refinement}. We compare the mean IoU on test split of \clothdata{}, after iteratively fine-tuning on 2 sets of scans from this new distribution. Based on experiments, we pick the full loss(eq. 5, main paper) as training loss and only train the last layer of the segmentation decoder and MLP of the Point Encoder. We fine-tune the model for 2 epochs only. 

\begin{table}[h]
\centering
\caption{Performance($\mathrm{IoU_{mean}}$) on \clothdata{}-test after network refinement.}
\resizebox{0.47\textwidth}{!}{
\begin{tabular}{lccccc}
\toprule
Layers trained &  Naive loss  &   Weighted cross-entropy & Full \\

\midrule
$\segdecoder$-last & 90.33  & 90.37 &  90.25 \\
$\segdecoder$-full  & 89.14  &  88.53 & 88.50  \\
$\segdecoder$-last+ $f_{\mathrm{MLP}}$ & 90.62  & 90.53  & 90.33 \\
$\segdecoder$-full+ $f_{\mathrm{MLP}}$ & 89.00 & 88.53  & 88.95 \\
$\segdecoder$-last+ $f_{\mathrm{MLP}}$ + $f^3$& 90.53 & 90.18 & 90.35 \\
$\segdecoder$-full+ $f_{\mathrm{MLP}}$ + $f^3$& 89.00 & 88.53 & 88.62 \\

\bottomrule
\end{tabular}
\label{tab:closet_refinement}
}
\end{table}

\paragraph{Segmenting 4D Scans using \clothmodel{} and \clothtool{}.}

\begin{figure*}[t]

    \centering
 	\includegraphics[height=0.17\textwidth]{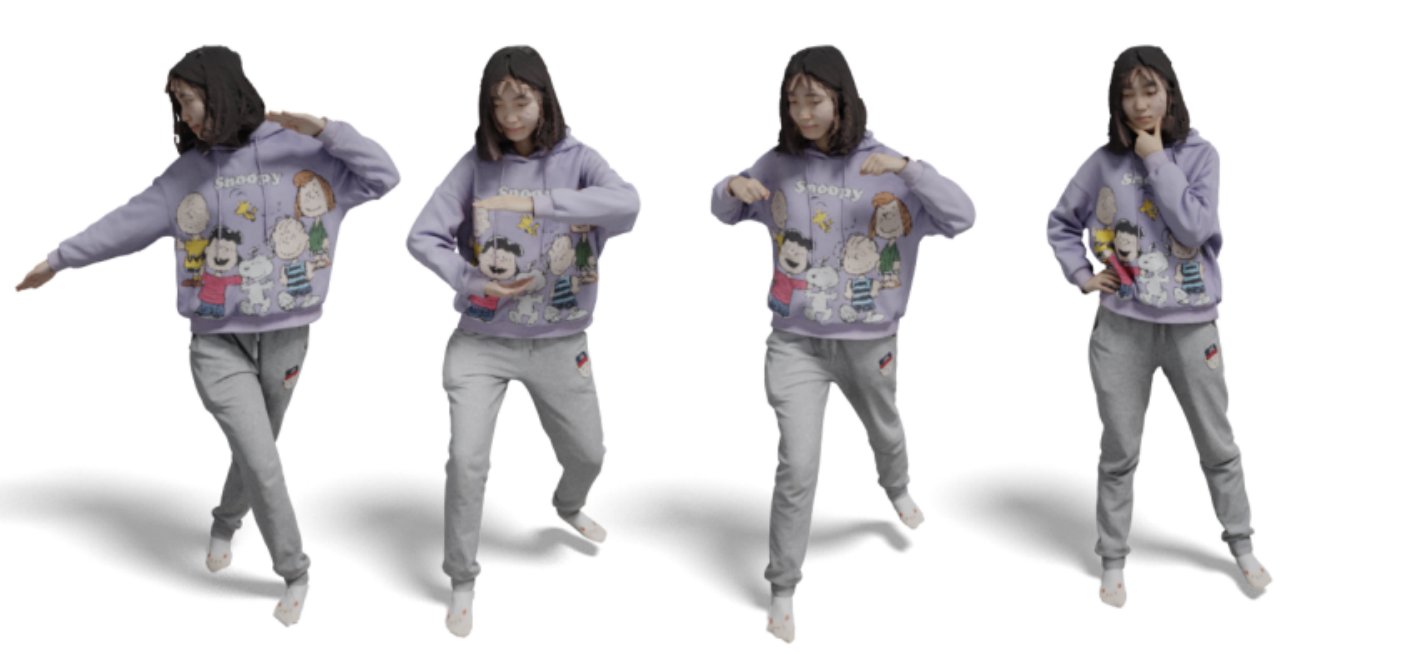}
    \includegraphics[height=0.17\textwidth]{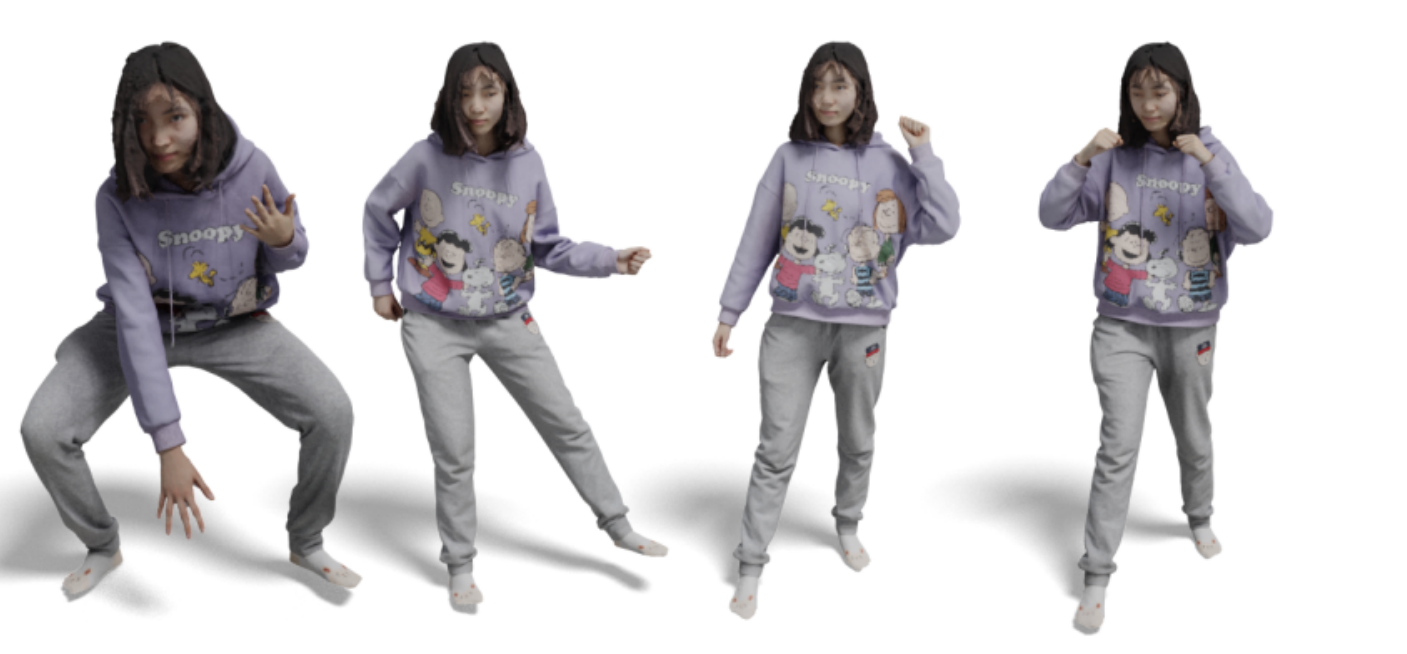}	
  \includegraphics[height=0.17\textwidth]{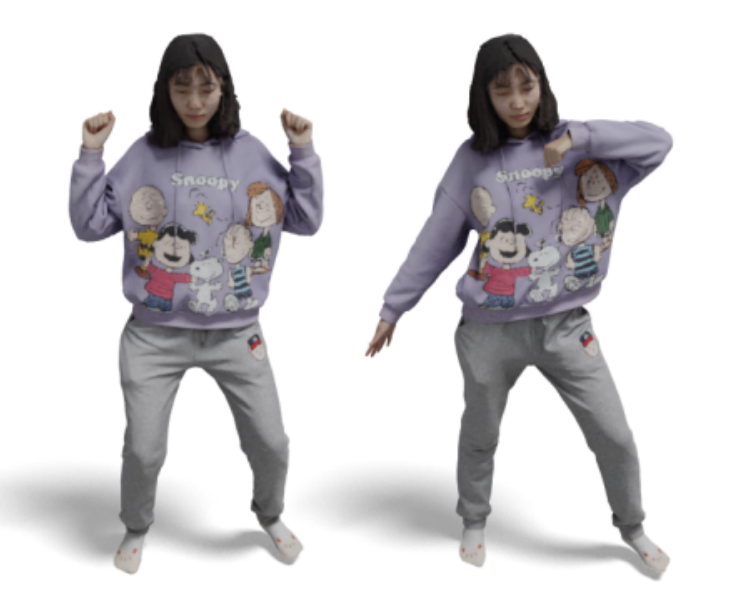}	
 
  \includegraphics[height=0.17\textwidth]{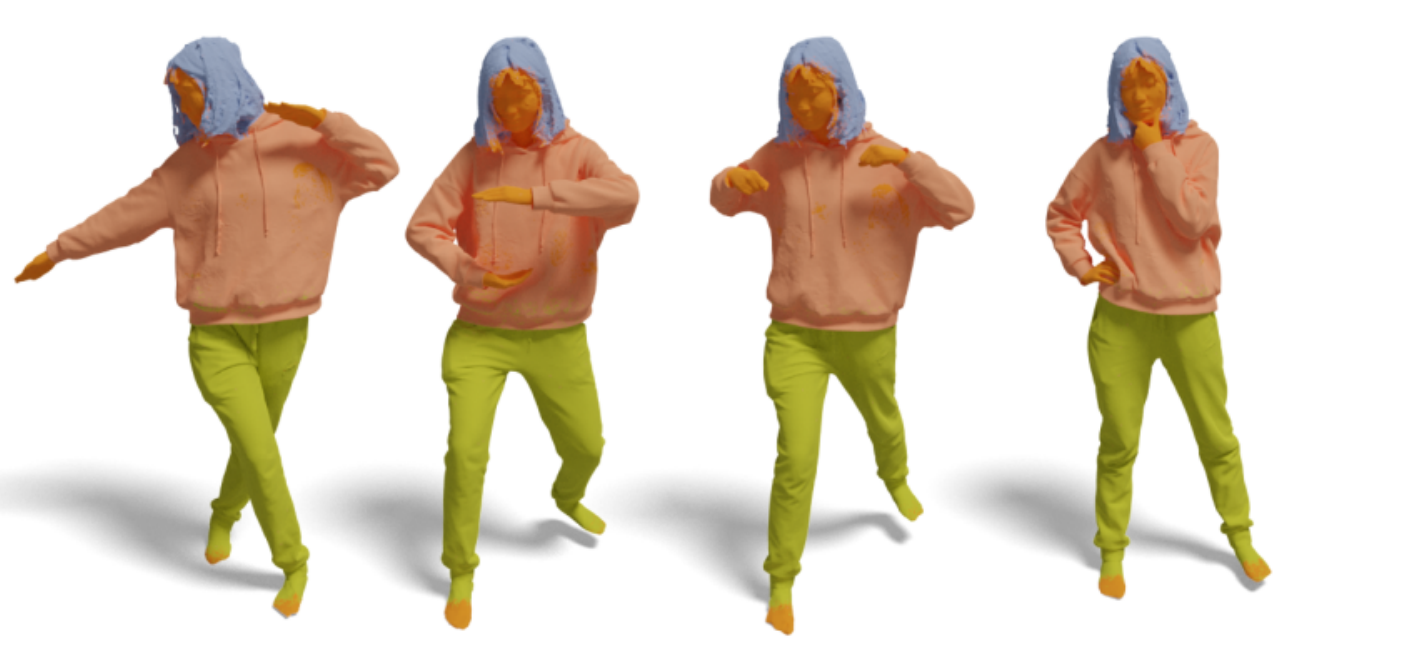}
   \includegraphics[height=0.17\textwidth]{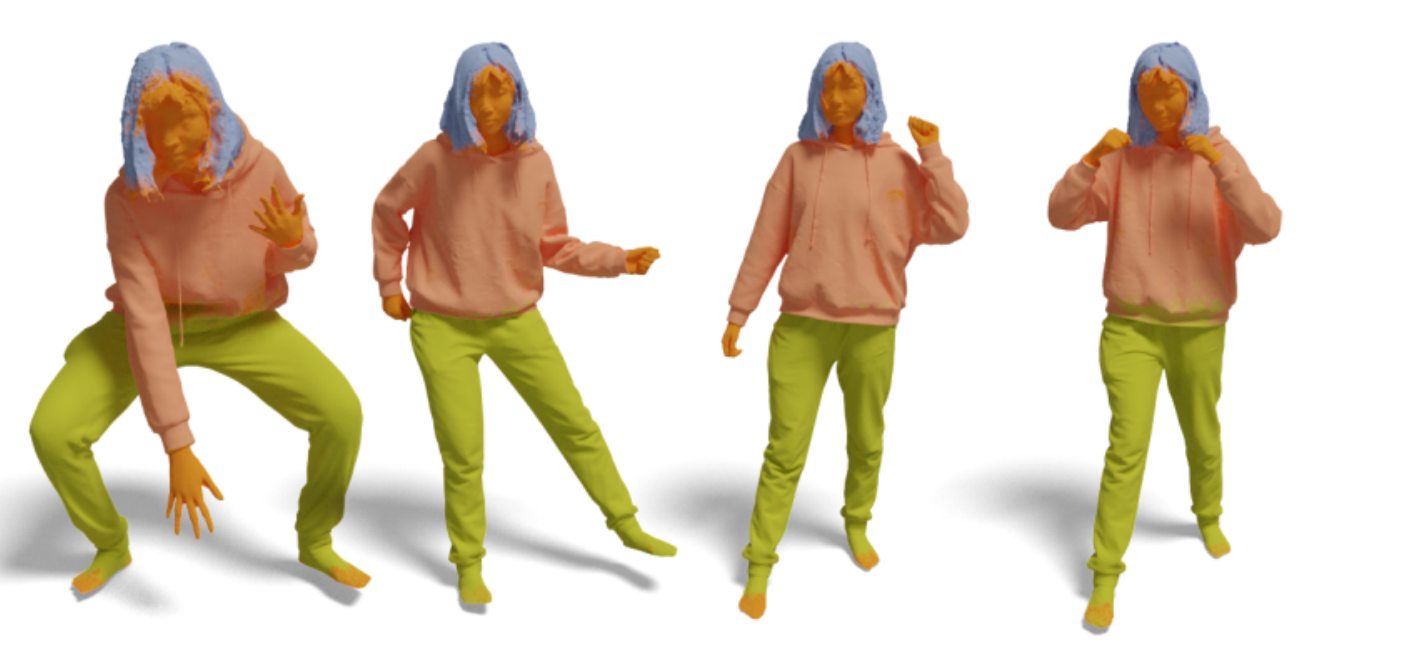}\includegraphics[height=0.17\textwidth]{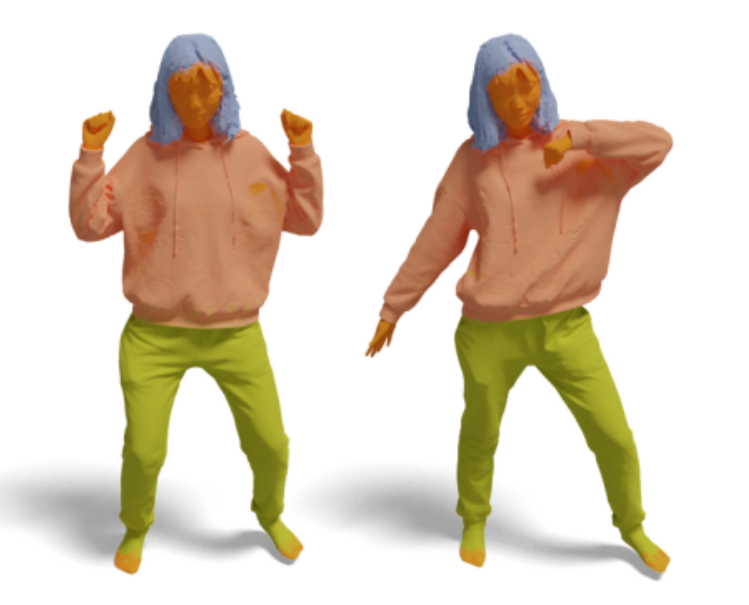}

 	\includegraphics[width=0.31\textwidth]{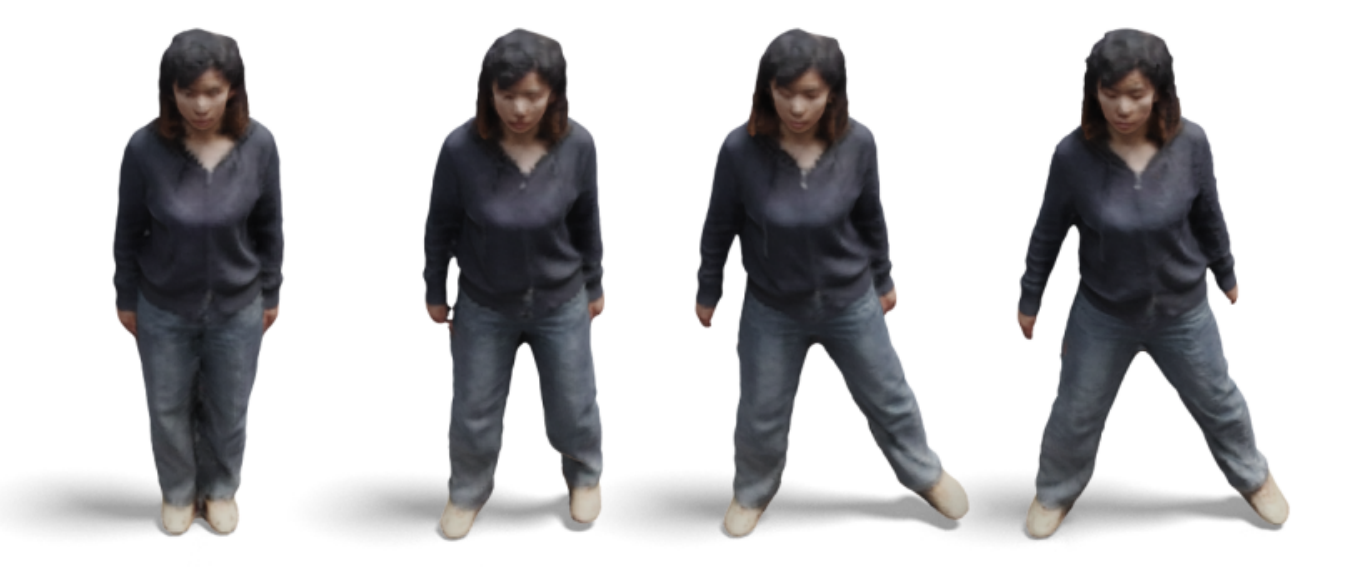}
	\includegraphics[width=0.36\textwidth]{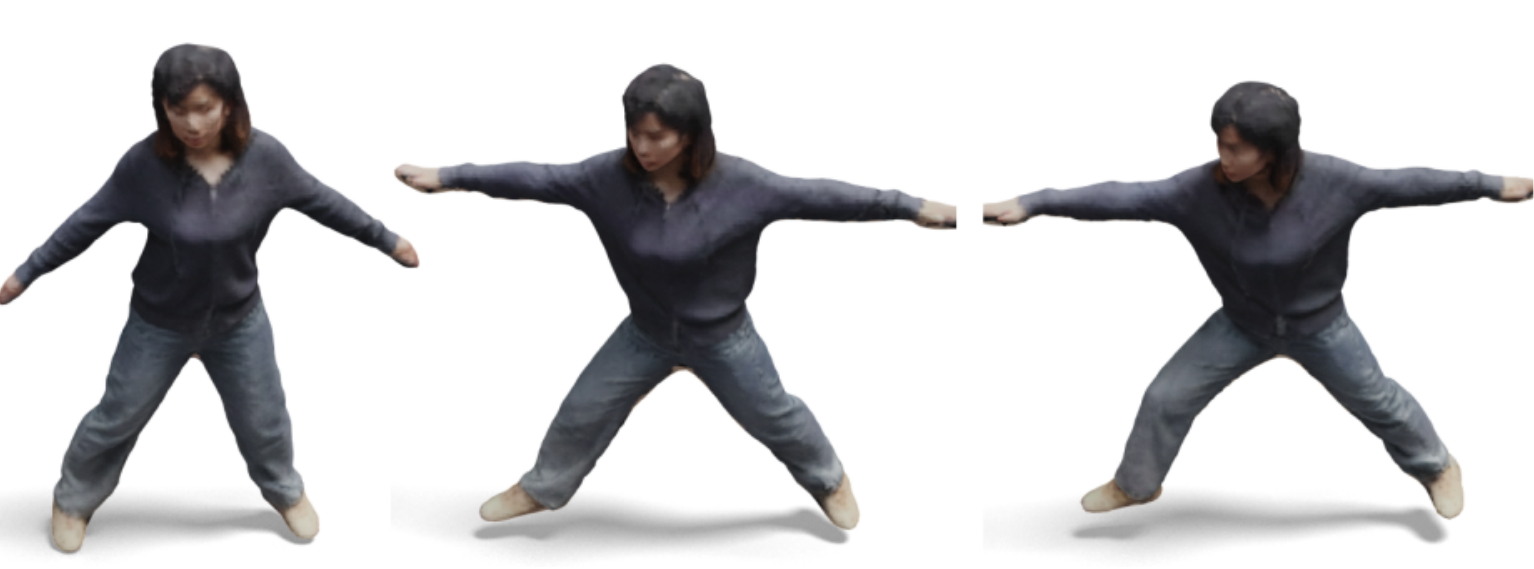}	
 	\includegraphics[width=0.28\textwidth]{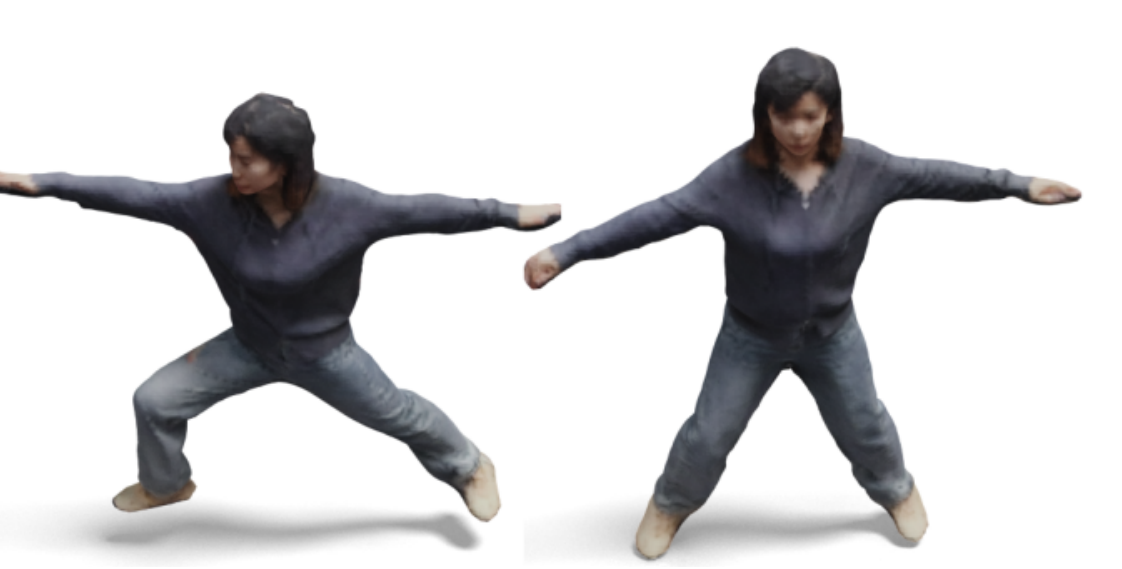}	
 
 	\includegraphics[width=0.31\textwidth]{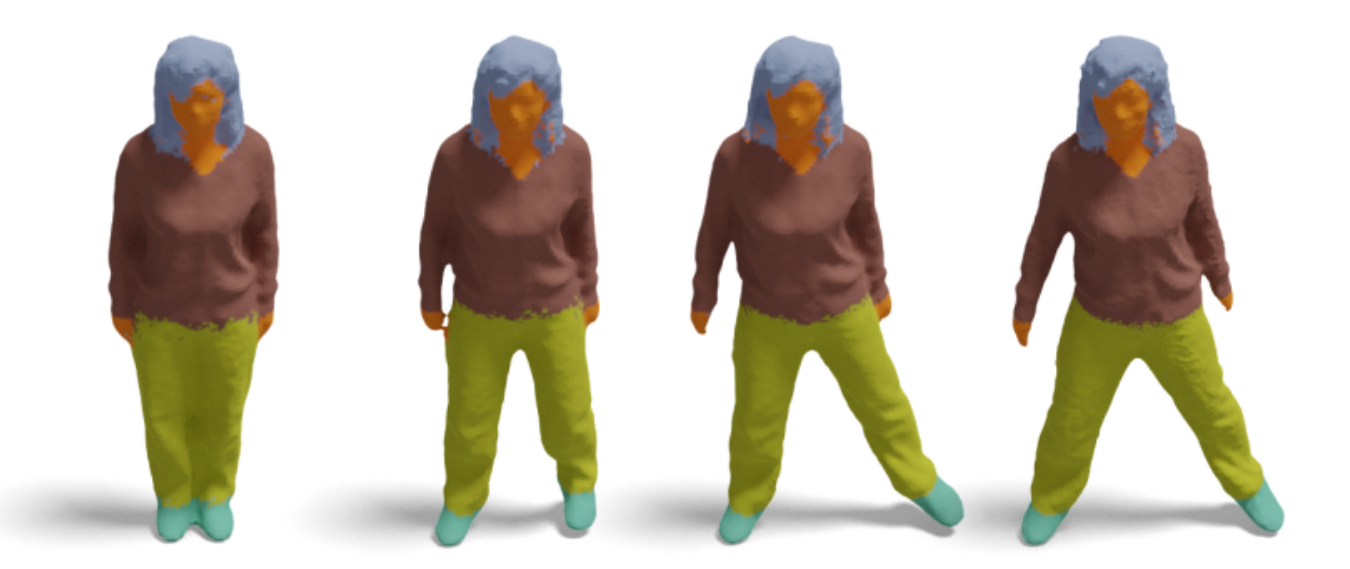}
	\includegraphics[width=0.36\textwidth]{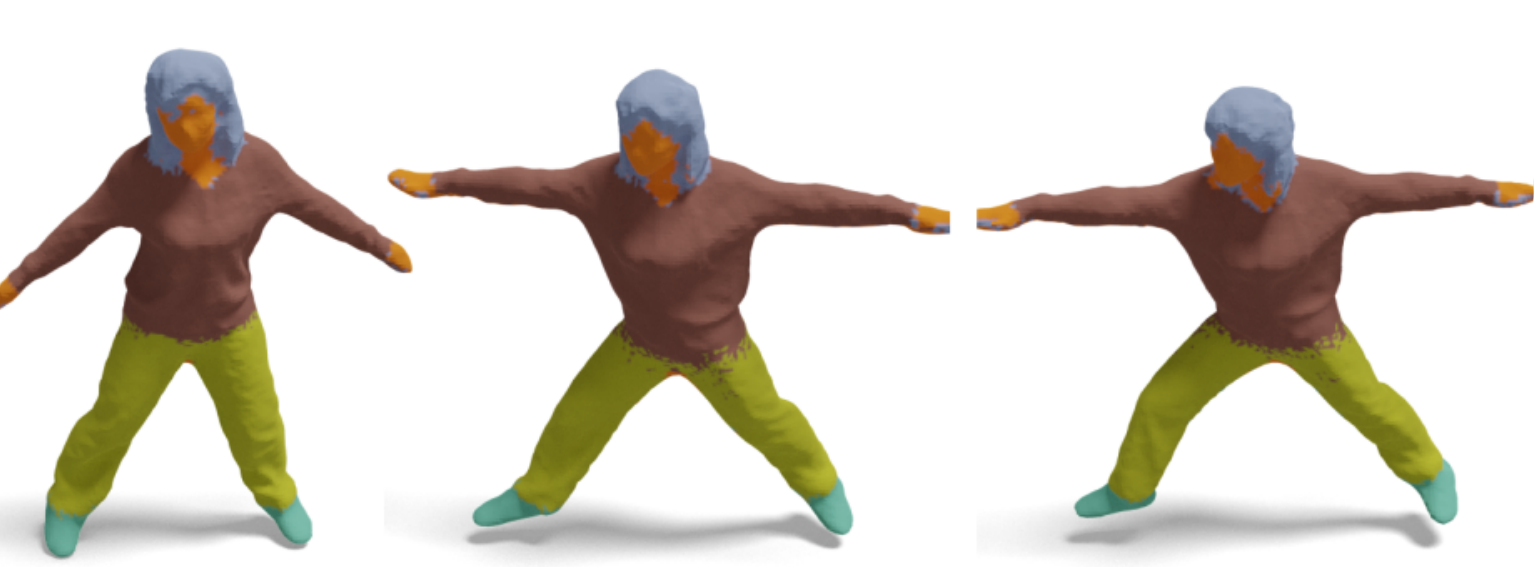}	
 	\includegraphics[width=0.28\textwidth]{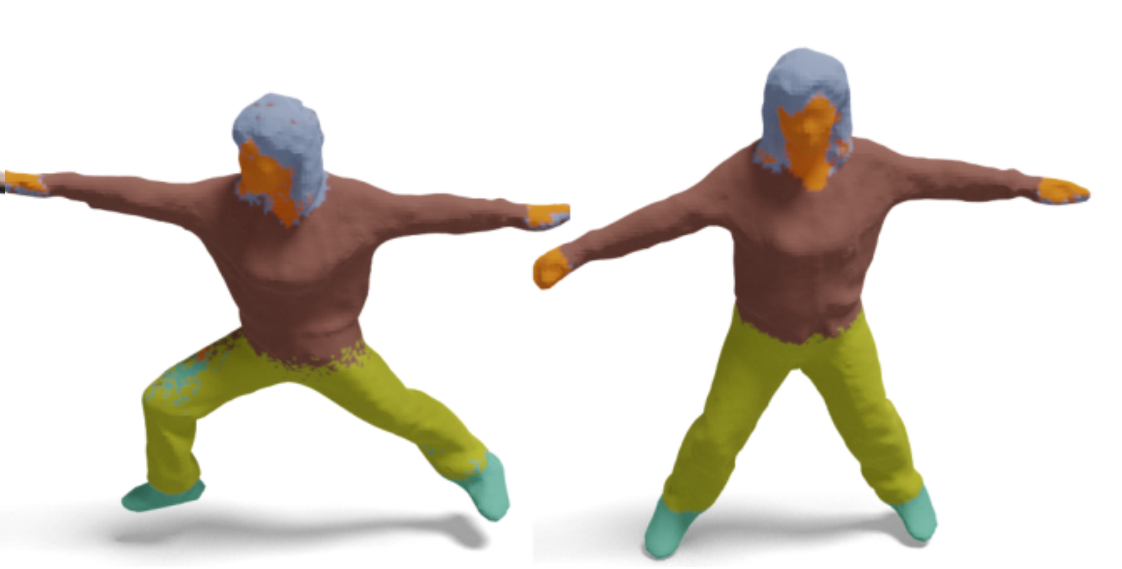}	
 
	\caption{\clothmodel{} is fine-tuned using \clothtool{} on a single frame of a sequence to improve generalization on the remaining frames. We show results of fine-tuned \clothmodel{} on THuman3.0~\cite{thuman3}(top)  and HuMMan~\cite{cai2022humman}(bottom). Fine-tuned networks result in consistent predictions.}
	\label{fig:result_public2}
\end{figure*}

We use the aforementioned setup to improve segmentation accuracy for a given 4d sequence. We randomly pick one frame of a 4D sequence and refine the model as per this scan. This is similar to one-shot fine-tuning. Then we generate the segmentation labels for the whole sequence. Since the model has now learned appearance and geometry features of one frame, this results in improved accuracy for remaining frames. We show results on a set of poses from THuman3.0 and HuMMan in\figref{fig:result_public2}. 

Finally, we have generated high quality segmentation labels of approximately 1000 scans(from diverse sources~\cite{thuman2,thuman3,cai2022humman,Jinka2022}) using \clothmodel{} and \clothtool{}. We will release this as \clothdata++.

%% file: main_arxiv.bbl
\begin{thebibliography}{66}
\providecommand{\natexlab}[1]{#1}
\providecommand{\url}[1]{\texttt{#1}}
\expandafter\ifx\csname urlstyle\endcsname\relax
  \providecommand{\doi}[1]{doi: #1}\else
  \providecommand{\doi}{doi: \begingroup \urlstyle{rm}\Url}\fi

\bibitem[3dm()]{3dmd}
3d{MD}.

\bibitem[axy()]{axyz}
{AXYZ} design.

\bibitem[lum()]{lumaai}
Luma{AI} {L}abs.

\bibitem[md()]{md}
{M}arvelous {D}esigner.

\bibitem[met()]{metashape}
Agisoft {M}etashape: Reconstruction from {I}mages.

\bibitem[ren()]{renderpeople}
{3D} {P}eople from {R}enderpeople.

\bibitem[tre()]{treedys}
Treedy static scanner.

\bibitem[twi()]{twindom}
Twindom {3D} {S}cans.

\bibitem[Bertiche et~al.(2019)Bertiche, Madadi, and Escalera]{cloth3d}
Hugo Bertiche, Meysam Madadi, and Sergio Escalera.
\newblock {CLOTH3D:} clothed 3d humans.
\newblock \emph{CoRR}, abs/1912.02792, 2019.

\bibitem[Bhatnagar et~al.(2019)Bhatnagar, Tiwari, Theobalt, and
  Pons-Moll]{bhatnagar2019mgn}
Bharat~Lal Bhatnagar, Garvita Tiwari, Christian Theobalt, and Gerard Pons-Moll.
\newblock Multi-garment net: Learning to dress {3D} people from images.
\newblock In \emph{ICCV}, 2019.

\bibitem[Bhatnagar et~al.(2020)Bhatnagar, Sminchisescu, Theobalt, and
  Pons-Moll]{bhatnagar2020ipnet}
Bharat~Lal Bhatnagar, Cristian Sminchisescu, Christian Theobalt, and Gerard
  Pons-Moll.
\newblock Combining implicit function learning and parametric models for 3d
  human reconstruction.
\newblock In \emph{European Conference on Computer Vision ({ECCV})}.
  {Springer}, 2020.

\bibitem[Cai et~al.(2022)Cai, Ren, Zeng, Lin, Yu, Wang, Fan, Gao, Yu, Pan,
  Hong, Zhang, Loy, Yang, and Liu]{cai2022humman}
Zhongang Cai, Daxuan Ren, Ailing Zeng, Zhengyu Lin, Tao Yu, Wenjia Wang,
  Xiangyu Fan, Yang Gao, Yifan Yu, Liang Pan, Fangzhou Hong, Mingyuan Zhang,
  Chen~Change Loy, Lei Yang, and Ziwei Liu.
\newblock {HuMMan}: Multi-modal 4d human dataset for versatile sensing and
  modeling.
\newblock In \emph{17th European Conference on Computer Vision, Tel Aviv,
  Israel, October 23--27, 2022, Proceedings, Part VII}, pages 557--577.
  Springer, 2022.

\bibitem[Charles et~al.(2017)Charles, Su, Kaichun, and
  Guibas]{8099499_pointnet}
R.~Qi Charles, Hao Su, Mo Kaichun, and Leonidas~J. Guibas.
\newblock Pointnet: Deep learning on point sets for 3d classification and
  segmentation.
\newblock In \emph{2017 IEEE Conference on Computer Vision and Pattern
  Recognition (CVPR)}, pages 77--85, 2017.

\bibitem[Chen et~al.(2022)Chen, Jiang, Song, Yang, Black, Geiger, and
  Hilliges]{chen2022gdna}
Xu Chen, Tianjian Jiang, Jie Song, Jinlong Yang, Michael~J Black, Andreas
  Geiger, and Otmar Hilliges.
\newblock gdna: Towards generative detailed neural avatars.
\newblock \emph{arXiv}, 2022.

\bibitem[Gong et~al.(2017)Gong, Liang, Shen, and Lin]{lip}
Ke Gong, Xiaodan Liang, Xiaohui Shen, and Liang Lin.
\newblock Look into person: Self-supervised structure-sensitive learning and
  {A} new benchmark for human parsing.
\newblock \emph{CoRR}, abs/1703.05446, 2017.

\bibitem[Gong et~al.(2018)Gong, Liang, Li, Chen, Yang, and Lin]{pgn}
Ke Gong, Xiaodan Liang, Yicheng Li, Yimin Chen, Ming Yang, and Liang Lin.
\newblock Instance-level human parsing via part grouping network.
\newblock In \emph{Proceedings of the European Conference on Computer Vision
  (ECCV)}, 2018.

\bibitem[Guo et~al.(2019)Guo, Huang, Zhang, Srikhanta, Cui, Li, R.Scott, Adam,
  and Belongie]{iMaterialist}
Sheng Guo, Weilin Huang, Xiao Zhang, Prasanna Srikhanta, Yin Cui, Yuan Li,
  Matthew R.Scott, Hartwig Adam, and Serge Belongie.
\newblock The imaterialist fashion attribute dataset.
\newblock \emph{arXiv preprint arXiv:1906.05750}, 2019.

\bibitem[He et~al.(2021)He, Yu, Liu, Yang, Sun, Wang, Fu, Zou, and
  Mian]{3d_segmentation_survey}
Yong He, Hongshan Yu, Xiaoyan Liu, Zhengeng Yang, Wei Sun, Yaonan Wang, Qiang
  Fu, Yanmei Zou, and Ajmal Mian.
\newblock Deep learning based 3d segmentation: {A} survey.
\newblock \emph{CoRR}, abs/2103.05423, 2021.

\bibitem[Ji et~al.(2006)Ji, Liu, Chen, and Wang]{easymeshcutting}
Zhongping Ji, Ligang Liu, Zhonggui Chen, and Guojin Wang.
\newblock Easy mesh cutting.
\newblock \emph{Computer Graphics Forum}, 25\penalty0 (3):\penalty0 283--291,
  2006.

\bibitem[Jinka et~al.(2022)Jinka, Srivastava, Pokhariya, Sharma, and
  Narayanan]{Jinka2022}
Sai~Sagar Jinka, Astitva Srivastava, Chandradeep Pokhariya, Avinash Sharma, and
  P.~J. Narayanan.
\newblock Sharp: Shape-aware reconstruction of people in loose clothing.
\newblock \emph{International Journal of Computer Vision}, 2022.

\bibitem[Kirkpatrick et~al.(2017)Kirkpatrick, Pascanu, Rabinowitz, Veness,
  Desjardins, Rusu, Milan, Quan, Ramalho, Grabska-Barwinska, Hassabis, Clopath,
  Kumaran, and Hadsell]{Kirkpatrick_2017}
James Kirkpatrick, Razvan Pascanu, Neil Rabinowitz, Joel Veness, Guillaume
  Desjardins, Andrei~A. Rusu, Kieran Milan, John Quan, Tiago Ramalho, Agnieszka
  Grabska-Barwinska, Demis Hassabis, Claudia Clopath, Dharshan Kumaran, and
  Raia Hadsell.
\newblock Overcoming catastrophic forgetting in neural networks.
\newblock \emph{Proceedings of the National Academy of Sciences}, 114\penalty0
  (13):\penalty0 3521--3526, 2017.

\bibitem[Kontogianni et~al.(2022)Kontogianni, Celikkan, Tang, and
  Schindler]{interacitve_siyu}
Theodora Kontogianni, Ekin Celikkan, Siyu Tang, and Konrad Schindler.
\newblock Interactive object segmentation in 3d point clouds, 2022.

\bibitem[L\"{a}hner et~al.(2018)L\"{a}hner, Cremers, and Tung]{deepwrinkles}
Zorah L\"{a}hner, Daniel Cremers, and Tony Tung.
\newblock Deepwrinkles: Accurate and realistic clothing modeling.
\newblock In \emph{Computer Vision – ECCV 2018: 15th European Conference,
  Munich, Germany, September 8-14, 2018, Proceedings, Part IV}, page 698–715,
  Berlin, Heidelberg, 2018. Springer-Verlag.

\bibitem[Lazova et~al.(2019)Lazova, Insafutdinov, and Pons-Moll]{lazova2019360}
Verica Lazova, Eldar Insafutdinov, and Gerard Pons-Moll.
\newblock 360-degree textures of people in clothing from a single image.
\newblock In \emph{2019 International Conference on 3D Vision (3DV)}, pages
  643--653. IEEE, 2019.

\bibitem[Li et~al.(2017)Li, Zhao, Wei, Lang, Li, and Feng]{mhparser}
Jianshu Li, Jian Zhao, Yunchao Wei, Congyan Lang, Yidong Li, and Jiashi Feng.
\newblock Towards real world human parsing: Multiple-human parsing in the wild.
\newblock \emph{CoRR}, abs/1705.07206, 2017.

\bibitem[Li and Hoiem(2018)]{8107520}
Zhizhong Li and Derek Hoiem.
\newblock Learning without forgetting.
\newblock \emph{IEEE Transactions on Pattern Analysis and Machine
  Intelligence}, 40\penalty0 (12):\penalty0 2935--2947, 2018.

\bibitem[Liu et~al.(2016)Liu, Luo, Qiu, Wang, and Tang]{liu2016deepfashion}
Ziwei Liu, Ping Luo, Shi Qiu, Xiaogang Wang, and Xiaoou Tang.
\newblock Deepfashion: Powering robust clothes recognition and retrieval with
  rich annotations.
\newblock In \emph{Proceedings of IEEE Conference on Computer Vision and
  Pattern Recognition (CVPR)}, 2016.

\bibitem[Loper et~al.(2015)Loper, Mahmood, Romero, Pons-Moll, and
  Black]{SMPL:2015}
Matthew Loper, Naureen Mahmood, Javier Romero, Gerard Pons-Moll, and Michael~J.
  Black.
\newblock {SMPL}: A skinned multi-person linear model.
\newblock \emph{ACM Trans. Graphics (Proc. SIGGRAPH Asia)}, 34\penalty0
  (6):\penalty0 248:1--248:16, 2015.

\bibitem[Ma et~al.(2020)Ma, Yang, Ranjan, Pujades, Pons-Moll, Tang, and
  Black]{cape}
Qianli Ma, Jinlong Yang, Anurag Ranjan, Sergi Pujades, Gerard Pons-Moll, Siyu
  Tang, and Michael~J. Black.
\newblock {Learning to Dress 3D People in Generative Clothing}.
\newblock In \emph{Computer Vision and Pattern Recognition (CVPR)}, 2020.

\bibitem[Ma et~al.(2021)Ma, Yang, Tang, and Black]{resynth}
Qianli Ma, Jinlong Yang, Siyu Tang, and Michael~J. Black.
\newblock The power of points for modeling humans in clothing.
\newblock In \emph{Proceedings of the IEEE/CVF International Conference on
  Computer Vision (ICCV)}, 2021.

\bibitem[Meyer and Do(2015)]{Meyer20153DGI}
Gregory~P. Meyer and Minh~N. Do.
\newblock 3d grabcut: Interactive foreground extraction for reconstructed 3d
  scenes.
\newblock In \emph{3DOR@Eurographics}, 2015.

\bibitem[Mihajlovic et~al.(2022)Mihajlovic, Saito, Bansal, Zollhoefer, and
  Tang]{coap}
Marko Mihajlovic, Shunsuke Saito, Aayush Bansal, Michael Zollhoefer, and Siyu
  Tang.
\newblock {COAP}: Compositional articulated occupancy of people.
\newblock In \emph{Proceedings IEEE Conf. on Computer Vision and Pattern
  Recognition (CVPR)}, 2022.

\bibitem[Musoni et~al.(2023)Musoni, Melzi, and Castellani]{gim3dplus}
Pietro Musoni, Simone Melzi, and Umberto Castellani.
\newblock Gim3d plus: A labeled 3d dataset to design data-driven solutions for
  dressed humans.
\newblock \emph{Graphical Models}, 129:\penalty0 101187, 2023.

\bibitem[Park et~al.(2019)Park, Florence, Straub, Newcombe, and
  Lovegrove]{park2019deepsdf}
Jeong~Joon Park, Peter Florence, Julian Straub, Richard Newcombe, and Steven
  Lovegrove.
\newblock Deepsdf: Learning continuous signed distance functions for shape
  representation.
\newblock In \emph{Proceedings of the IEEE/CVF conference on computer vision
  and pattern recognition}, pages 165--174, 2019.

\bibitem[Patel et~al.(2020)Patel, Liao, and Pons-Moll]{patel2020}
Chaitanya Patel, Zhouyincheng Liao, and Gerard Pons-Moll.
\newblock The virtual tailor: Predicting clothing in {3D} as a function of
  human pose, shape and garment style.
\newblock In \emph{{IEEE} Conference on Computer Vision and Pattern Recognition
  (CVPR)}. {IEEE}, 2020.

\bibitem[Pietro~Musoni(2022)]{gim3d}
Umberto~Castellani Pietro~Musoni, Simone~Melzi.
\newblock {GIM3D}: A 3d dataset for garment segmentation.
\newblock \emph{placeholder}, 2022.

\bibitem[Pons-Moll et~al.(2017)Pons-Moll, Pujades, Hu, and Black]{clothcap}
Gerard Pons-Moll, Sergi Pujades, Sonny Hu, and Michael Black.
\newblock Clothcap: Seamless 4d clothing capture and retargeting.
\newblock \emph{ACM Transactions on Graphics, (Proc. SIGGRAPH)}, 36\penalty0
  (4), 2017.
\newblock Two first authors contributed equally.

\bibitem[Qi et~al.(2017)Qi, Yi, Su, and Guibas]{pointnet2}
Charles~R. Qi, Li Yi, Hao Su, and Leonidas~J. Guibas.
\newblock Pointnet++: Deep hierarchical feature learning on point sets in a
  metric space.
\newblock In \emph{Proceedings of the 31st International Conference on Neural
  Information Processing Systems}, page 5105–5114, Red Hook, NY, USA, 2017.
  Curran Associates Inc.

\bibitem[Qian et~al.(2021)Qian, Liu, Xu, and Lu]{iseg3d}
Sucheng Qian, Liu Liu, Wenqiang Xu, and Cewu Lu.
\newblock iseg3d: An interactive 3d shape segmentation tool.
\newblock 2021.

\bibitem[Qin et~al.(2020)Qin, Zhang, Huang, Dehghan, Zaiane, and
  Jagersand]{u2net}
Xuebin Qin, Zichen Zhang, Chenyang Huang, Masood Dehghan, Osmar Zaiane, and
  Martin Jagersand.
\newblock U2-net: Going deeper with nested u-structure for salient object
  detection.
\newblock page 107404, 2020.

\bibitem[Rother et~al.(2004)Rother, Kolmogorov, and Blake]{rother2004grabcut}
C. Rother, V. Kolmogorov, and A. Blake.
\newblock Grabcut: Interactive foreground extraction using iterated graph cuts.
\newblock 2004.

\bibitem[Saito et~al.(2021)Saito, Yang, Ma, and Black]{scanimate}
Shunsuke Saito, Jinlong Yang, Qianli Ma, and Michael~J. Black.
\newblock {SCANimate}: Weakly supervised learning of skinned clothed avatar
  networks.
\newblock In \emph{Proceedings IEEE/CVF Conf.~on Computer Vision and Pattern
  Recognition (CVPR)}, 2021.

\bibitem[Shao et~al.(2022)Shao, Zheng, Zhang, Sun, and Liu]{thuman5}
Ruizhi Shao, Zerong Zheng, Hongwen Zhang, Jingxiang Sun, and Yebin Liu.
\newblock Diffustereo: High quality human reconstruction via diffusion-based
  stereo using sparse cameras.
\newblock In \emph{ECCV}, 2022.

\bibitem[Sofiiuk et~al.(2020)Sofiiuk, Petrov, Barinova, and
  Konushin]{petrov20fBRS}
Konstantin Sofiiuk, Ilya Petrov, Olga Barinova, and Anton Konushin.
\newblock F-brs: Rethinking backpropagating refinement for interactive
  segmentation.
\newblock In \emph{{IEEE} Conference on Computer Vision and Pattern Recognition
  (CVPR)}. {IEEE}, 2020.

\bibitem[Su et~al.(2023)Su, Yu, Wang, and Liu]{thuman3}
Zhaoqi Su, Tao Yu, Yangang Wang, and Yebin Liu.
\newblock Deepcloth: Neural garment representation for shape and style editing.
\newblock \emph{IEEE Transactions on Pattern Analysis and Machine
  Intelligence}, 45\penalty0 (2):\penalty0 1581--1593, 2023.

\bibitem[Thomas et~al.(2019)Thomas, Qi, Deschaud, Marcotegui, Goulette, and
  Guibas]{thomas2019KPConv}
Hugues Thomas, Charles~R. Qi, Jean-Emmanuel Deschaud, Beatriz Marcotegui,
  Fran{\c{c}}ois Goulette, and Leonidas~J. Guibas.
\newblock Kpconv: Flexible and deformable convolution for point clouds.
\newblock \emph{Proceedings of the IEEE International Conference on Computer
  Vision}, 2019.

\bibitem[Tiwari et~al.(2020)Tiwari, Bhatnagar, Tung, and
  Pons-Moll]{tiwari20sizer}
Garvita Tiwari, Bharat~Lal Bhatnagar, Tony Tung, and Gerard Pons-Moll.
\newblock {SIZER}: A dataset and model for parsing {3D} clothing and learning
  size sensitive {3D} clothing.
\newblock In \emph{ECCV}, 2020.

\bibitem[Tiwari et~al.(2021)Tiwari, Sarafianos, Tung, and
  Pons-Moll]{tiwari21neuralgif}
Garvita Tiwari, Nikolaos Sarafianos, Tony Tung, and Gerard Pons-Moll.
\newblock {Neural-GIF}: Neural generalized implicit functions for animating
  people in clothing.
\newblock In \emph{ICCV}, 2021.

\bibitem[van~den Oord et~al.(2017)van~den Oord, Vinyals, and
  Kavukcuoglu]{vqvae}
Aaron van~den Oord, Oriol Vinyals, and Koray Kavukcuoglu.
\newblock Neural discrete representation learning.
\newblock In \emph{Proceedings of the 31st International Conference on Neural
  Information Processing Systems}, page 6309–6318, Red Hook, NY, USA, 2017.
  Curran Associates Inc.

\bibitem[Wang et~al.(2023)Wang, Zhang, Su, and Zhu]{continualsurvery}
Liyuan Wang, Xingxing Zhang, Hang Su, and Jun Zhu.
\newblock A comprehensive survey of continual learning: Theory, method and
  application, 2023.

\bibitem[Wang et~al.(2018{\natexlab{a}})Wang, Ceylan, Popovic, and
  Mitra]{garmentdesign_Wang_SA18}
Tuanfeng~Y. Wang, Duygu Ceylan, Jovan Popovic, and Niloy~J. Mitra.
\newblock Learning a shared shape space for multimodal garment design.
\newblock \emph{ACM Trans. Graph.}, 37\penalty0 (6):\penalty0 1:1--1:14,
  2018{\natexlab{a}}.

\bibitem[Wang et~al.(2018{\natexlab{b}})Wang, Yu, Huang, and
  Neumann]{wang2018sgpn}
Weiyue Wang, Ronald Yu, Qiangui Huang, and Ulrich Neumann.
\newblock Sgpn: Similarity group proposal network for 3d point cloud instance
  segmentation.
\newblock In \emph{CVPR}, 2018{\natexlab{b}}.

\bibitem[Wang et~al.(2019)Wang, Sun, Liu, Sarma, Bronstein, and Solomon]{dgcnn}
Yue Wang, Yongbin Sun, Ziwei Liu, Sanjay~E. Sarma, Michael~M. Bronstein, and
  Justin~M. Solomon.
\newblock Dynamic graph cnn for learning on point clouds.
\newblock \emph{ACM Transactions on Graphics (TOG)}, 2019.

\bibitem[Wiersma et~al.(2022)Wiersma, Nasikun, Eisemann, and
  Hildebrandt]{Wiersma2022DeltaConv}
R Wiersma, A. Nasikun, E Eisemann, and K Hildebrandt.
\newblock Deltaconv: Anisotropic operators for geometric deep learning on point
  clouds.
\newblock \emph{Transactions on Graphics}, 41\penalty0 (4), 2022.

\bibitem[Yamaguchi et~al.(2012)Yamaguchi, Kiapour, Ortiz, and
  Berg]{parsingclothing}
Kota Yamaguchi, Mohammad~Hadi Kiapour, Luis~E. Ortiz, and Tamara~L. Berg.
\newblock Parsing clothing in fashion photographs.
\newblock \emph{2012 IEEE Conference on Computer Vision and Pattern
  Recognition}, pages 3570--3577, 2012.

\bibitem[Yamaguchi et~al.(2013)Yamaguchi, Kiapour, and
  Berg]{yamaguchi2013paperdoll}
Kota Yamaguchi, M.~Hadi Kiapour, and Tamara~L. Berg.
\newblock Paper doll parsing: Retrieving similar styles to parse clothing
  items.
\newblock In \emph{{IEEE} International Conference on Computer Vision, {ICCV}
  2013, Sydney, Australia, December 1-8, 2013}, pages 3519--3526. {IEEE}
  Computer Society, 2013.

\bibitem[Yang and Luo(2014)]{yang2014cvpr}
W. Yang and L. Luo, P.and~Lin.
\newblock Clothing co-parsing by joint image segmentation and labeling.
\newblock 2014.

\bibitem[Yenamandra et~al.(2021)Yenamandra, Tewari, Bernard, Seidel, Elgharib,
  Cremers, and Theobalt]{yenamandra2020i3dmm}
T Yenamandra, A Tewari, F Bernard, HP Seidel, M Elgharib, D Cremers, and C
  Theobalt.
\newblock i3dmm: Deep implicit 3d morphable model of human heads.
\newblock In \emph{Proceedings of the IEEE / CVF Conference on Computer Vision
  and Pattern Recognition (CVPR)}, 2021.

\bibitem[Yu et~al.(2021)Yu, Zheng, Guo, Liu, Dai, and Liu]{thuman2}
Tao Yu, Zerong Zheng, Kaiwen Guo, Pengpeng Liu, Qionghai Dai, and Yebin Liu.
\newblock Function4d: Real-time human volumetric capture from very sparse
  consumer rgbd sensors.
\newblock In \emph{IEEE Conference on Computer Vision and Pattern Recognition
  (CVPR2021)}, 2021.

\bibitem[Zhang et~al.(2017)Zhang, Pujades, Black, and Pons-Moll]{buff}
Chao Zhang, Sergi Pujades, Michael~J. Black, and Gerard Pons-Moll.
\newblock Detailed, accurate, human shape estimation from clothed 3d scan
  sequences.
\newblock In \emph{The IEEE Conference on Computer Vision and Pattern
  Recognition (CVPR)}, 2017.

\bibitem[Zhao et~al.(2020)Zhao, Jiang, Jia, Torr, and Koltun]{pointtransformer}
Hengshuang Zhao, Li Jiang, Jiaya Jia, Philip Torr, and Vladlen Koltun.
\newblock Point transformer, 2020.

\bibitem[Zhao et~al.(2018)Zhao, Li, Cheng, Zhou, Sim, Yan, and
  Feng]{crowdedscenes}
Jian Zhao, Jianshu Li, Yu Cheng, Li Zhou, Terence Sim, Shuicheng Yan, and
  Jiashi Feng.
\newblock Understanding humans in crowded scenes: Deep nested adversarial
  learning and {A} new benchmark for multi-human parsing.
\newblock \emph{CoRR}, abs/1804.03287, 2018.

\bibitem[Zheng et~al.(2021)Zheng, Shao, Zhang, Yu, Zheng, Dai, and
  Liu]{multihuman}
Yang Zheng, Ruizhi Shao, Yuxiang Zhang, Tao Yu, Zerong Zheng, Qionghai Dai, and
  Yebin Liu.
\newblock Deepmulticap: Performance capture of multiple characters using sparse
  multiview cameras.
\newblock In \emph{IEEE Conference on Computer Vision (ICCV 2021)}, 2021.

\bibitem[Zheng et~al.(2022)Zheng, Huang, Yu, Zhang, Guo, and Liu]{thuman4}
Zerong Zheng, Han Huang, Tao Yu, Hongwen Zhang, Yandong Guo, and Yebin Liu.
\newblock Structured local radiance fields for human avatar modeling.
\newblock In \emph{Proceedings of the IEEE/CVF Conference on Computer Vision
  and Pattern Recognition (CVPR)}, 2022.

\bibitem[Zhou et~al.(2018)Zhou, Park, and Koltun]{open3d}
Qian-Yi Zhou, Jaesik Park, and Vladlen Koltun.
\newblock Open3d: A modern library for 3d data processing, 2018.
\newblock cite arxiv:1801.09847Comment: http://www.open3d.org.

\bibitem[Zhu et~al.(2020)Zhu, Cao, Jin, Chen, Du, Wang, Cui, and
  Han]{deepfashion3d}
Heming Zhu, Yu Cao, Hang Jin, Weikai Chen, Dong Du, Zhangye Wang, Shuguang Cui,
  and Xiaoguang Han.
\newblock Deep fashion3d: A dataset and benchmark for 3d garment reconstruction
  from single images, 2020.

\end{thebibliography}
